\documentclass[10pt,twocolumn,letterpaper,notitlepage]{article}
\usepackage{cite}
\usepackage{amsmath}
\usepackage{texnames}
\usepackage{graphicx}
\usepackage[table]{xcolor}
\usepackage{ahsmeeting}

\usepackage{multirow}
\usepackage{multicol}

\usepackage{url}
\usepackage{lipsum}
\usepackage[labelfont=bf, labelsep=period, textfont=bf]{caption}
\usepackage[colorlinks=true]{hyperref}
\definecolor{halfgreen}{RGB}{0,128,0}
\definecolor{ahsred}{RGB}{192,0,0}

\usepackage{stfloats}

\meeting{}

\newcounter{isorefi}
{\endlist}

\begin{document}

\title{A Vision-Based Control Method for Autonomous Landing of\\  Vertical Flight Aircraft On a Moving Platform Without Using GPS}

\author{
  \begin{tabular}{cc}
    \shortstack{\textbf{Bochan Lee} \\ Graduate Research Assistant \\ Texas A\&M University \\
College Station, TX, USA}
 &
   \shortstack{\textbf{Vishnu Saj}\\ Graduate Student \\ Texas A\&M University \\
College Station, TX, USA}\\
    \\
    \shortstack{\textbf{Moble Benedict}\\ Assistant Professor \\ Texas A\&M University \\
College Station, TX, USA}
 &
   \shortstack{\textbf{Dileep Kalathil}\\ Assistant Professor \\ Texas A\&M University \\
College Station, TX, USA}
\end{tabular}
    }
\date{}

\abstract{The paper discusses a novel vision-based estimation and control approach to enable fully autonomous tracking and landing of vertical take-off and landing (VTOL) capable unmanned aerial vehicles (UAVs) on moving platforms without relying on a GPS signal. A unique feature of the present method is that it accomplishes this task without tracking the landing pad itself; however, by utilizing a standardized visual cue installed normal to the landing pad and parallel to the pilot's/vehicle's line of sight. A computer vision system using a single monocular camera is developed to detect the visual cue and then accurately estimate the heading of the UAV and its relative distances in all three directions to the landing pad. Through comparison with a Vicon-based motion capture system, the capability of the present vision system to measure distances in real-time within an accuracy of less than a centimeter and heading within a degree with the right visual cue, is demonstrated. A gain-scheduled proportional integral derivative (PID) control system is integrated with the vision system and then implemented on a quad-rotor-UAV dynamic model in a realistic simulation program called Gazebo. Extensive simulations are conducted to demonstrate the ability of the controller to achieve robust tracking and landing on platforms moving in arbitrary trajectories. Repeated flight tests, using both stationary and moving platforms are successfully conducted with less than 5 centimeters of landing error.}

\maketitle
\section{Notation}

\symbolentry{A}{Camera intrinsic matrix}
\symbolentry{c_u}{Corner position-column}
\symbolentry{c_v}{Corner position-row}
\symbolentry{f_x}{Focal length of camera in horizontal direction}
\symbolentry{f_y}{Focal length of camera in vertical direction}
\symbolentry{K_D}{Derivative gain}
\symbolentry{K_I}{Integral gain}
\symbolentry{K_P}{Proportional gain}
\symbolentry{R}{Rotation matrix}
\symbolentry{R_{basic}}{Basic rotation matrix}
\symbolentry{R_{modified}}{Modified rotation matrix}
\symbolentry{s}{Scaling factor}
\symbolentry{t}{Translation vector}
\symbolentry{u}{Image pixel position-column}
\symbolentry{u_0}{Center of image-column}
\symbolentry{v}{Image pixel position-row}
\symbolentry{v_0}{Center of image-row}
\symbolentry{X}{Sideward relative distance, meter}
\symbolentry{Y}{Vertical relative distance, meter}
\symbolentry{Z}{Forward relative distance, meter}
\symbolentry{\alpha}{Relative yaw angle, rad}
\symbolentry{\beta}{Relative pitch angle, rad}
\symbolentry{\gamma}{Relative roll angle, rad}
\symbolentry{\theta}{Aircraft pitch angle, deg}
\symbolentry{\rho_u}{Parameter to convert pixels to SI units-column}
\symbolentry{\rho_v}{Parameter to convert pixels to SI units-row}
\symbolentry{\phi}{Aircraft roll angle, deg}
\symbolentry{\psi}{Aircraft yaw angle, deg}
\symbolentry{\nabla f(u,v)}{Image gradient}

\section{Introduction}

Fully autonomous approach and landing of vertical take-off and landing (VTOL) capable unmanned aerial vehicles (UAVs) on moving platforms without relying on GPS is an extremely challenging task. In order to accomplish this, the flight controller has to first perceive the motion of the platform using machine vision, then find the relative distances, velocities, and heading of the aircraft with respect to the platform and then control the aircraft so that it can keep up with the platform motion and finally execute a touch-down. Such a maneuver is demanding even for human pilots and therefore, its automation is of significant interest whether the aircraft is manned or unmanned.

Out of all the possible moving platforms, the most difficult one to land on is a small ship at sea due to the limited landing space, six degrees of freedom ship deck motions, limited visual references for pilots, and lack of alternative landing spots as shown in \href{https://youtu.be/lDISL-jsF-Q?t=60}{Video} \cite{Ruptly}. Despite these challenges, the Navy helicopter pilots have managed to successfully land on ships for decades. Therefore, it is imperative to understand and gain inspiration from the the Navy helicopter ship landing procedure, which is discussed in Refs. \cite{lumsden1998challenges, colwell2002maritime}. The present study began by asking two key questions: (1) what is the secret behind the success of Navy helicopter pilots when it comes to ship landing? and (2) can some of the key ideas from this procedure be applied to VTOL UAVs landing on moving platforms?

Contrary to intuition, Navy helicopter pilots land their aircraft without visually focusing on the landing spot. In fact, Navy pilots are trained not to follow ship deck motions for two main reasons. First, spatial disorientation can occur when a pilot has no fixed, visible horizon to refer to, which is a critical element for maintaining a proper sense of helicopter attitude independent of ship motions. Second, constantly changing the helicopter pose to match ship deck motions can trigger unstable helicopter dynamics introducing serious potential hazards. Therefore, a pilot tries to control the helicopter in a stable manner independent of the ship's roll and pitch motions, and then lands vertically. Vertical landing on a moving ship deck is tested and proven to be safe within the operating limits. It is also recommended to land quickly in order to prevent the ship deck from impacting one of the helicopter landing gear skids or wheels causing a rollover. 

\begin{figure}[hbt!]
\centering
\includegraphics[width=0.35\textwidth]{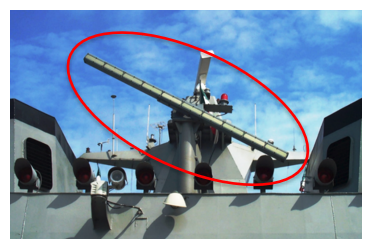}
\caption{Horizon Reference Bar}
\label{hrb}
\end{figure}

If the pilot is not looking at the ship deck while landing, then what is the visual cue that he/she is referring to? The key visual aid that helps pilots to land safely on ships is a "horizon reference bar" shown in Fig. \ref{hrb}, which is gyro-stabilized to indicate a perfect horizon independent of ship motions and is widely used in most modern Navies  \cite{stingl1970vtol,nato}. Pilots therefore control the helicopter by referencing the horizon bar instead of responding to ship motions. 

The present study adopts this proven ship landing method and automates it to land on moving platforms. First, a camera plays the role of pilot's eyes, and a computer vision system processes the visual information from images. Then, a control system uses the visual data to manipulate the UAV similar to a pilot controlling a helicopter. This vision-based control system can either be image- or position-based, depending on how the visual data is used. The image-based method directly uses visual information obtained from a 2D camera view while position-based control first estimates the pose of an object by transforming the data from a 2D camera plane into 3D space. Previous efforts to land a UAV on a moving platform have used image-based control to center the landing platform in a downward-facing 2D camera view \cite{lee2012autonomous}, and position-based control to minimize the relative distance from the landing platform in 3D \cite{wenzel2011automatic}. Despite both techniques showing limited feasibility for UAV applications, these approaches cannot be extended to general landing cases because they use external infrastructure such as a Vicon motion capture system \cite{vicon} to feed states into the control loop \cite{lee2012autonomous} or infrared(IR)-LEDs on the landing platform and onboard IR camera \cite{wenzel2011automatic}. There have been other studies which also achieved autonomous UAV landing on a moving platform by using a proportional derivative (PD) controller in the presence of time delays \cite{daly2015coordinated}, and a linear quadratic regulator (LQR) control system in real-time \cite{ghamry2016real}. However, along with the UAV, they also control the moving platform to be placed at a favorable position for UAV landing, which cannot be applied to a general case of an independently moving platform. 

For a practical vision-based control solution, one should achieve the landing task with common sensors available on UAVs and without manipulating the landing platform motions. In recent years, there have been several studies that sought for a general solution by combining the information from the UAV's inertial measurement unit (IMU) with the estimated pose from captured images \cite{araar2017vision} or combining UAV's IMU information with measured distances from a distance sensor \cite{falanga2017vision}. There have also been other studies which utilize either red, green and blue (RGB) color information of a landing platform \cite{bi2013implementation} or an optical-flow based control method which uses relative velocity and proximity information \cite{herisse2011landing}. Although these vision-based methods achieved autonomous UAV landing on platforms undergoing simple motions by using minimum sensors such as a single onboard camera and IMU, all of them visually tracked the landing platform, unlike a helicopter ship landing procedure where the pilots do not look at the ship deck. An approach that involves tracking the motion of the landing spot is not ideal for the VTOL UAV landing because (1) in order to track the landing spot, the UAV needs to approach from a high altitude, which might not be desirable or even possible in many applications, (2) in scenarios where the platform is undergoing complex motions, for example, a small ship in moderately-high sea states or a Humvee traversing rugged terrain, a sophisticated vision system is required to capture the large, unpredictable platform motions, and (3) even if the platform motion can be tracked precisely, there is always a potential hazard caused by the control system forcing the UAV to match the complex platform movements. 

\begin{figure}[hbt!]
\centering
\includegraphics[width=0.4\textwidth]{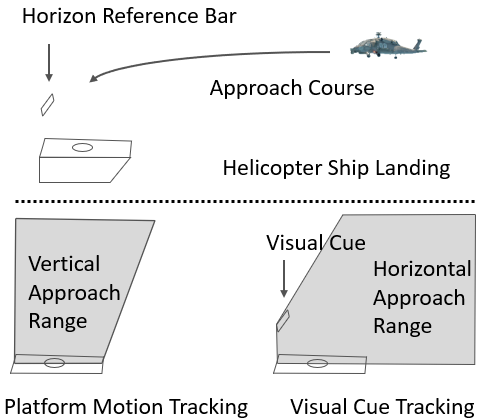}
\caption{Comparison of Previous Platform Motion Tracking with Present Visual Cue Tracking Method}
\label{method_comp}
\end{figure}

Conversely, this paper presents a vision-based autonomous landing solution for VTOL UAVs which overcomes these drawbacks by implementing the Navy helicopter ship landing procedure by tracking a gyro-stabilized horizon reference bar rather than the moving platform itself. In the present study, a single front-facing camera tracks a standardized visual cue (analogous to a horizon bar) that is located beyond the landing spot and installed parallel to the approach course. As shown in Fig.  \ref{method_comp}, referring to the visual cue allows the UAV to approach horizontally similar to the helicopter ship landing procedure. Therefore, its operating range is greater in the horizontal direction and it can be used for both approach and landing phase. However, the operating range for a platform-motion-tracking approach is limited to the vertical space, and therefore, it can be used only for the final landing phase and requires another method for approach. Therefore, the method presented in this study is fundamentally different than previous UAV landing techniques, and is applicable for all types of VTOL aircraft.

For the present study, instead of a standard horizon bar, a checkerboard pattern is used as a visual cue since it is easily distinguishable from its surroundings and also offers redundancy with multiple corners that can be detected. Gain-scheduled proportional integral derivative (PID) controllers are configured to command roll, pitch, yaw, and throttle to the UAV. As the UAV advances to the landing spot, multiple layers of controllers are activated according to the relative distance from the landing platform because that allows the UAV to approach at a higher speed when it is farther away and then slow-down and carefully adjust its pose when it reaches close proximity of the landing spot and eventually executes landing maneuver similar to how a remote helicopter pilot would do. In addition, the robust autonomous flight characteristics are even more enhanced by configuring different flight modes, which include a scanning mode to search for the visual cue in case of detection failure, a tracking mode that is activated during normal flight conditions, and safety mode for responding appropriately to unexpected situations.   

To demonstrate the feasibility of this novel method, a systematic approach has been taken from development to validation. The outline of the paper is as follows: (1) computer vision system development and validation using the Vicon motion capture system \cite{vicon}, (2) control system architecture and the configured flight modes, (3) results from the Gazebo simulation \cite{gazebo} conducted for various scenarios, (4) flight-testing results from autonomous landings on both stationary and moving platforms, and (5) the conclusions inferred from this study.

\section{Computer Vision System}

Vision-based control has made its mark in autonomous navigation systems with the advancement of computer vision algorithms. There have been many studies in this field to develop methodologies for detecting and extracting desired visual features from the surrounding environment \cite{compvision1,compvision2,compvision3}. The extracted data are used for various purposes such as pose estimation, object recognition, reconstruction, and so forth. Among multiple methods, the projective transformation techniques\cite{projtrans} are closely referenced for computing relative position and orientation. 

\begin{figure}[hbt!]
\centering
\includegraphics[width=0.45\textwidth]{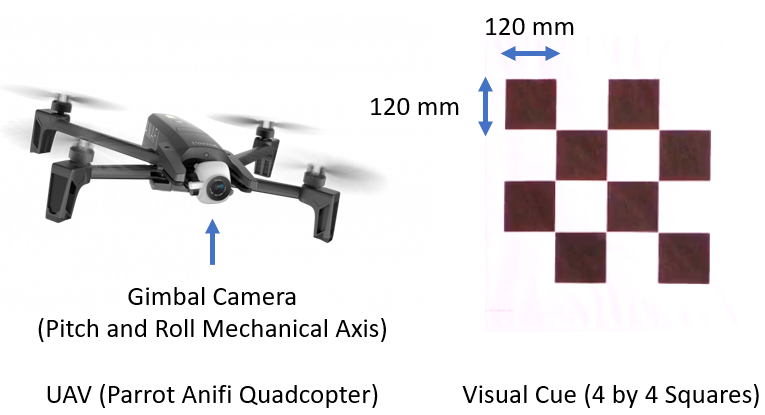}
\caption{UAV with Gimbal Camera and Visual Cue}
\label{exp_vision}
\end{figure}

In this study, a computer vision system is developed to provide accurate position and orientation information. The contributions of the present study in computer vision include improvement of a previous detection method \cite{projtrans}, integration of the detection and estimation methods, and validation of the entire vision system using the Vicon motion capture system. The Parrot Anafi quad-rotor UAV and the checker-board-based visual cue which have been used throughout the study are shown in Fig. \ref{exp_vision}.

The onboard front-facing camera has pitch and roll mechanical gimbals, which compensate for the pitch and roll motion of the UAV and keeps the camera level with the horizon. Thus, images captured by the camera do not experience the effects of the pitch and roll motions. This reduces the complexity caused by rotational motions. However, the effect of yaw is reflected in the images, which is utilized to extract the heading of the UAV with respect to the visual cue. The maximum resolution of the camera is 4k ultra-high definition, UHD (4096 x 2160) and the higher resolution helps in increasing the detection range. The size of squares on the checkerboard is the other factor that affects the detection range. As the square size increases, the effective range for detection also increases. However, there is a trade-off between the maximum effective range and the closest proximity from the visual cue since the camera has to be located farther away for the visual cue detection in case of a bigger square size. The effective range for each case is found through experiments and are provided in Table \ref{selection}.

\begin{table}[ht] \begin{minipage}{\columnwidth} \centering
\caption{\hbox{Valid Range for Varying Resolution \& Square Size}}
\label{selection}
\begin{tabular}{lcc} \arrayrulecolor{halfgreen} \hline \hline
Resolution & 80 mm Square & 120 mm Square \\ \hline
qHD(540p) & 13 m & 15.5 m \\ 
HD(720p) & 14 m & 17.5 m \\
FHD(1080p) & 15 m & 19 m \\
4K UHD(2160p) & 17 m & 25 m \\ \hline \hline
\end{tabular} \end{minipage}
\end{table}

In the present study, the square size is chosen to be 120 mm since it can be detected accurately from distances as high as 25 meters, and the square size is small enough to stay within the camera view even when the visual cue is only 1 meter away. Even though higher resolution increases the effective range, it also increases the data size to process. Therefore, HD 720p (1280 x 720) resolution is selected considering the latency in livestreaming and image processing for a relatively high bandwidth integrated vision-based feedback control system as well as a good effective range. In the following sections, detailed discussions of the detection and estimation methods are provided along with its validation.

\subsection{Detection}

\begin{figure}[hbt!]
\centering
\includegraphics[width=0.485\textwidth]{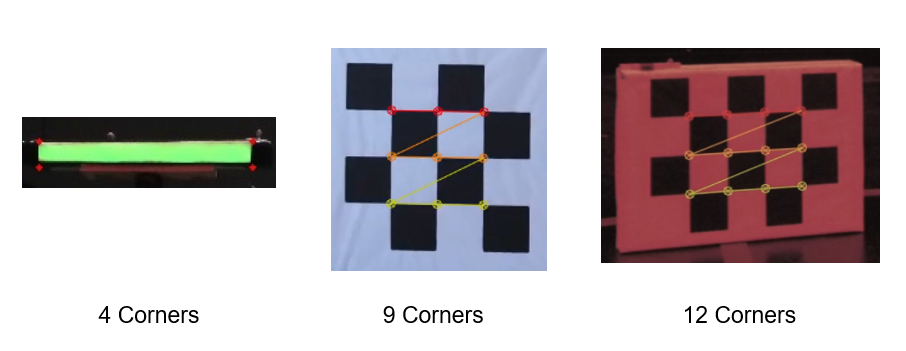}
\caption{Available Representative Visual Cues}
\label{visualcues}
\end{figure}

Detection is one of the most important aspects in the present computer vision system. The computation of the UAV's relative positions and heading angle depends heavily on how accurately the visual cue is detected. Note that for the present approach to work, the visual cue does not have to be of any special shape or size as long as its dimensions are known and it can be distinguished from its surroundings. Hence, a number of different visual cues are tested and representative cases are shown in Fig. \ref{visualcues}. 

Visual cues that vary in shape, size, and the number of corners are tested to determine an appropriate one for a given environment. 
Initially, a simple green colored bar is tested as a visual cue. It is basically a scaled-down version of the horizon reference bar used for ship landing. The green bar is detected using an HSV (hue, saturation, value) filter \cite{Cheng01colorimage}. Once the green bar is separated out from the surroundings, the positions of corners are detected using the Harris corner method \cite{compvision1}. However, the direct application of the HSV filtering method is not ideal since the HSV filter cannot guarantee clear and perfect capture of the entire bar portion from the background in a dynamic situation. Once the boundary of the bar is not distinctive, the Harris corner algorithm also yields false corners which results in an inaccurate estimation of the UAV's position and heading.
In order to solve this issue, a new algorithm is developed to enclose the green bar in a bounding rectangle with sharp corners.  

Despite the feasibility of the green bar, a visual cue that has more distinctive features and redundancy is considered in order to achieve robust VTOL UAV tracking and landing. The redundancy can be enhanced by having more corners. Hence, the checkerboard is selected as the visual cue for this study because its geometry exhibits local image features such as edges, lines, and corners which are greatly helpful in detection. 

\begin{figure}[hbt!]
\centering
\vspace{-0.4cm}
\includegraphics[width=0.48\textwidth]{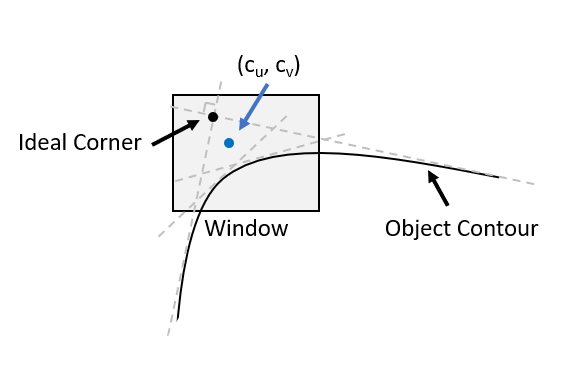}
\caption{F\"orstner Corner Detection Method}
\label{subpixel}
\end{figure}

In addition, in order to increase the accuracy in detection, F\"orstner corner detection method, which is a more advanced detection method than the Harris corner detection, is adopted for the present computer vision system \cite{forstner1987fast}. The F\"orstner corner detection increases the accuracy by sub-pixel refinement process. It is based on the fact that an ideal corner is a single point that tangent lines of the object cross perpendicular to each other as shown in Fig. \ref{subpixel}. 

As shown in Fig. \ref{subpixel}, the pixel information around a corner is not perfectly clear.
Therefore, an approximation process for defining the corner position $(c_u, c_v)$ is required and is expressed in Eq.~\eqref{forstner}.

\begin{equation}\label{forstner}
    (\hat{c_u},\hat{c_v})=\underset{c_u,c_v}{\arg\min}\sum_{u,v \in N}([\nabla f(u,v)]^T(u-c_u,v-c_v))^2
\end{equation}

The image gradient $\nabla f(u, v)$ at the image pixel position $(u, v)$ is perpendicular to line from $(u, v)$ to corner position $(c_u, c_v)$. In order to obtain $(c_u, c_v)$, a least square estimation in a small window is used. The projection of line from $(c_u, c_v)$ to $(u, v)$ on to the tangent line at $(u, v)$ is required to be maximized in the given window. In other words, the projection of image gradient on to the line segment connecting $(c_u, c_v)$ and $(u, v)$ has to be minimized for all $(u, v)$ inside the window $N$. This method is implemented by using the existing OpenCV checkerboard detector \cite{findchess}. 

This method accurately detects the corners, which improves the estimation quality; however, the detected corners are not recognized in any specific order. Since the order is not assigned, it can possibly experience unintended changes in the order, depending on the captured image. This poses an issue to establish a correct coordinate system since it yields the correct direction of axes only when the order of corners is recognized from the top left to bottom right. Hence, the present detection method is improved to resolve this issue by assigning the proper order of corners considering the checkerboard geometry and is then applied to simulations and flight tests. Using this method, different numbers of corners are also tested along with varying resolutions and square sizes as a visual cue selection process. The final selection is 9 corners for redundancy as well as considering the entire size of a visual cue with 120-millimeter squares.

\subsection{Estimation}

\begin{figure}[hbt!]
\centering
\includegraphics[width=0.48\textwidth]{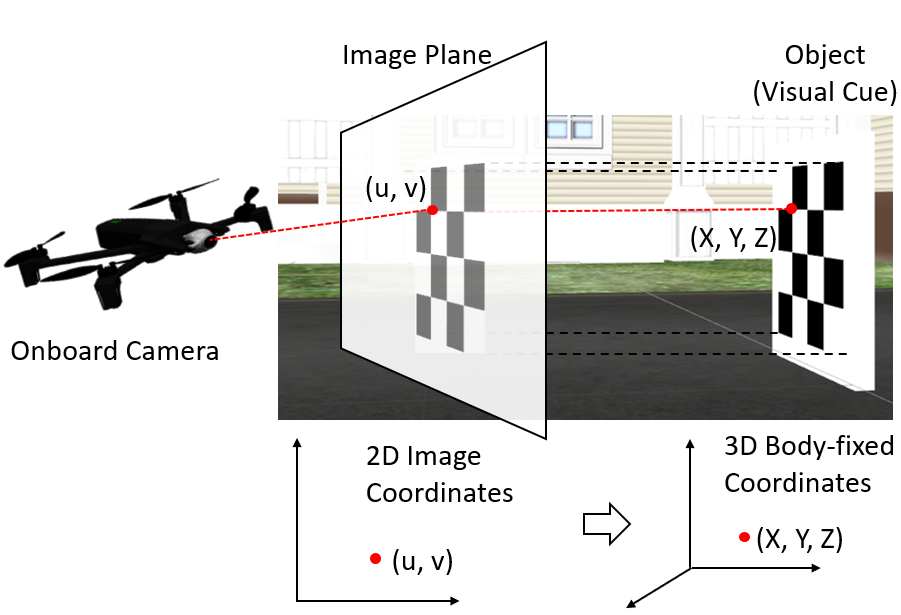}
\caption{Geometric Relation between Image and Object}
\label{Camera_model}
\end{figure}

The estimation is based on a single camera calibration method using a planar object \cite{est1,est2}. The geometric relation between a camera and a planar object is shown in Fig. \ref{Camera_model}.

A conventional pinhole camera model is used to derive the geometric relation. Consider a three-dimensional object (visual cue) in the space captured by a camera. A 3-D coordinate system can be defined with respect to the object and there is a 2-D coordinate system associated with the image frame. $(X, Y, Z)$ is a point on the object described in the visual cue body-fixed frame and $(u, v)$ is a corresponding point on the image frame (pixel position in the image). The relationship between the image coordinates and the body-fixed coordinates in matrix form is described in Eq.~\eqref{img_coord}. 

\begin{equation}
\label{img_coord}
s
 \begin{bmatrix} u \\ v \\ 1 \end{bmatrix}
 =
 \boldsymbol{A}
  \begin{bmatrix}
   \boldsymbol{R} & t \\
   0_{1x3} & 1\\
   \end{bmatrix}^{-1}
 \begin{bmatrix} X \\ Y \\ Z \\1 \end{bmatrix}  
\end{equation}

It is derived in a homogeneous coordinate system and converted to a cartesian coordinate system. $(u, v)$ is an image coordinate system and $s$ is a scaling factor that comes from transforming the homogeneous coordinates to cartesian coordinates. $R$ is a 3 x 3 rotation matrix and $t$ is a 3 x 1 translation vector. $R$ matrix has the information about the camera orientation with respect to the visual cue and $t$ gives the information of how far the camera is from the particular $(X, Y, Z)$ point in the visual cue body-fixed frame. $O_{1x3}$ is 1 x 3 zero matrix. The matrix which includes $R$, $t$, $O_{1x3}$, and 1 is called the camera extrinsic matrix, which varies from image to image. The matrix $A$ is called the camera intrinsic matrix and its components are shown in Eq.~\eqref{A_mat}.

\begin{equation} \label{A_mat}
\boldsymbol{A}
 =
  \begin{bmatrix}
   1/\rho_{u} & 0 & u_{0} \\
   0 & 1/\rho_{v} & v_{0} \\
   0 & 0 & 1 \\
  \end{bmatrix}
  \begin{bmatrix}
   f_{x} & 0 & 0 & 0 \\
   0 & f_{y} & 0 & 0 \\
   0 & 0 & 1 & 0 \\
  \end{bmatrix}
\end{equation}

$1/\rho_{u}$ and $1/\rho_{v}$ are parameters that scale the image coordinates, which are in pixels to values in the International System of Units (SI). $u_{0}$ and $v_{0}$ are the center position in the image plane. The first matrix shifts the center of the image plane to the top left corner point. The second matrix contains the focal length information $f_x$ and $f_y$, which are the focal lengths of the camera in the $x$ and $y$ directions, respectively. The product of two matrices forms the camera intrinsic matrix, which depends on the particular camera. This unique camera intrinsic matrix can be found using a camera calibration \cite{solvepnp}. 

By solving Eq.\eqref{img_coord}, rotation and the translation vectors are obtained. In computer vision, the estimation of $R$ and $t$ are called pose estimation. There are different algorithms to determine the pose of the camera. A common class of algorithms called the Perspective-n-Point (PnP) algorithm is applied. Given a set of $n$ 3-D coordinates of an object and its corresponding 2-D projections on the image, this algorithm can estimate the pose of the camera. There are 6 degrees of freedom (DOF) for a camera, which are 3 DOF in rotation (roll, pitch, yaw) and 3 DOF in translation $(X, Y, Z)$. A minimum of 3 points are required to find a solution, but the solution is not unique. There should be a minimum of 4 points to obtain a unique solution; however, it can be more reliable and redundant when there are more points. 
An iterative method is used for the PnP algorithm since it is robust for objects which consist of a planar surface. The iterative method is based on Levenberg-Marquardt optimization \cite{lev1,lev2}. In this method, the function finds such a pose that minimizes re-projection error, which is the sum of squared distances between the observed image points $(u, v)$ and projected object points $(X, Y, Z)$. By default, the iterative algorithm sets the initial value of rotation and translation as zero and then updates during each iteration. 

The OpenCV RANSAC method \cite{ransac, projtrans} is used to find a rough estimation for the extrinsic matrix. The RANSAC method also identifies the outliers and removes them during the calculation. The initial guess and inliers are fed into the iterative algorithm to have a more accurate estimation. The OpenCV solvePnP algorithm \cite{solvepnp} returns a rotation vector and the translation vector. The rotation vector can be converted to the rotation matrix using the Rodrigues function. Thus, once the rotation matrix ($R$) and translation vector ($t$) are obtained, $-R^{-1}t$ gives the relative distances in 3D from the camera to the origin of the visual cue body-fixed frame.

The present estimation method is modified from the existing algorithm to take advantage of the gimbal camera. Since the gimbal corrects for the roll and pitch motion of a UAV, the images only reflect the yaw motion (heading angle) with respect to the visual cue. Thus, roll and pitch angles computed by the image can be regarded as camera noise and gimbal correction errors. Therefore, the roll and pitch angles are excluded from the position estimation. The basic and modified rotation matrix are specified in Eq.~\eqref{basic_R} and ~\eqref{modified_R}, respectively. 

\begin{equation} \label{basic_R}
\boldsymbol{R_{basic}}
 =
  \begin{bmatrix}
   c{\alpha}c{\beta} & c{\alpha}s{\beta}s{\gamma}-s{\alpha}c{\gamma} & c{\alpha}s{\beta}c{\gamma}+s{\alpha}s{\gamma} \\
   s{\alpha}c{\beta} & s{\alpha}s{\beta}s{\gamma}-c{\alpha}c{\gamma} & s{\alpha}s{\beta}c{\gamma}+c{\alpha}s{\gamma} \\
   -s{\beta} & c{\beta}s{\gamma} & c{\beta}c{\gamma} \\
  \end{bmatrix}
\end{equation}

\begin{equation} \label{modified_R}
\boldsymbol{R_{modified}}
 =
  \begin{bmatrix}
   \cos{\alpha} & -\sin{\alpha} & 0 \\
   \sin{\alpha} & -\cos{\alpha} & 0 \\
   0 & 0 & 1 \\
  \end{bmatrix}
\end{equation}

$\alpha$, $\beta$, and $\gamma$ represent yaw, pitch, and roll angles, respectively. Each row of the $R$ matrix contributes to their respective $(X, Y, Z)$ coordinates. By setting pitch and roll angle to zero, the modified rotation matrix is obtained from the basic rotation matrix. The yaw angle is taken into consideration because it is required to set a UAV approaching course by maintaining a designated heading angle. The third row of the modified rotation matrix shows that the $Z$-coordinate is independent of the yaw angle $\alpha$. Hence, the yaw angle has no contributions in the estimation of forward relative distance ($Z$-coordinate) between the UAV and the visual cue. However, the sideward relative distance ($X$-coordinate) and vertical relative distance ($Y$-coordinate) are heavily dependent on the yaw angle. 

\begin{figure}[hbt!]
\vspace{-0.3cm}
 \raggedleft
\includegraphics[width=0.485\textwidth]{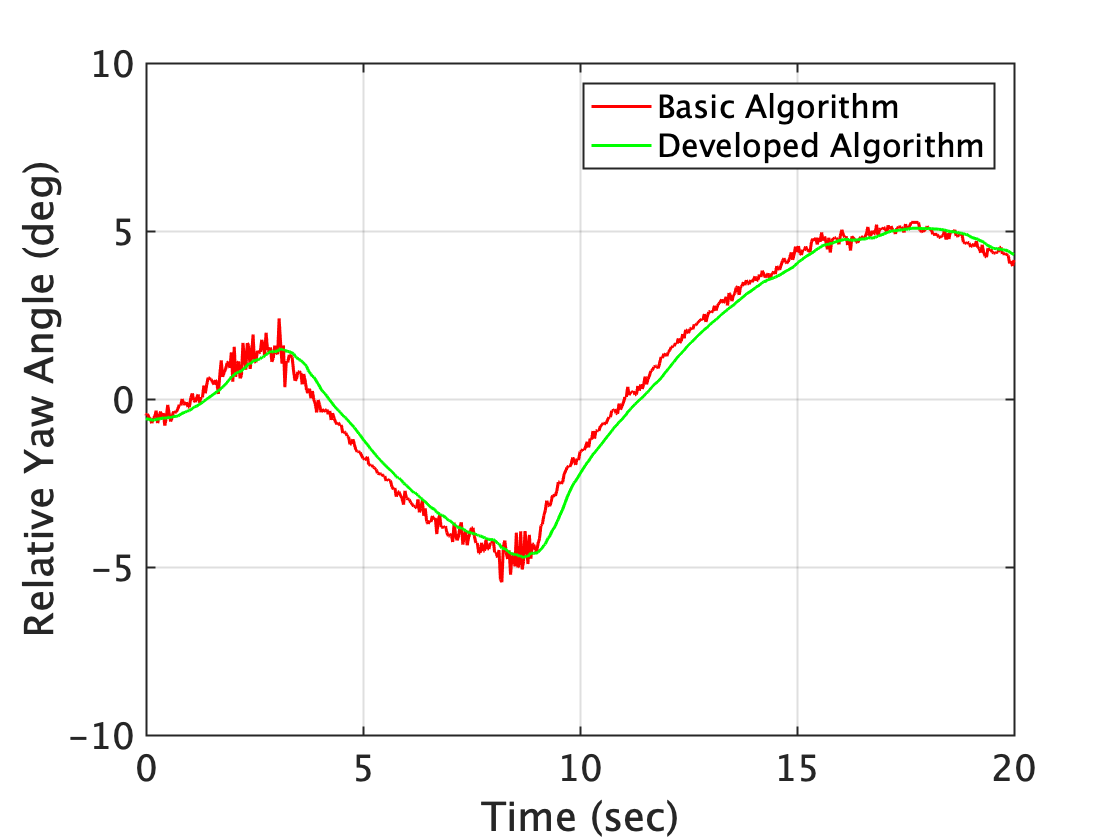}
\caption{Effect of Moving Average Filter on Yaw}
\label{average}
\end{figure}

It was apparent from multiple experiments that the yaw angle estimation becomes noisy as the camera gets farther away from the visual cue. Since the estimation is based on the number of visual cue pixels in the image, the changes in pixels with distance results in the noise. Despite this sensitivity issue, yaw estimation still shows a reasonable trend within the range that the forward relative distance is accurately estimated. To utilize the yaw estimation trend, instead of directly taking the noisy yaw angle estimation, a moving average filter is configured and the results are shown in Fig. \ref{average}. 

The red line denotes the yaw angle estimation results by the basic algorithm and the blue line denotes the results after the moving average filter is applied. The moving average is calculated using the history of the previous estimation data. A lower and upper bound are defined for the next yaw angle estimation value with respect to the current average. Whenever the estimated yaw angle is out of the defined range, it rejects that value and takes an average value. In this way, the moving averaging filter provides stable yaw estimation by tracking a trend in a further distance, and the accuracy naturally increases as it gets closer to the visual cue due to having more pixel data in the image.

In conclusion, the present estimation method is developed to satisfy the practical requirements for the UAV tracking and landing maneuver. When the UAV flies to the landing pad from a further distance, having a continuous smooth estimation for the current position is more important than the accuracy as long as the estimation is within a reasonable boundary. If the vision system provides fluctuating estimations to the control system, the UAV will move in an arbitrary manner in response to the fluctuations. When the UAV is close to the landing pad, the accuracy is crucial and the vision system provides smooth and precise estimations with the help of the moving average filter and more pixel data in the image.  

\subsection{Validation}

The results obtained from the developed computer vision system are validated through position and attitude measurements using a Vicon motion capture system shown in Fig. \ref{vicon}.  

\begin{figure}[hbt!]
\centering
\includegraphics[width=0.48\textwidth]{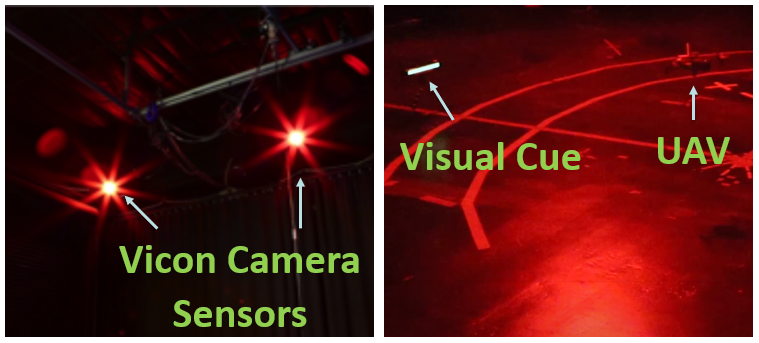}
\caption{Experimental Setup in Vicon System}
\label{vicon}
\end{figure}

Vicon is a motion capture system widely used in the entertainment industry to track the motion of people and other objects even up to sub-millimeter resolution. The Vicon system used for this study can capture images at 120Hz using eight 16-megapixel digital video cameras. Camera-mounted strobes illuminate small, retro-reflective markers, which are identified and processed by the imagers. The Vicon cameras track reflective markers on both the UAV and the visual cue to obtain precise position and orientation data, which is then used as the ground truth for this study to validate the computer vision system. The comparisons of the forward relative distance and sideward relative distance are shown in Fig. \ref{vicon_result_f} and \ref{vicon_result_s}, respectively.

\begin{figure}[!hbt]
  \ContinuedFloat*  \raggedleft
   \includegraphics[width=1.0\linewidth]{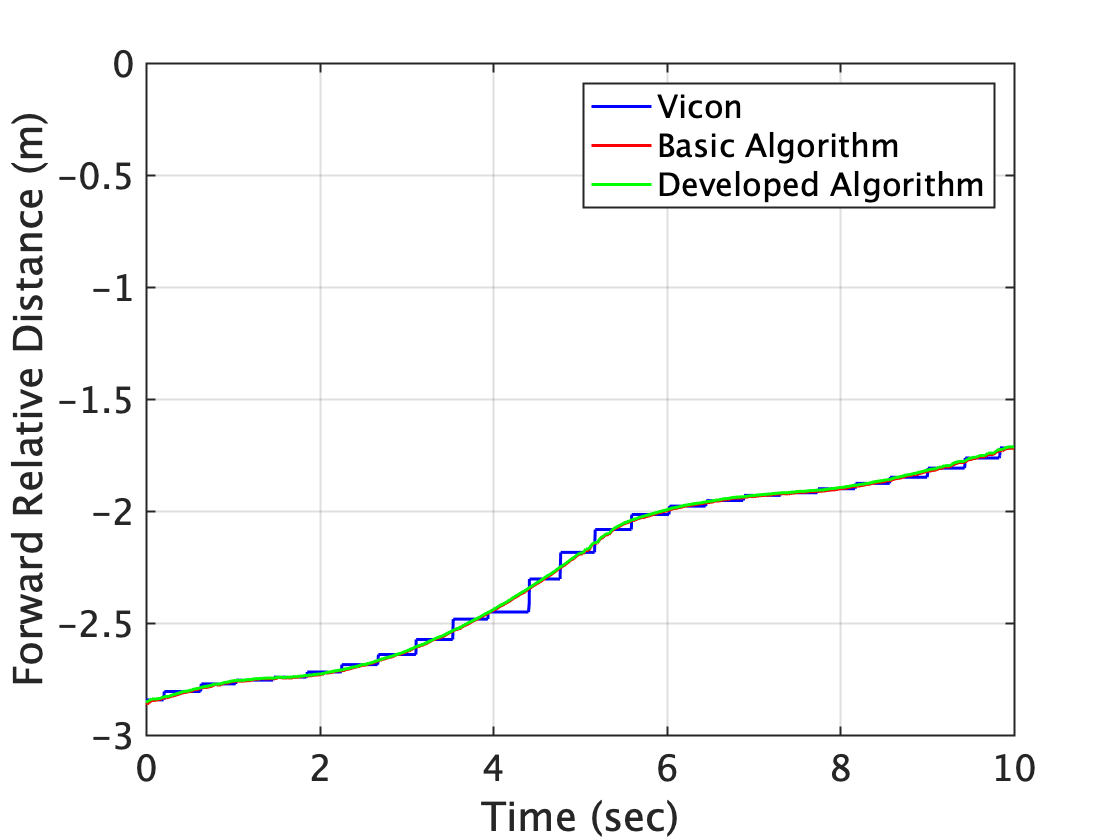}
  \caption{\label{vicon_result_f}Validation of Forward Distance Estimation} 
\end{figure}
\begin{figure}[!hbt]
  \ContinuedFloat  \raggedleft
   \includegraphics[width=1.0\linewidth]{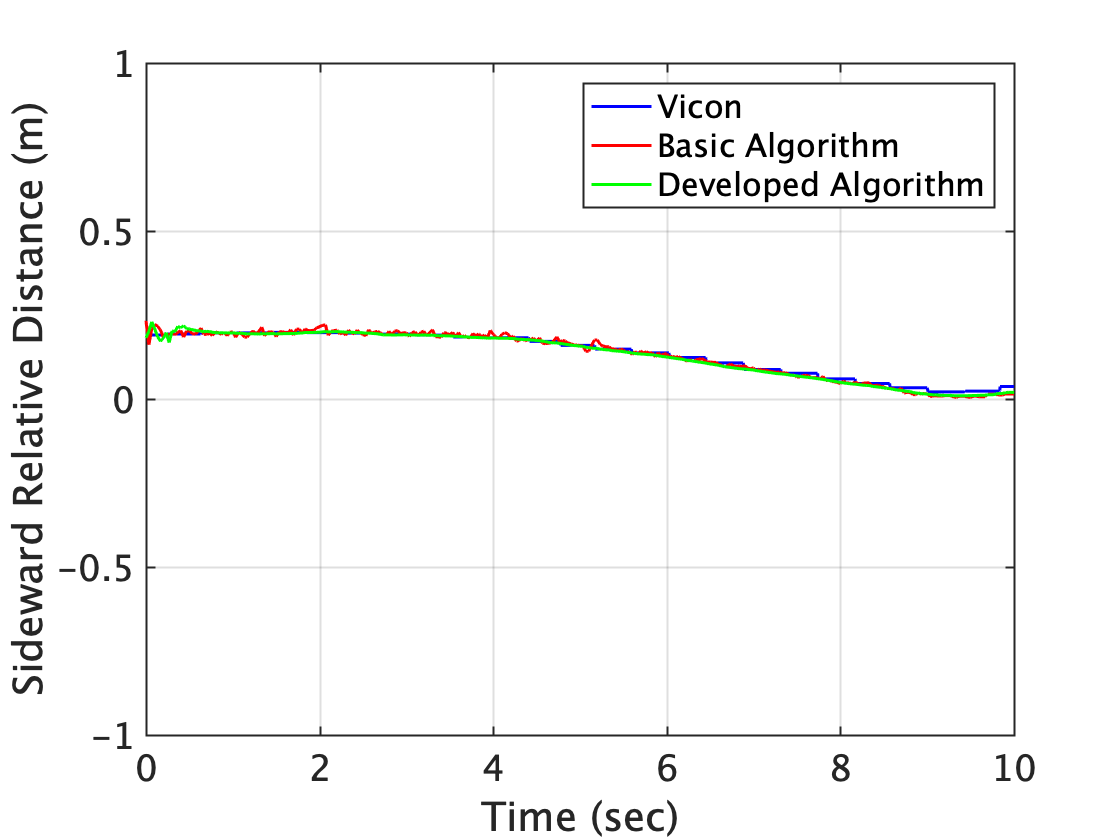}
  \caption{\label{vicon_result_s}Validation of Sideward Distance Estimation} 
\end{figure}

The blue, red, and green lines denote the Vicon measurements, baseline algorithm estimations, and estimations from the present improved algorithm, respectively. As seen from the figure, the baseline and the improved algorithms have no difference when it comes to the forward relative distance estimation, and both these methods can obtain the same level of accuracy as the Vicon measurements. The sideward distance estimation by the baseline algorithm has fluctuations; however, the improved algorithm shows smooth results due to the moving average filter. The maximum error in sideward distance estimation is 1 centimeter, when compared to the Vicon result. The comparisons of the vertical relative distance and relative yaw angle are as shown in Fig. \ref{vicon_result_v} and \ref{vicon_result_v}, respectively. 

\begin{figure}[!hbt]
  \ContinuedFloat*  \raggedleft
   \includegraphics[width=1.0\linewidth]{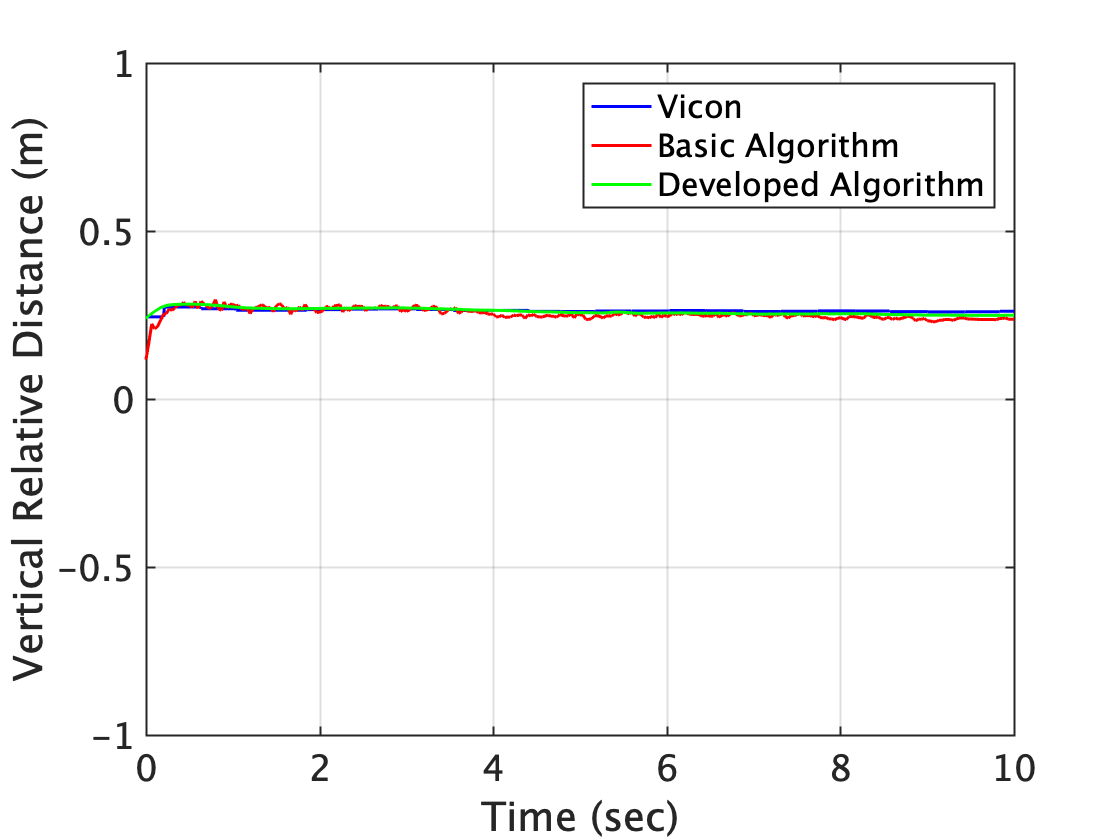}
  \caption{\label{vicon_result_v}Validation of Vertical Distance Estimation} 
\end{figure}
\begin{figure}[!hbt]
  \ContinuedFloat  \raggedleft
   \includegraphics[width=1.0\linewidth]{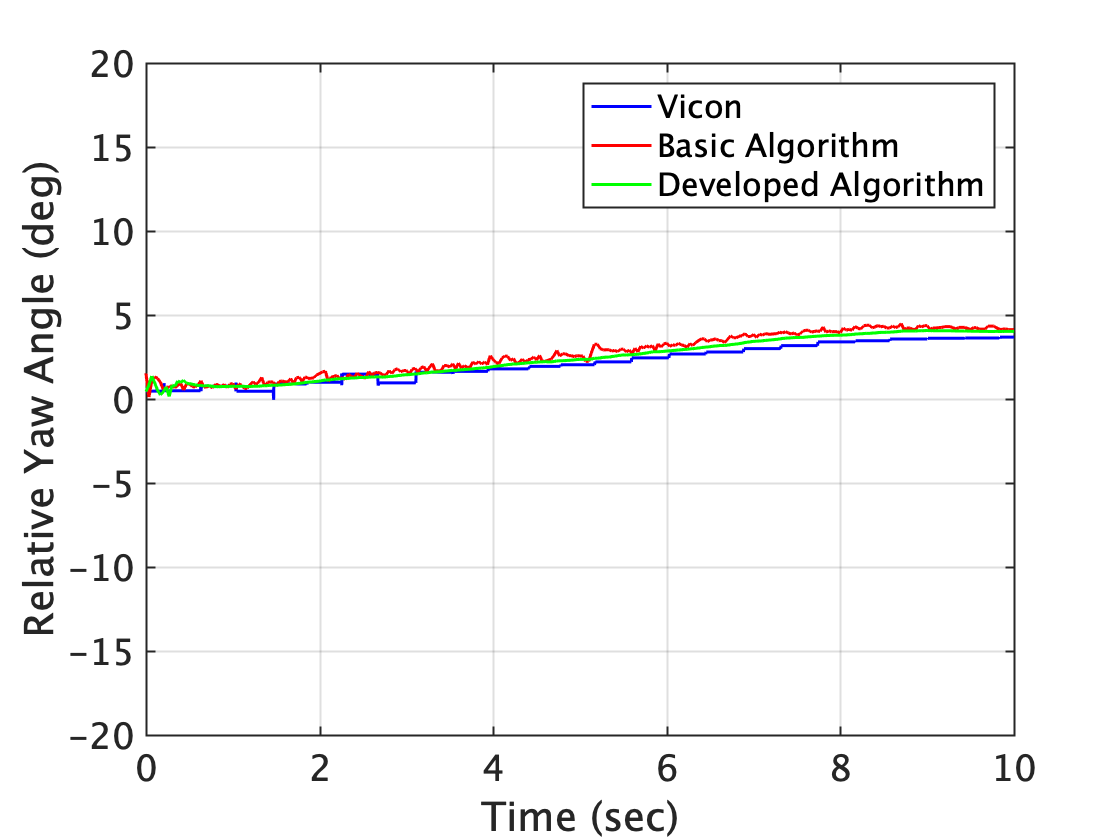}
  \caption{\label{vicon_result_y}Validation of Yaw Angle Estimation} 
\end{figure}

In both vertical relative distance and relative yaw angle comparison, the improved algorithm yields smooth estimation results. The maximum error in the vertical relative distance is below 1 centimeter and the maximum error in the relative yaw angle is 1 degree. Through the Vicon experiments, the ability of the present computer vision system to detect the visual cue and precisely estimate the position and orientation is demonstrated. 

\section{Flight Control System}

The objective is to develop a flight control system that will enable any type of VTOL aircraft to land precisely on moving platforms. The present vision-based controller will form the outer-loop of a cascaded feedback control system where it is assumed that the inner-loop will handle the basic attitude stability of the aircraft as well as generates the required pitch, roll, and yaw rates as demanded by the outer-loop. Therefore, the inner loop will be aircraft dependent. In this study, a quadcopter with fixed pitch propellers is selected and the inner-loop controls the rotating speed of each propeller. Proportional integral derivative (PID) controllers with scheduled gains are designed to achieve a realistic flight dynamic behavior as well as precise tracking and landing of the VTOL UAV. The standard form of the PID controller is represented in Eq.~\eqref{PID}. 

\begin{equation} \label{PID}
u(t) = K_{P}e(t) + K_{I}\int_{0}^{\tau} e(\tau) d\tau + K_{D}\frac{d}{dt}e(t)
\end{equation}

$e(t)$ denotes an error between a setpoint and relative position data with respect to the visual cue. Since the positions and heading angle are computed from the vision system, disturbances and noises are already included. The update rate of the outer loop of the closed-loop feedback control system is 0.1 seconds including the detection, estimation, and live-streaming through the computer vision system. $K_{P}$, $K_{I}$, and $K_{D}$ gains are tuned based on the effect of each parameter as shown in Table \ref{tuning} \cite{ang2005pid}.

\begin{table}[ht] \begin{minipage}{\columnwidth} \centering
\caption{Effect of $K_{P}$, $K_{I}$, and $K_{D}$}
\label{tuning}
\begin{tabular}{lccc} \arrayrulecolor{halfgreen} \hline \hline
Characteristic & Increase $K_{P}$ & Increase $K_{I}$ & Increase $K_{D}$ \\ \hline
\multirow{2}{*}{Rise Time} &  \multirow{2}{*}{Decrease} & \multicolumn{1}{c}{Small} & \multicolumn{1}{c}{Small}\\ 
                           &                            & \multicolumn{1}{c}{Decrease} & \multicolumn{1}{c}{Decrease}\\ 
\\
Overshoot & Increase & Increase & Decrease\\
\\

\multirow{2}{*}{Settling Time} &  \multicolumn{1}{c}{Small} & \multirow{2}{*}{Increase} & \multirow{2}{*}{Decrease}\\ 
                               &  \multicolumn{1}{c}{Increase} &                           & \\ 

\\
\multicolumn{1}{l}{Steady-state} &  \multirow{2}{*}{Decrease} & \multicolumn{1}{c}{Large} & \multicolumn{1}{c}{Minor}\\ 
\multicolumn{1}{l}{Error}  &                         & \multicolumn{1}{c}{Decrease} &\multicolumn{1}{c}{Change}\\ 

\\
Stability & Degrade & Degrade & Improve\\ \hline \hline
\end{tabular} \end{minipage}
\end{table}

\begin{figure*}[b]
\centering
\includegraphics[width=0.9\textwidth]{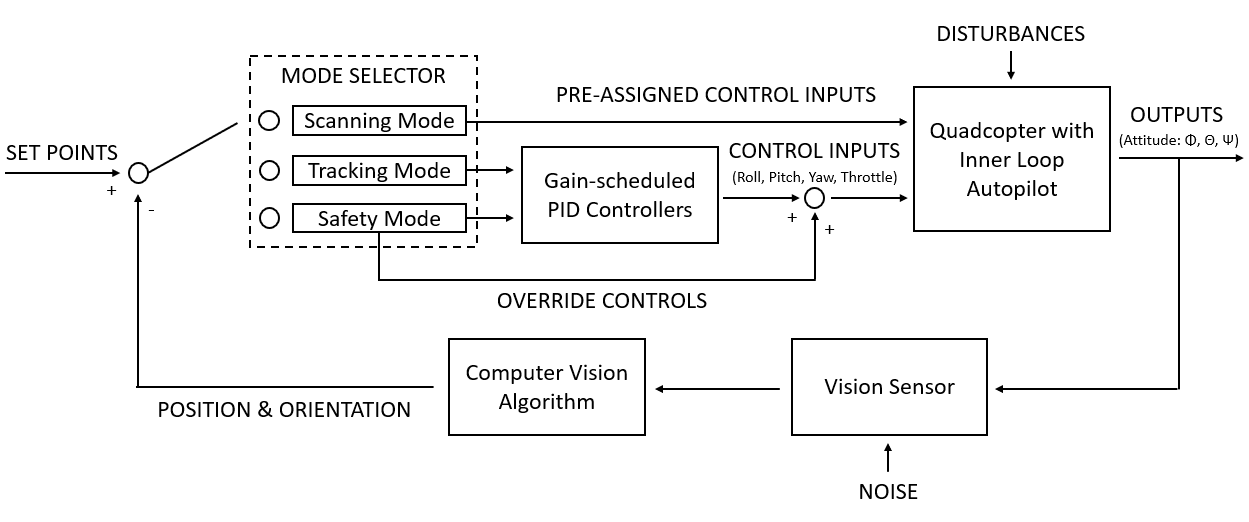}
\caption{Vision-based Control System Architecture}
\label{architecture}
\end{figure*}

The different sets of gains are scheduled based on relative positioning data to control the behavior of a UAV as desired. It allows the UAV to fly at high-speed when it is relatively far away from the visual cue and then slow down the relative speed for cautious tracking when it gets closer to the landing point. The flight mode selector is also configured in order to adapt the UAV movements to the flight conditions and enhance safety. The states are the relative positions in three dimensions and the heading angle of the UAV with respect to the visual cue obtained from the computer vision system. The control inputs are pitch, roll, yaw, and throttle commands. Each control magnitude ranges from -100 to 100 as a percentage of the total input. The outputs are UAV attitude and raw images from the camera. The UAV attitude consists of roll angle $\phi$, pitch angle $\theta$, and yaw angle $\psi$.

\subsection{Control Architecture}

The vision-based PID control system with scheduled gains is structured as shown in Fig. \ref{architecture}. Setpoints are the desired relative positions and heading angle of the UAV. The mode selector automatically chooses a proper flight mode based on the flight conditions. Depending on the selected flight mode, the control inputs are obtained through the gain-scheduled PID controllers or from the pre-assigned values. The parameters that are fed back include the estimated UAV position and heading angle obtained from the computer vision system.

\subsection{Flight Modes}

A robust flight control system should be able to operate under various flight conditions. Thus, flight modes are configured to fly the UAV in an appropriate manner given the scenario. They consist of the mode selector and gain-scheduled PID controllers in order to obtain the proper control inputs depending on the scenario. Three different modes are developed to enhance safety and tracking capability as follows.

\begin{enumerate}

\item \textbf{Scanning mode:} In the case of a visual cue detection failure, the scanning mode is activated anytime during the flight to search for the visual cue. Once this mode is engaged, the UAV starts yawing at a constant rate by commanding the pre-assigned control inputs instead of relying on the gain-scheduled PID controllers. It continues to yaw until it detects the visual cue and one successful detection disengages this mode instantly. This allows the UAV to track the visual cue irrespective of its initial heading orientation as long as the visual cue is within the detection range. During the flight, if the UAV loses the visual cue from the camera view, it stops moving and then scans for the cue in order to get back on track. Hence, the scanning mode enhances safety as well as increases operational range and robustness.

\item \textbf{Tracking mode:} This mode is activated once the visual cue is properly detected. Based on the relative distance and heading angle, different gain-scheduled PID controllers are engaged as shown in Fig. \ref{scheduled_gain}. 

\begin{figure*}[b]
\centering
\includegraphics[width=0.85\textwidth]{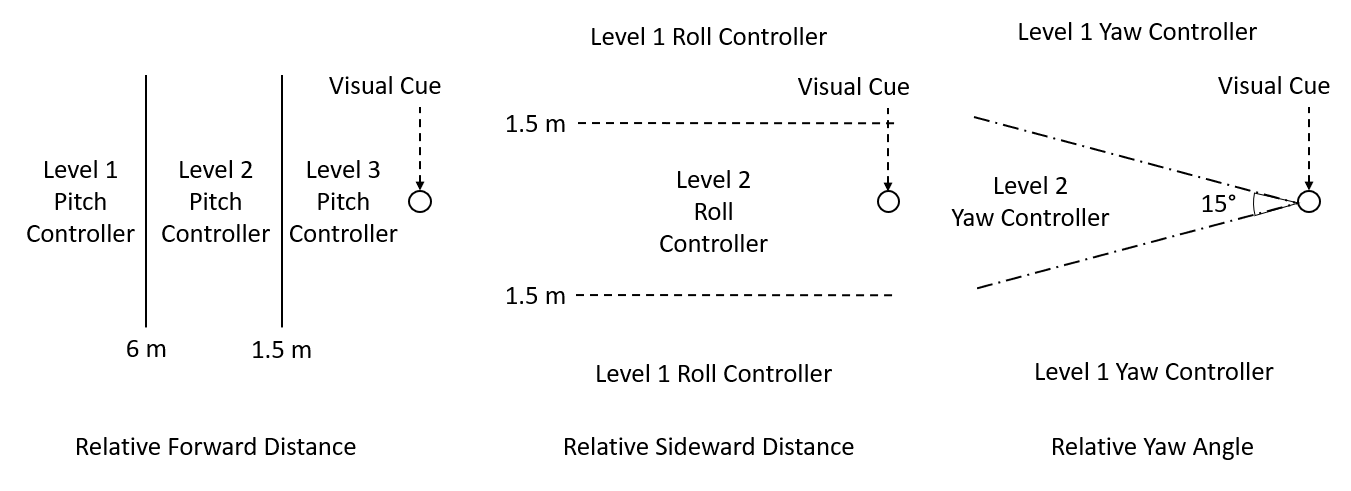}
\caption{Composition of Gain-scheduled PID Controllers}
\label{scheduled_gain}
\end{figure*}

First, three different pitch controllers are configured based on the forward relative distance. Level 1 pitch PID controller gains are scheduled to track the visual cue at the maximum speed when the distance to the landing spot is more than 6 meters. Level 2 pitch PID controller gains are scheduled to track the visual cue at moderate speeds when the distance range is between 6 and 1.5 meters. The controller is tuned to prevent overshoots and oscillations. Level 3 pitch PID controller gains are scheduled to track the visual cue within a range of 1.5 meters. It focuses on minimizing steady-state error for precise tracking and mimics the cautious final approach maneuver of a helicopter.
Second, two different roll controllers are configured based on the sideward relative distance. Level 1 roll controller gains are scheduled to track the visual cue when the sideward distance is more than 1.5 meters and is tuned to achieve fast-tracking, no overshoots, and no oscillations. Level 2 roll controller gains are scheduled to track the visual cue precisely within the range of 1.5 meters focusing on steady-state error minimization.
Third, two different yaw controllers are configured based on the relative heading (yaw) angle to maintain the desired approach angle. In order to prevent the visual cue from getting out of the camera view by commanding large pitch and roll motions with excessive yaw angle, it gives priority to the yaw correction by reducing the magnitude of pitch and roll control inputs. Level 1 yaw controller is designed to slow down the pitch and roll motion as well as correct the yaw angle quickly when the relative yaw angle is more than 15 degrees. Level 2 yaw controller is designed to correct the yaw angle precisely without slowing down the pitch and roll motion when the relative yaw angle is less than 15 degrees.
Lastly, one throttle PID controller is configured to control the heave motion based on the vertical relative distance. Considering the vertical displacement that the UAV can have within the camera view, one set of PID gains is enough to handle its heave motion. 

\item \textbf{Safety mode:} This mode is activated to enhance safety by responding appropriately to three different situations shown in Fig. \ref{safety_mode} and the safety mode supersedes any other mode. 

\begin{figure*}[t!]
\centering
\includegraphics[width=0.9\textwidth]{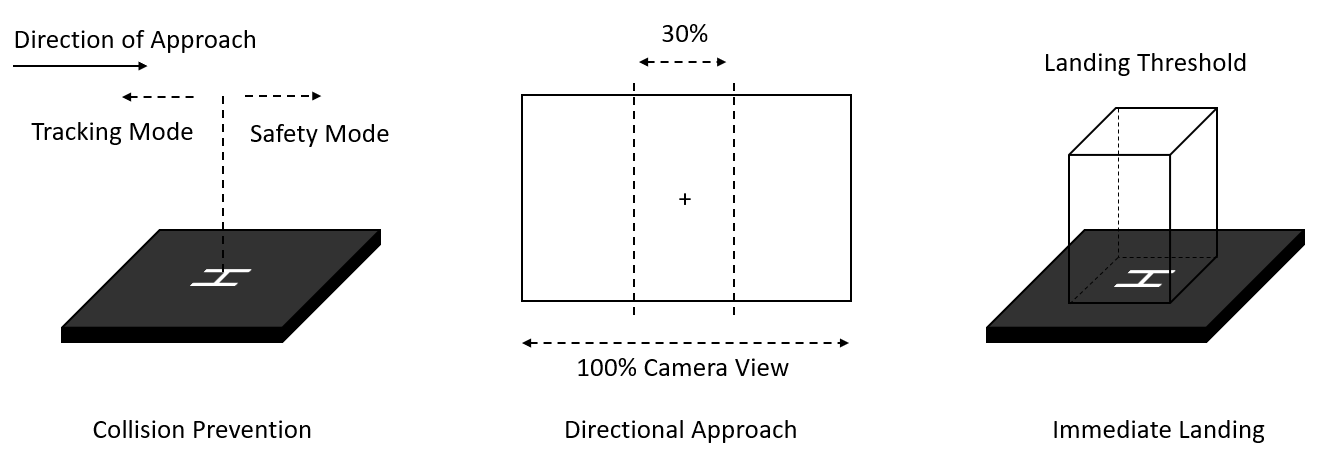}
\caption{Schematic of Safety Mode}
\label{safety_mode}
\end{figure*}

First, a backward (nose-up) pitch is applied instantly to prevent a collision when crossing the landing pad centerline instead of relying on the pitch PID controller. Second, the directional approach is designed to control the UAV when estimated relative positions are not reliable. Particularly, there exists a distance range where the forward distance estimation is accurate enough but the sideward distance and relative yaw angle estimation are not accurate due to a small number of pixels. In this case, it directly uses pixel position data in a 2D camera plane to control roll motion instead of estimating a position from an image. It does not correct for yaw angle but the roll is commanded if the visual cue is located in the area which is more than 15 percent offset from the center of the image in order to make sure the visual cue is not lost from the camera view. Third, an immediate landing is attempted to reduce the time for the final vertical movement by commanding throttle down instantly when the UAV enters into the landing threshold.

\end{enumerate}

\section{Simulations}

The objective is to verify every aspect of the vision-based flight control system under different scenarios and to demonstrate that extreme accuracy can be achieved by adopting this novel method of tracking a visual cue installed parallel to a UAV approach course instead of tracking a landing platform.

\begin{figure*}[b!]
\centering
\includegraphics[width=1.0\textwidth]{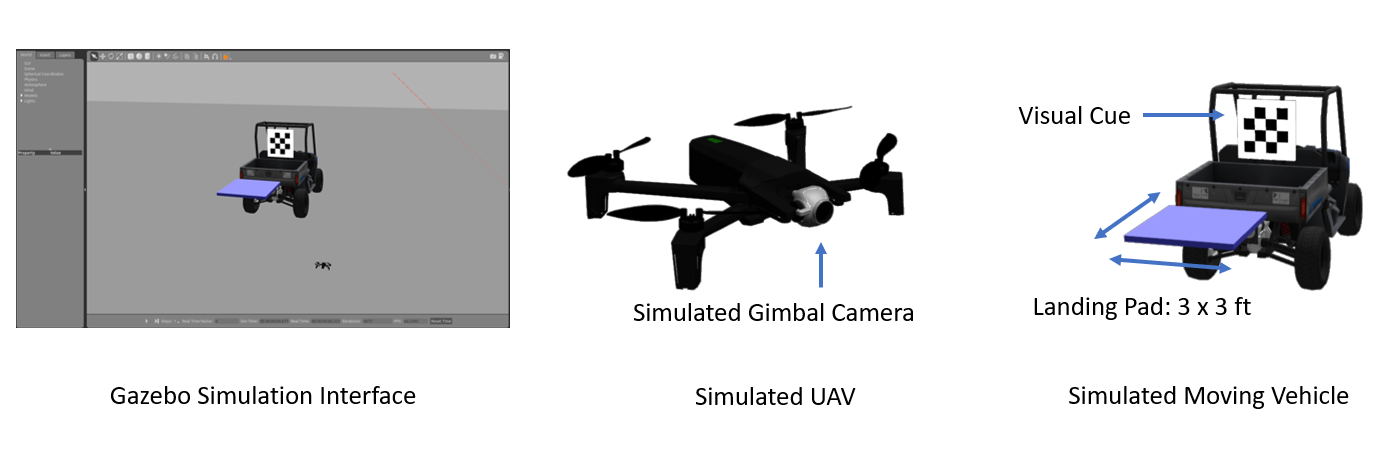}
\caption{Simulated UAV and Moving Platform in Gazebo Simulation Program}
\label{gazebo_total} 
\end{figure*}

In the present study, the entire coding is done in the Python programming language and a realistic robot simulation program, Gazebo \cite{gazebo}, is used to simulate the physical and visual surroundings of a UAV as shown in Fig. \ref{gazebo_total} and \href{https://youtu.be/8kAeVzGJyN8}{Video} \cite{simvideo}. 

The simulated UAV model has the same dimensions, mass moments of inertia, gimbal camera, and flight mechanism as the physical UAV, Parrot Anafi. The simulation also implements a moving vehicle that carries the visual cue and landing platform as shown in Fig. \ref{gazebo_total}. As determined by vision system experiments, the visual cue has a checkerboard pattern consists of 4 x 4 squares with a side length of 120 mm each, and the landing platform size is 3 x 3 ft. These characteristics remain the same throughout simulations and flight tests.

The simulations are conducted in the same way as the actual flight tests including live-streaming, real-time vision processing, and feedback control loop. First, the simulated UAV is connected via WIFI to the base station computer. Once the UAV takes off, the images from the simulated camera are streamed to the computer. The images are processed one by one to obtain the relative position and heading information, which is utilized by the feedback controller running on the computer to generate the control inputs. Then, the computed control inputs are sent back to the UAV via the WIFI connection. This one cycle from streaming to sending commands back to the UAV takes 0.1 seconds. This simulated system, once verified, can be directly applied to a real flight test by simply switching an IP address from simulated UAV to physical UAV. The following simulation results show the tracking capability and landing accuracy in detail.

\subsection{Landing Accuracy}

\begin{figure*}[t]
\centering
\includegraphics[width=1.0\textwidth]{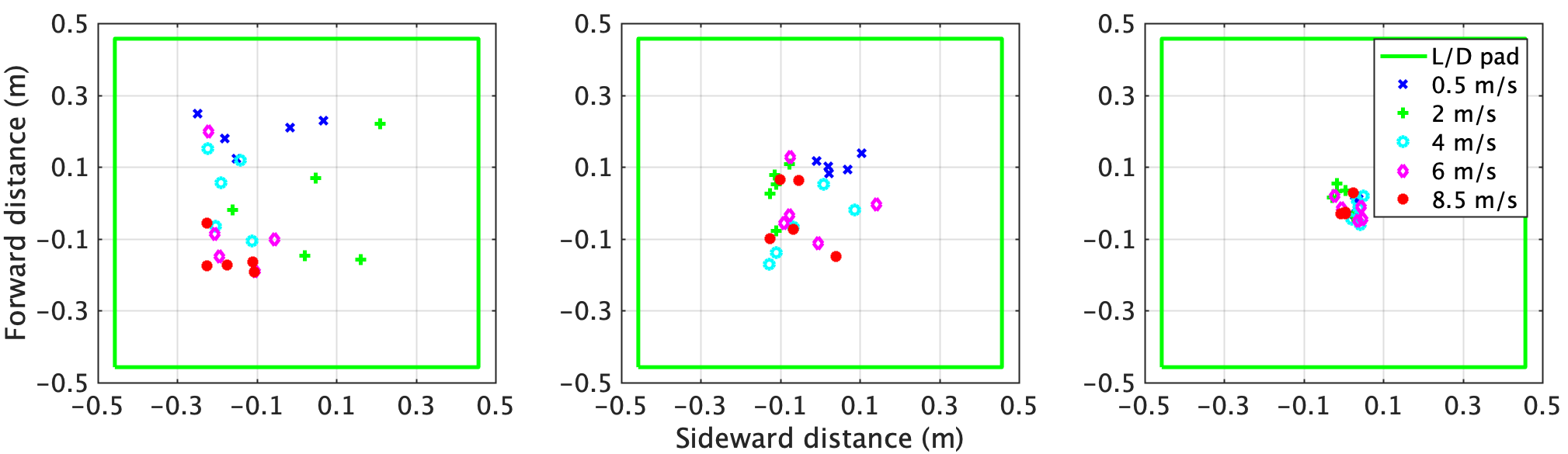}
\caption{Landing Points with Landing Threshold of 25 x 25cm(left), 15 x 15cm(center), and 5 x 5 cm(right)}
\label{landing_accuracy} 
\end{figure*}

To demonstrate the tracking and landing accuracy of the present vertical-visual-cue based method, multiple simulations are conducted with different landing thresholds and varying the speed of the mov
ing platform. The final landing locations from 100 independent simulations are plotted in Fig. \ref{landing_accuracy}. 

The green line depicts the landing pad and each mark denotes the landing point for different speeds of the moving platform. Note that landing anywhere within the 25 x 25 cm area can be regarded as a safe landing considering the UAV size. The present vision-based control system achieves precise landing within a 5 x 5 cm landing threshold for all platform speeds. When the platform is moving at 8.5 m/s, it requires the UAV to fly close to its maximum speed but it still can make accurate landings. The dispersion of landing points in the figure can be attributed to the assigned landing threshold. The time history of relative distance for each landing case is shown in Fig. \ref{landing_time}.

\begin{figure*}[b]
\centering
\includegraphics[width=1.0\textwidth]{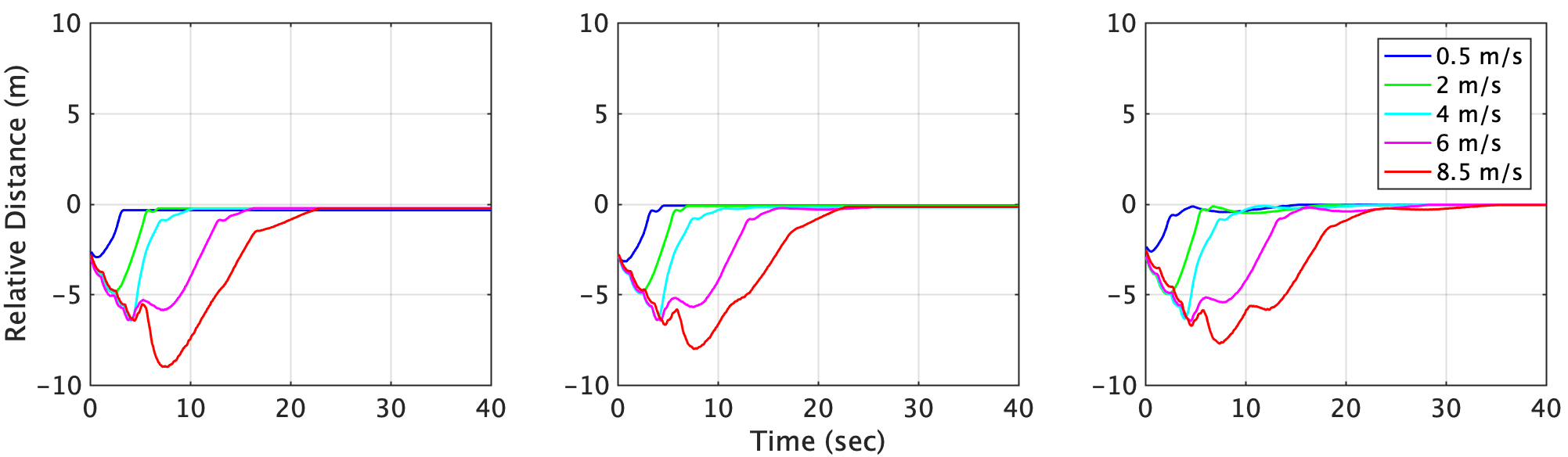}
\caption{Relative Distance with Landing Threshold of 25 x 25cm(left), 15 x 15cm(center), and 5 x 5 cm(right)}
\label{landing_time} 
\end{figure*}

Each line denotes the relative distance between the UAV and landing pad center at different platform speeds. By the time the UAV starts tracking after take-off, the platform is already accelerating forward up to its designated speed. Thus, the relative distance increases at the beginning of the tracking phase. Also, it takes more time to land as the landing threshold is made smaller. The average time to achieve a landing threshold in each case is specified in Table \ref{landing_time_table}.

\begin{table}[ht] \begin{minipage}{\columnwidth} \centering
\caption{Average Time to Achieve Landing Threshold}
\label{landing_time_table}
\begin{tabular}{lccc} \arrayrulecolor{halfgreen} \hline \hline
Vehicle Speed & 5x5cm & 15x15cm & 25x25cm\\ \hline
0.5m/s & 15.3s & 4.6s & 3.3s \\
2.0m/s & 25.6s & 7.0s & 6.8s\\
4.0m/s & 26.9s & 13.8s & 10.3s \\
6.0m/s & 28.6s & 16.4s & 16.2s \\
8.5m/s & 35.2s & 23.0s & 22.9s\\ \hline \hline
\end{tabular} \end{minipage}
\end{table}

In conclusion, the landing accuracy depends on an assigned landing threshold. The present system can achieve a precise landing, however, there is a trade-off between the time and landing accuracy. Thus, the landing threshold should be selected based on the priority of a mission, which could be accuracy or time. In the following simulations, the medium landing threshold of 15 x 15 cm is applied.

\subsection{Tracking Capability}

To examine the tracking capability, simulations are conducted by varying the speed and the course of the moving platform. The four cases discussed in this paper include the platform moving forward in a straight line at speeds 0.5 m/s and 8.5 m/s as well as the platform moving in an S-pattern and a circular pattern. The trajectories, final landing points, set-point tracking data in time, and time history of control inputs are analyzed for each case.

\begin{enumerate}

\item \textbf{Case-1: Platform moving in a straight line at 0.5 m/s:} The platform is programmed to move forward at 0.5 m/s speed and the trajectories viewed from the top and side are shown in Figs. \ref{top_view_0} and \ref{side_view_0}, respectively. 

\begin{figure}[!hbt]
  \ContinuedFloat*  \raggedleft
   \includegraphics[width=0.9\linewidth]{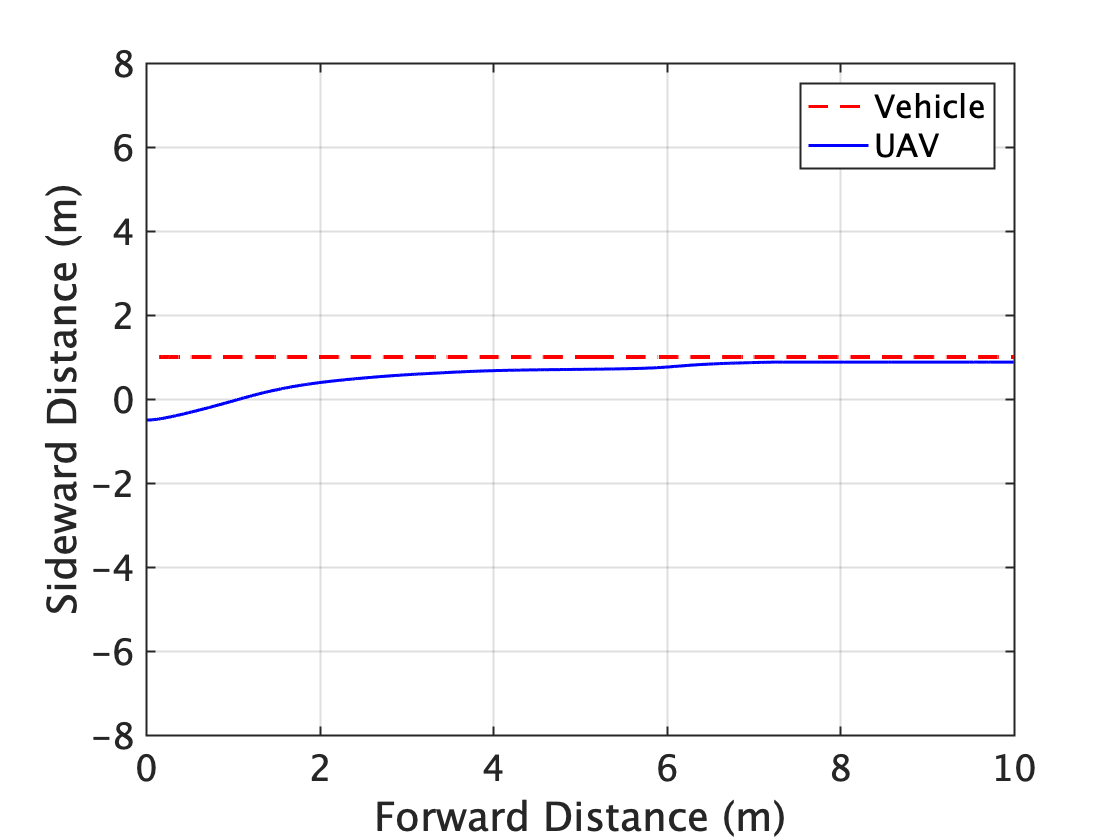}
  \caption{\label{top_view_0}Trajectories in Top View (Case-1)} 
\end{figure}
\begin{figure}[!hbt]
  \ContinuedFloat  \raggedleft
   \includegraphics[width=0.9\linewidth]{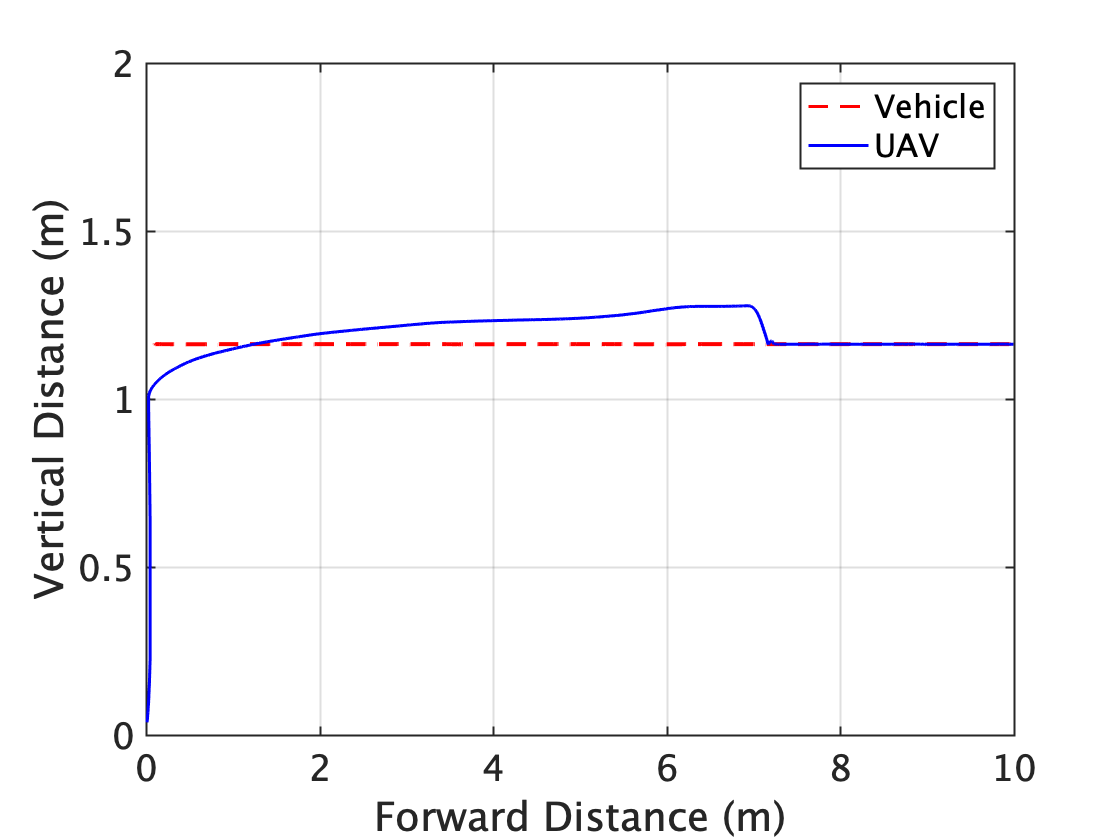}
  \caption{\label{side_view_0}Trajectories in Side View (Case-1)} 
\end{figure}

The solid blue line denotes the UAV trajectory whereas the dashed red line denotes the platform trajectory from take-off to landing. The initial position of the UAV is on the ground at a sideward distance of 1.5 meters from the platform. The UAV takes off and then tracks the visual cue until satisfying the landing condition. Once the UAV enters the 15 x 15 cm landing space from the pad center, it lands quickly.

The forward relative distance and pitch command input in time are shown in Figs. \ref{forward_0} and \ref{pitch_0}, respectively. The blue and red solid lines denote the time history of the forward relative distance and pitch command input, respectively. The pitch command input ranges from -100 to 100 as percentage input where -100 is the maximum pitch backward and 100 is the maximum pitch forward. Initially, the distance increases since the UAV has to pick up speed from 0 m/s speed while the platform is already moving forward at 0.5 m/s. Due to the different levels of gain-scheduled PID controllers activated based on the relative distance, the tracking rate changes from fast-tracking at a large distance to precise tracking in the proximity of the landing pad. It takes 4.4 seconds to reach the landing threshold of 15 cm from the landing pad center and the final landing deviation is 2 cm.

\begin{figure}[!hbt]
  \ContinuedFloat*  \raggedleft
   \includegraphics[width=0.9\linewidth]{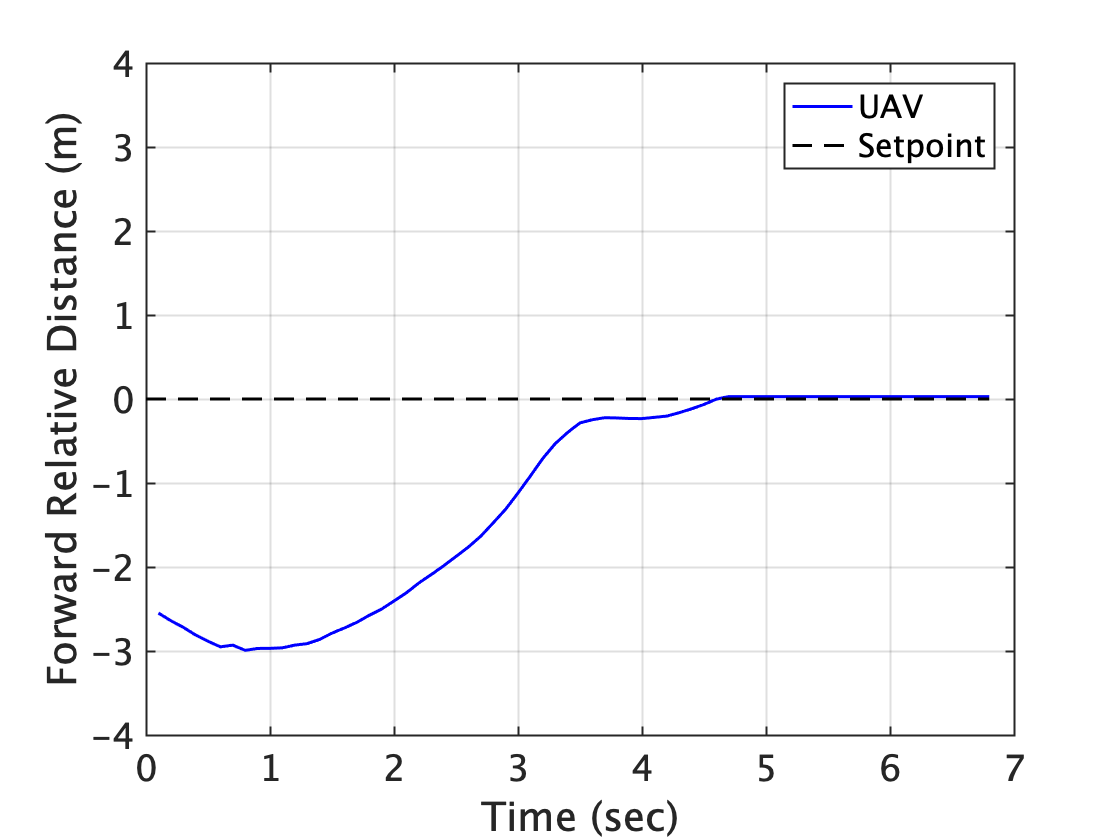}
  \caption{\label{forward_0}Forward Relative Distance (Case-1)} 
\end{figure}
\begin{figure}[!hbt]
  \ContinuedFloat  \raggedleft
   \includegraphics[width=0.9\linewidth]{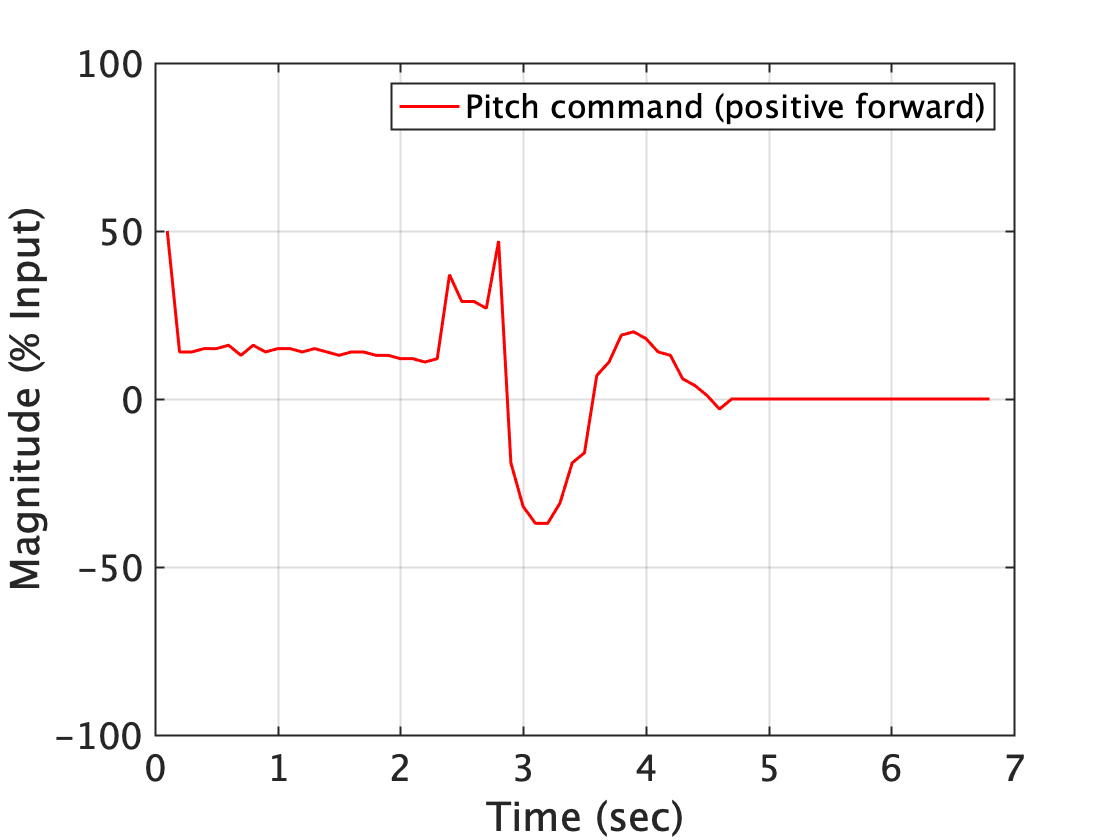}
  \caption{\label{pitch_0}Pitch Command Input (Case-1)} 
\end{figure}

The sideward relative distance and roll command input in time are shown in Figs. \ref{side_0} and \ref{roll_0}, respectively. The blue and red solid lines denote the time history of sideward relative distance and roll command input, respectively. The roll command input ranges from -100 to 100 as percentage input where -100 is maximum roll left and 100 is maximum roll right. For an initial 1 second, \hbox{the roll is commanded based on the visual cue position in}

\begin{figure}[b]
  \ContinuedFloat*  \raggedleft
   \includegraphics[width=0.9\linewidth]{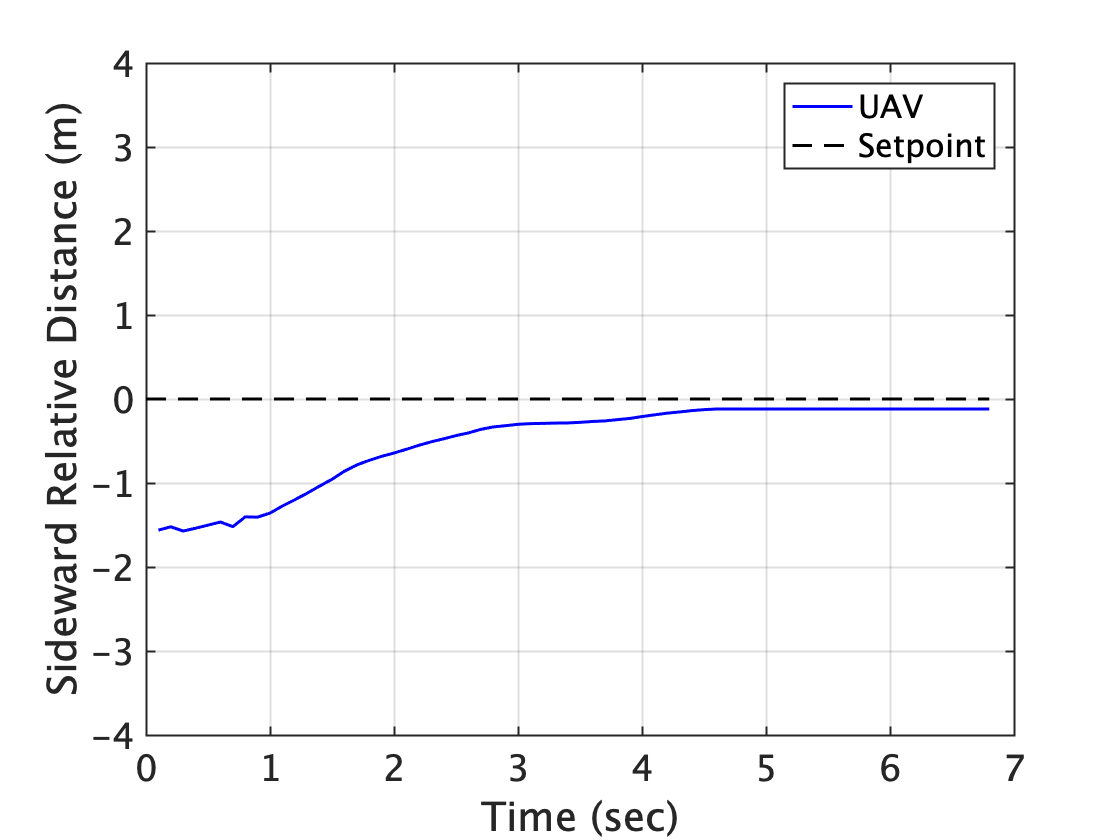}
  \caption{\label{side_0}Sideward Relative Distance (Case-1)} 
\end{figure}
\begin{figure}[!hbt]
  \ContinuedFloat  \raggedleft
   \includegraphics[width=0.9\linewidth]{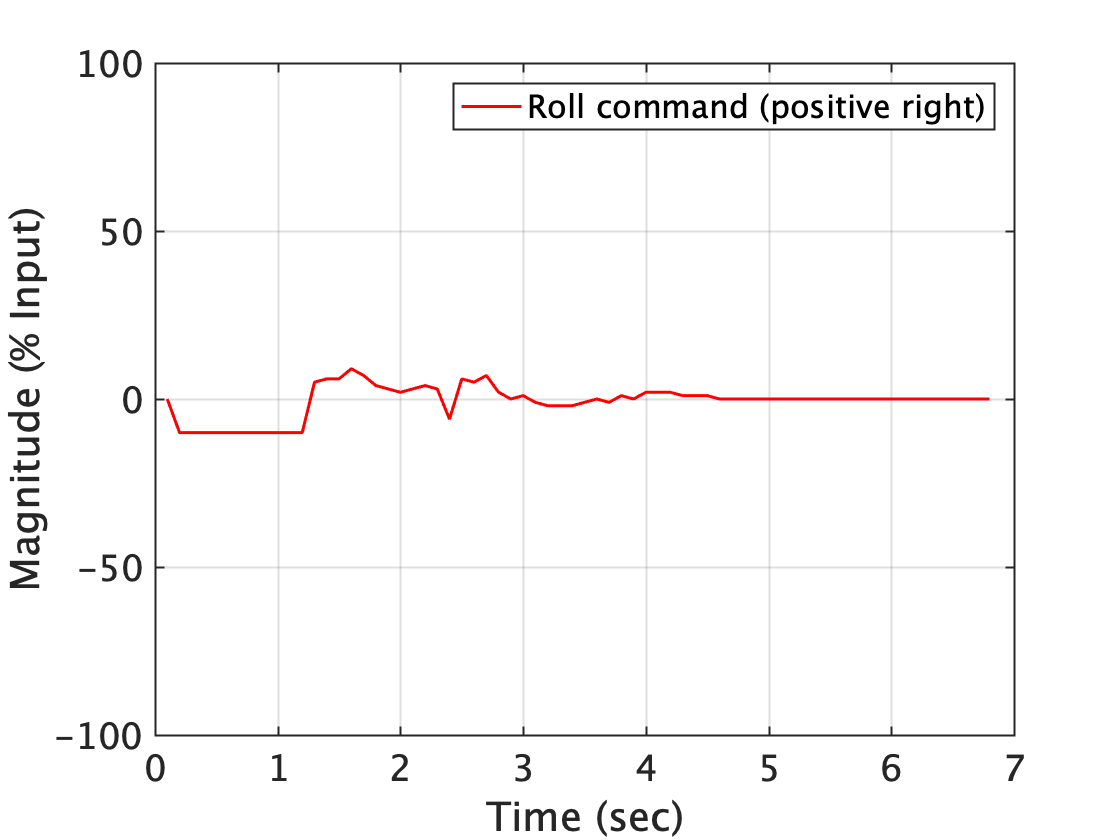}
  \caption{\label{roll_0}Roll Command Input (Case-1)} 
\end{figure}

a 2D camera view, which is the directional approach of safety mode. It prevents responding to the noisy sideward distance estimates that happen while the moving average method is not activated. It takes 10 previous data (1 second) to activate the moving average method. The final landing deviation in the sideward direction is 11 cm. 

The vertical relative distance and throttle command input in time are shown in Figs. \ref{vertical_0} and \ref{throttle_0}, respectively.

\begin{figure}[!hbt]

  \ContinuedFloat* \raggedleft
   \includegraphics[width=0.9\linewidth]{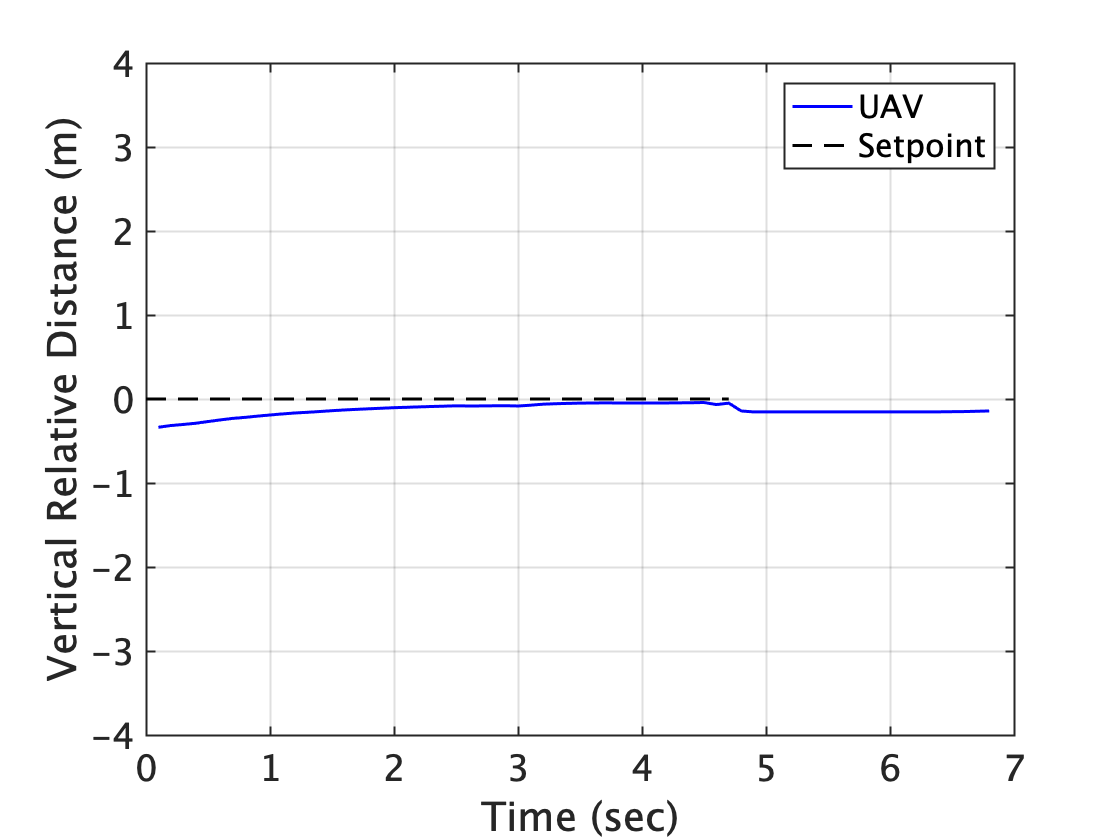}
  \caption{\label{vertical_0}Vertical Relative Distance (Case-1)} 
\end{figure}
\begin{figure}[b]
  \vspace{-0.4cm}
  \ContinuedFloat \raggedleft
   \includegraphics[width=0.9\linewidth]{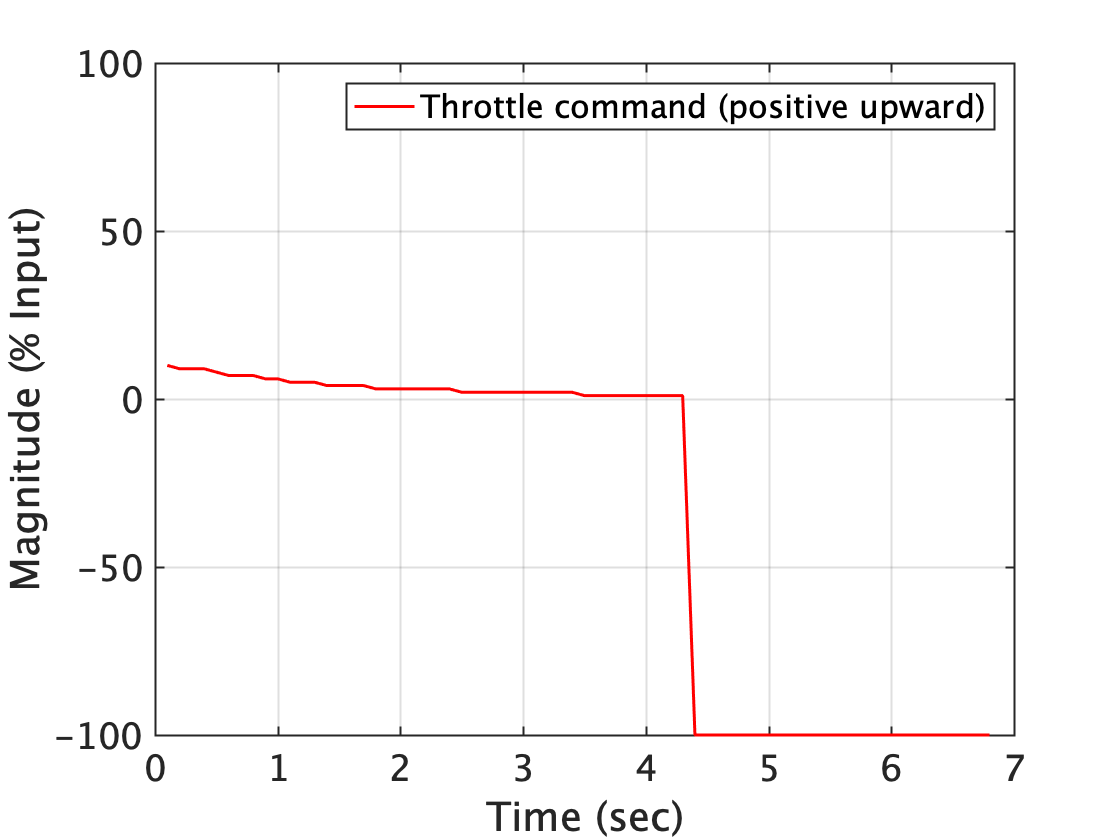}
  \caption{\label{throttle_0}Throttle Command Input (Case-1)} 
\end{figure}

The blue and red solid lines denote the time history of vertical relative distance and throttle command input, respectively. The zero setpoint means 15 cm above the landing pad. The throttle command input ranges from -100 to 100 as percentage input where -100 is maximum throttle down and 100 is maximum throttle up. At 4.4 seconds, the immediate landing is conducted by applying the maximum throttle down command.

The relative yaw angle and yaw command input in time are shown in Figs. \ref{yaw_0} and \ref{pedal_0}, respectively. The blue and red solid lines denote the time history of relative yaw angle and yaw command input, respectively. The zero setpoint means that a heading angle is set to zero for making a parallel approach. The yaw command input ranges from -100 to 100 as percentage input where -100 is the maximum yaw left and 100 is the maximum yaw right. As in the case of the roll, it takes one second for the moving average method to be active. Therefore, instead of responding to initial fluctuating yaw angle estimations, the directional approach of safety mode commands zero yaw for the first one second. Then, it starts regulating the yaw angle by the yaw PID controller once the moving average method is active.

\begin{figure}[!hbt]
  \vspace{0.4cm}
  \ContinuedFloat* \raggedleft
   \includegraphics[width=0.9\linewidth]{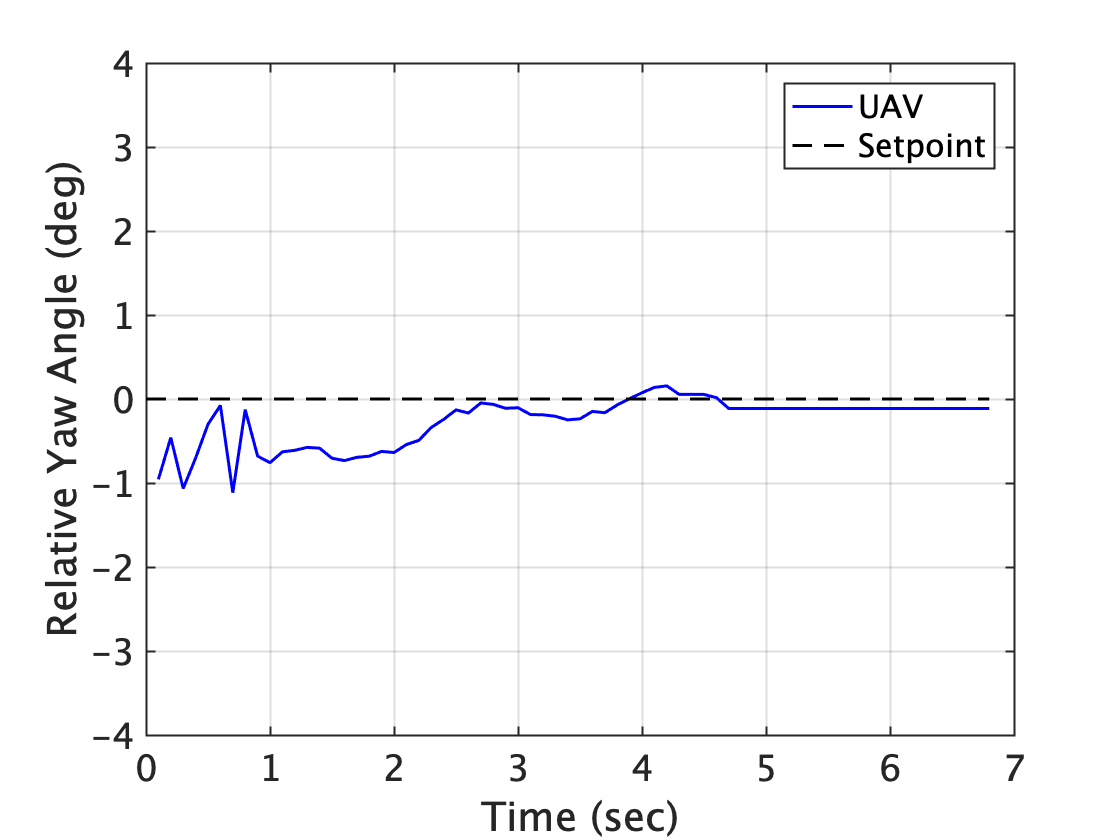}
  \caption{\label{yaw_0}Relative Yaw Angle (Case-1)} 
\end{figure}
\begin{figure}[b]
  \vspace{-0.4cm}
  \ContinuedFloat \raggedleft
   \includegraphics[width=0.9\linewidth]{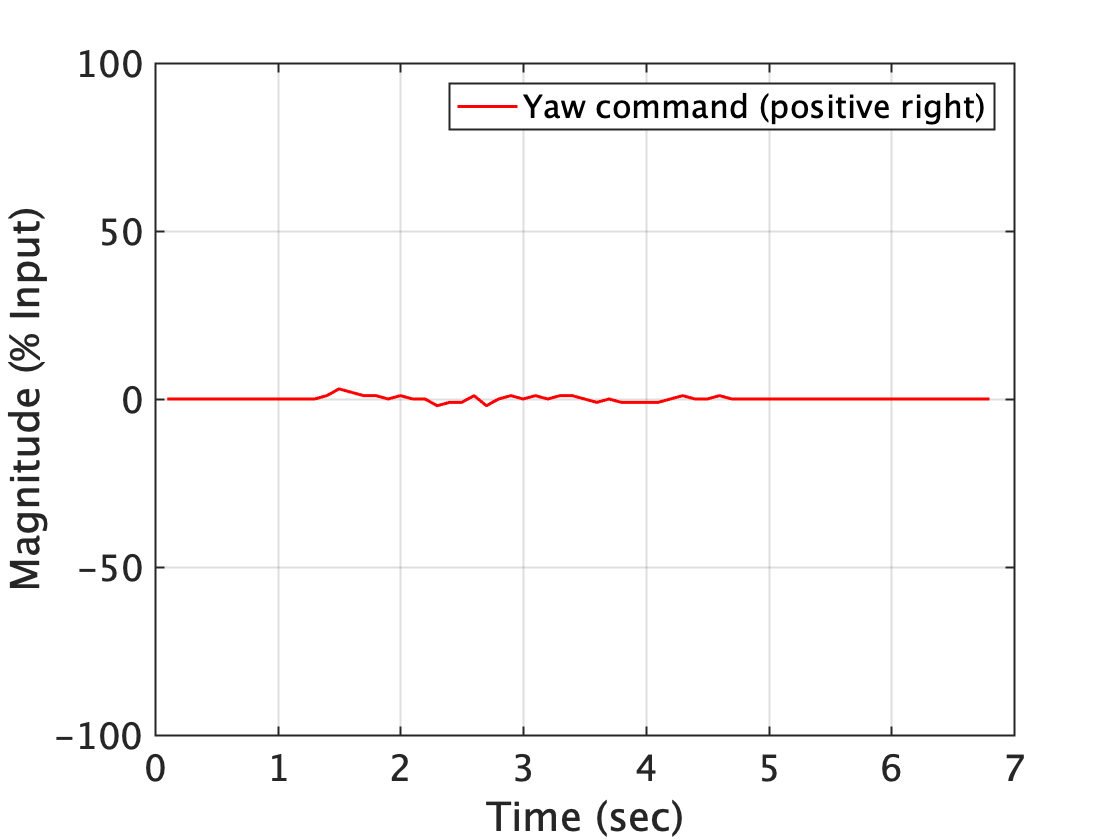}
  \caption{\label{pedal_0}Yaw Command Input (Case-1)} 
\end{figure}

\item \textbf{Case-2: Platform moving in a straight line at 8.5 m/s:} The platform is programmed to move forward at 8.5 m/s, which requires the UAV to fly at its maximum speed during tracking. Therefore, it occasionally experiences control saturation; however, it still makes a successful approach and landing. The trajectories viewed from the top and side are shown in Figs. \ref{top_view_8} and \ref{side_view_8}, respectively. The solid blue line denotes the UAV trajectory whereas the dashed red line denotes the platform trajectory from take-off to landing. The overshoots and undershoots observed in the trajectories are due to the control saturation when the UAV flies with its maximum pitch forward command which limits other commands. 

\begin{figure}[!hbt]
  \ContinuedFloat* \raggedleft
   \includegraphics[width=0.9\linewidth]{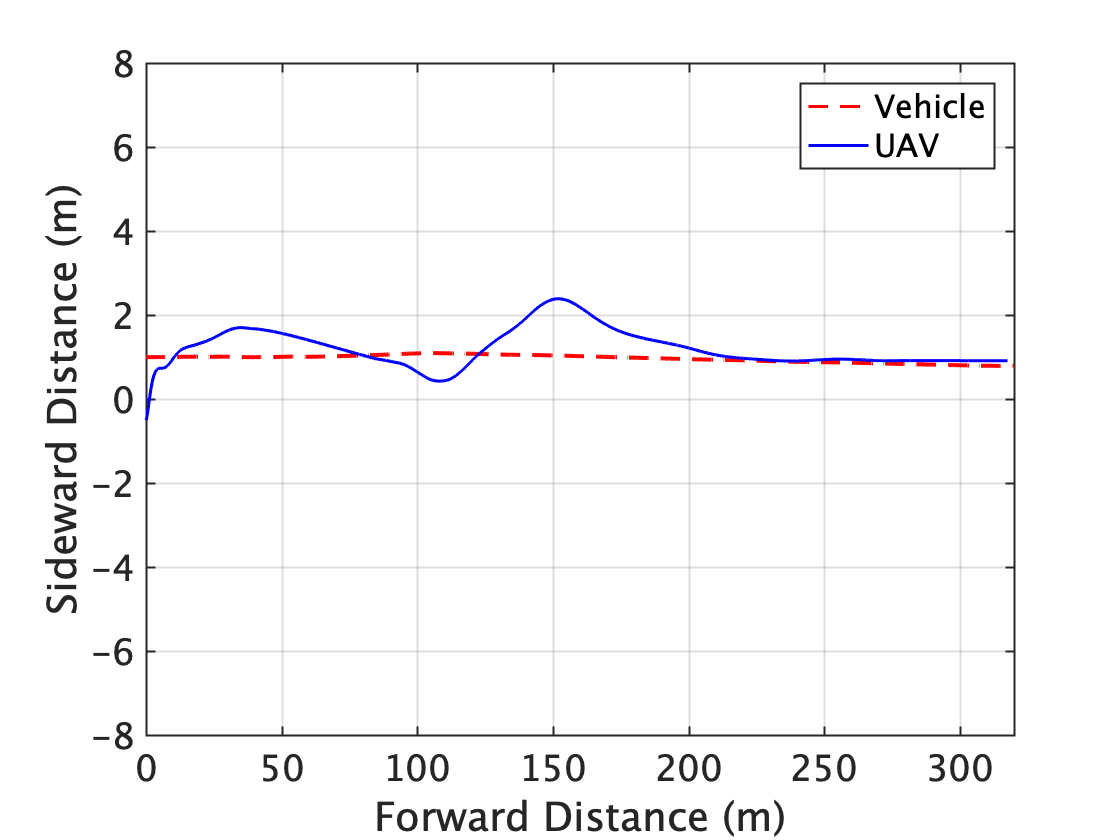}
  \caption{\label{top_view_8}Trajectories in Top View (Case-2)} 
\end{figure}
\begin{figure}[!hbt]
\vspace{-0.375cm}
  \ContinuedFloat \raggedleft
   \includegraphics[width=0.9\linewidth]{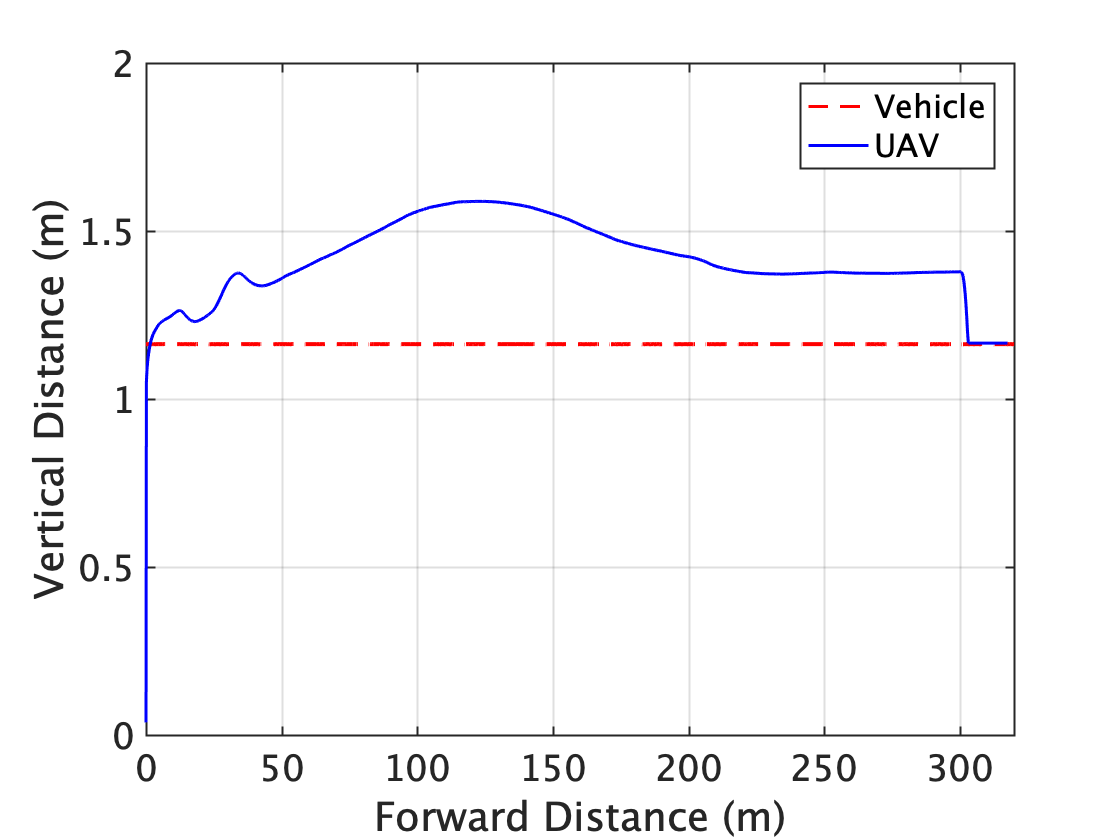}
  \caption{\label{side_view_8}Trajectories in Side View (Case-2)} 
\end{figure}

The forward relative distance and pitch command input in time are shown in Figs. \ref{forward_8} and \ref{pitch_8}, respectively. Up to 7.5 seconds, the relative distance increases since the platform is moving faster than the UAV. When the relative distance is further than 6 meters, the level 1 pitch controller yields maximum forward pitch for high-speed tracking. Since UAV control inputs are achieved through propeller speed variation, the maximum forward pitch that requires the highest rotating speed for the two rear propellers affects the capability to yield the other control inputs as desired. Level 2 and 3 pitch controllers are activated at distances of 6 and 1.5 meters, respectively. It takes 23 seconds to reach the landing threshold of 15 cm from the pad center and the final landing deviation in the forward direction is 12 cm. 

\begin{figure}[!hbt]
  \ContinuedFloat* \raggedleft
   \includegraphics[width=0.9\linewidth]{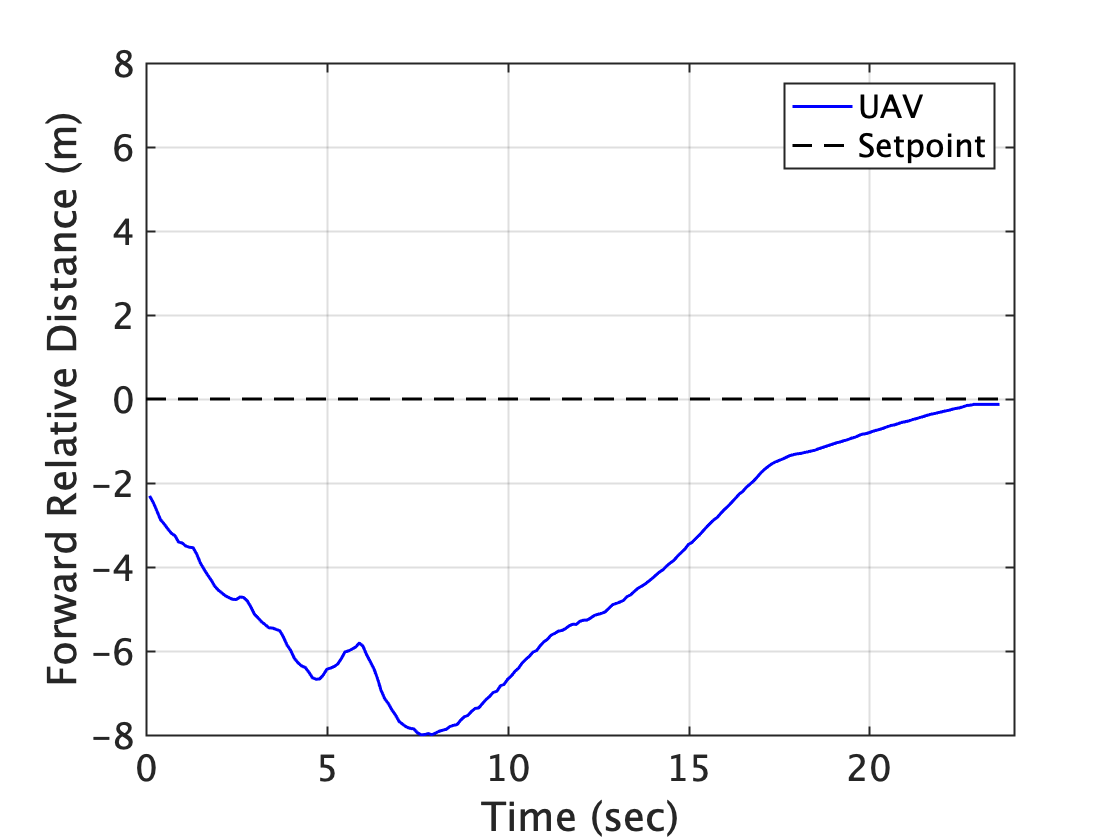}
  \caption{\label{forward_8}Forward Relative Distance (Case-2)} 
\end{figure}
\begin{figure}[!hbt]
\vspace{-0.375cm}
  \ContinuedFloat \raggedleft
   \includegraphics[width=0.9\linewidth]{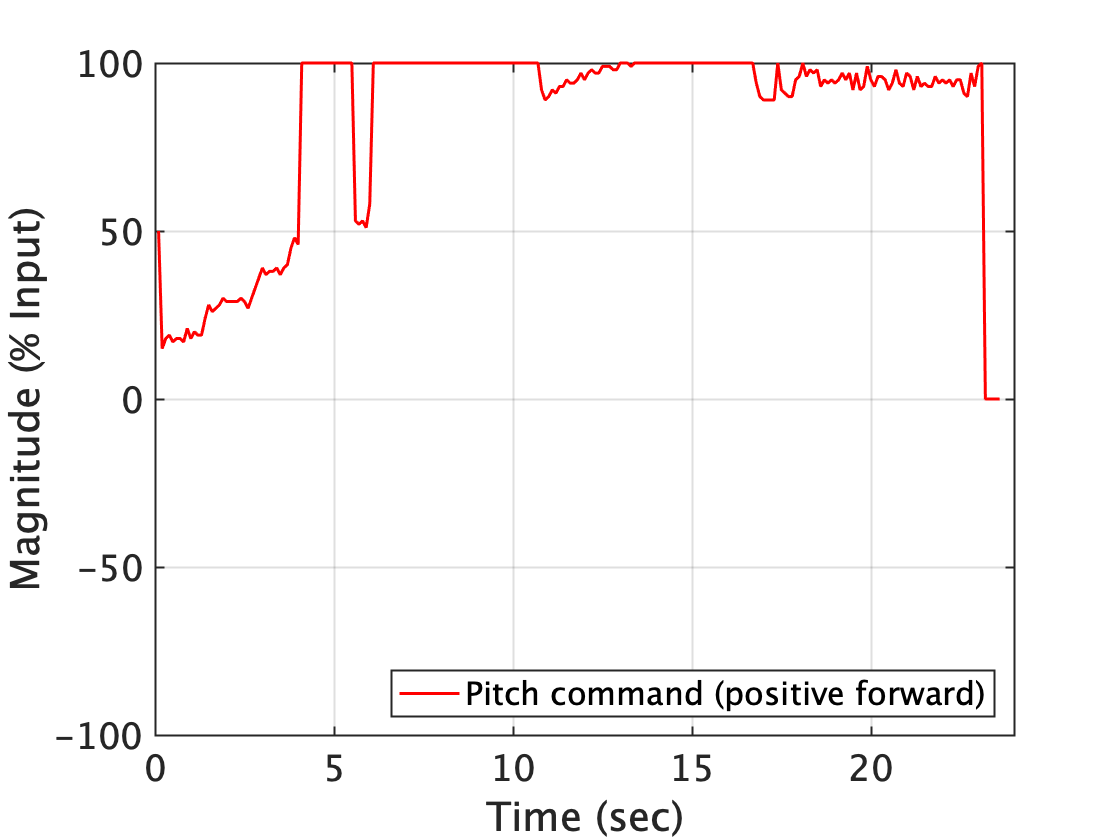}
  \caption{\label{pitch_8}Pitch Command Input (Case-2)} 
\end{figure}

The sideward relative distance and roll command input in time are shown in Figs. \ref{side_8} and \ref{roll_8}, respectively.

\begin{figure}[b]
  \ContinuedFloat* \raggedleft
   \includegraphics[width=0.9\linewidth]{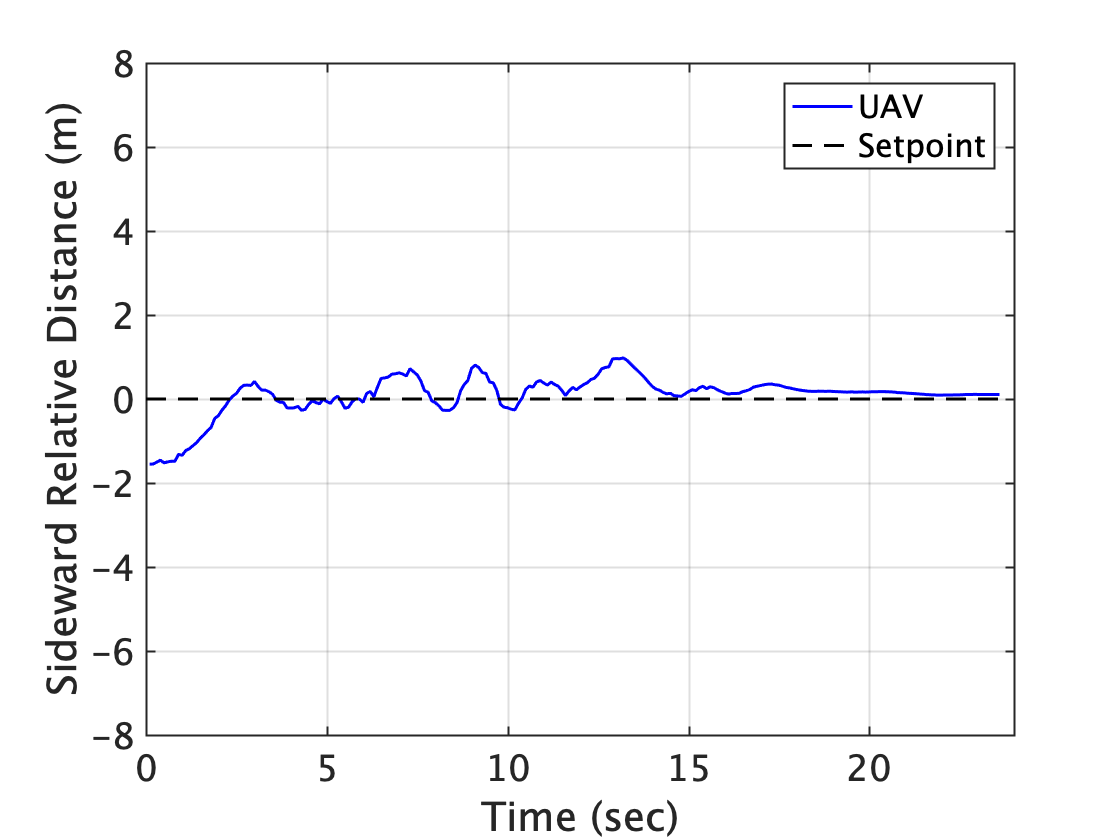}
  \caption{\label{side_8}Sideward Relative Distance (Case-2)} 
\end{figure}
\begin{figure}[!hbt]
 \vspace{-0.3cm}
  \ContinuedFloat \raggedleft
   \includegraphics[width=0.9\linewidth]{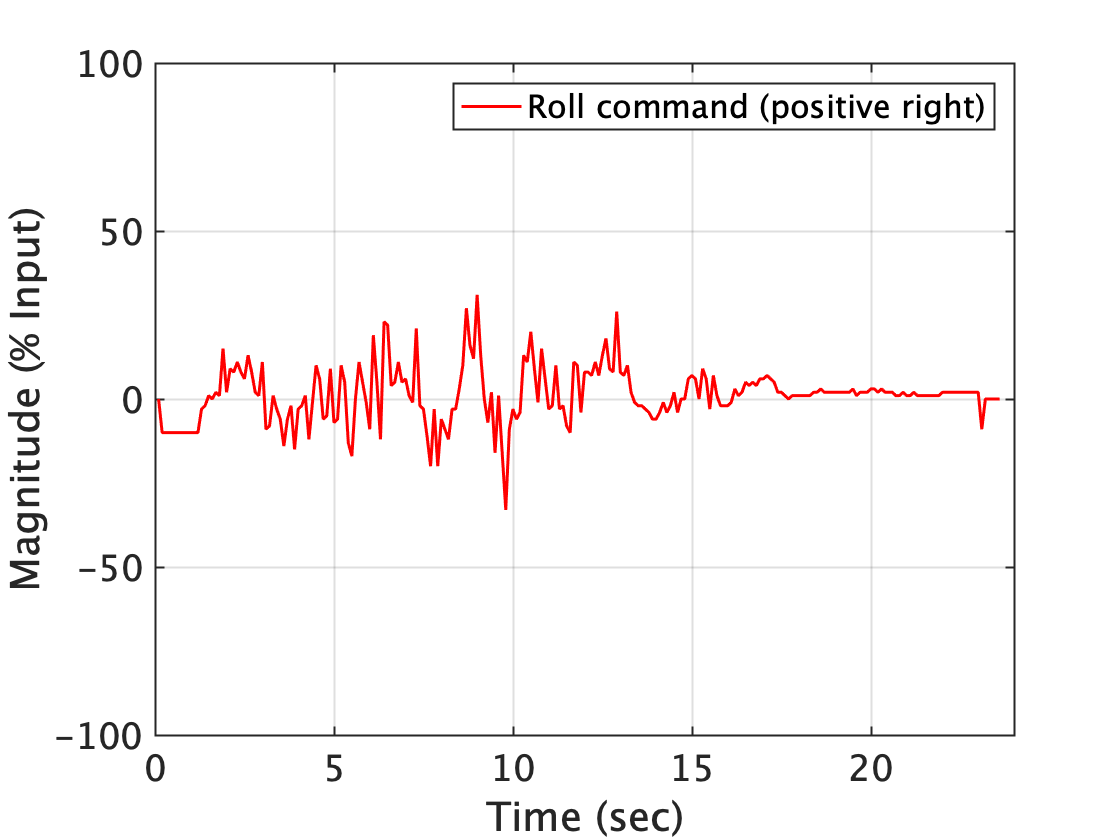}
  \caption{\label{roll_8}Roll Command Input (Case-2)} 
\end{figure}

Initially, the directional approach of safety mode is activated to command roll with a magnitude of 10 based on the visual cue position in a 2D camera view, and level 2 roll controller takes over the control after 1.1 seconds. Between 4 and 17 seconds, it experiences limited control authority caused by maximum pitch forward commands. Thus, the relative distance is fully regulated after 17 seconds. The final sideward landing deviation is 11 cm. 

The vertical relative distance and throttle command input in time are shown in Figs. \ref{vertical_8} and \ref{throttle_8}, respectively.

\begin{figure}[!hbt]
 \vspace{-0.4cm}
  \ContinuedFloat* \raggedleft
   \includegraphics[width=0.9\linewidth]{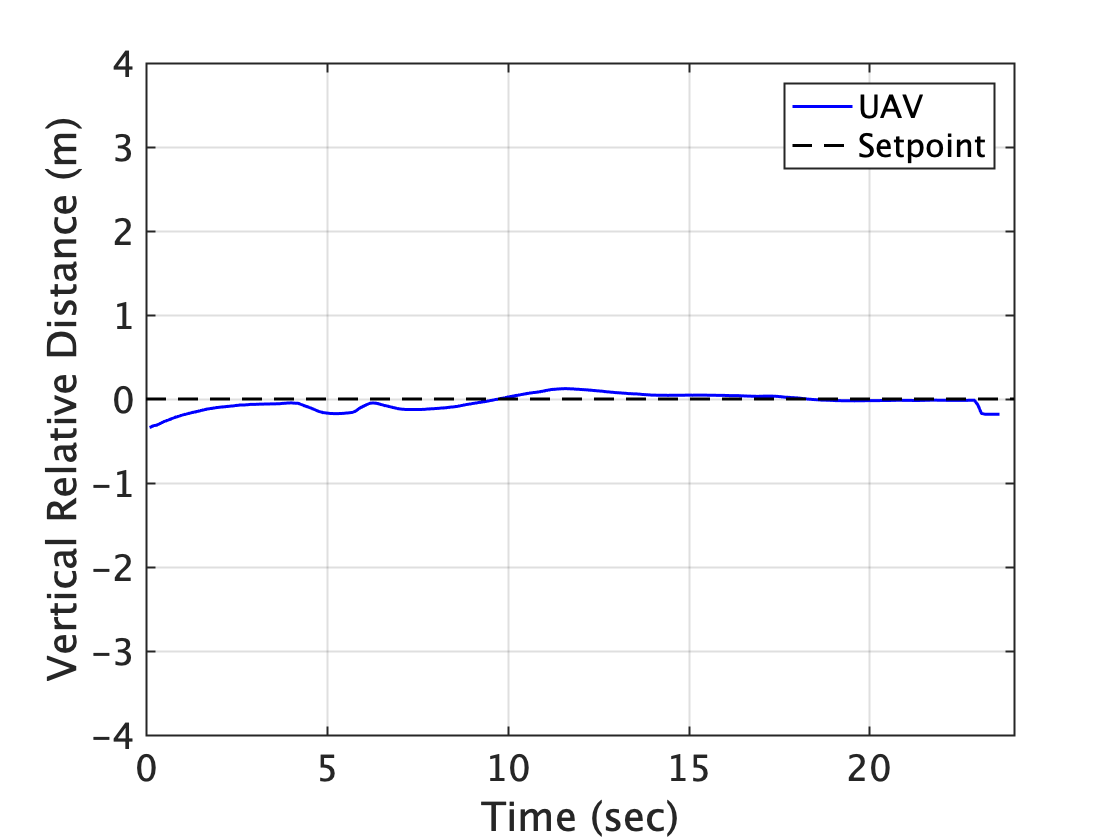}
  \caption{\label{vertical_8}Vertical Relative Distance (Case-2)} 
\end{figure}
\begin{figure}[b!ht]
 \vspace{-0.7cm}
  \ContinuedFloat \raggedleft
  \includegraphics[width=0.9\linewidth]{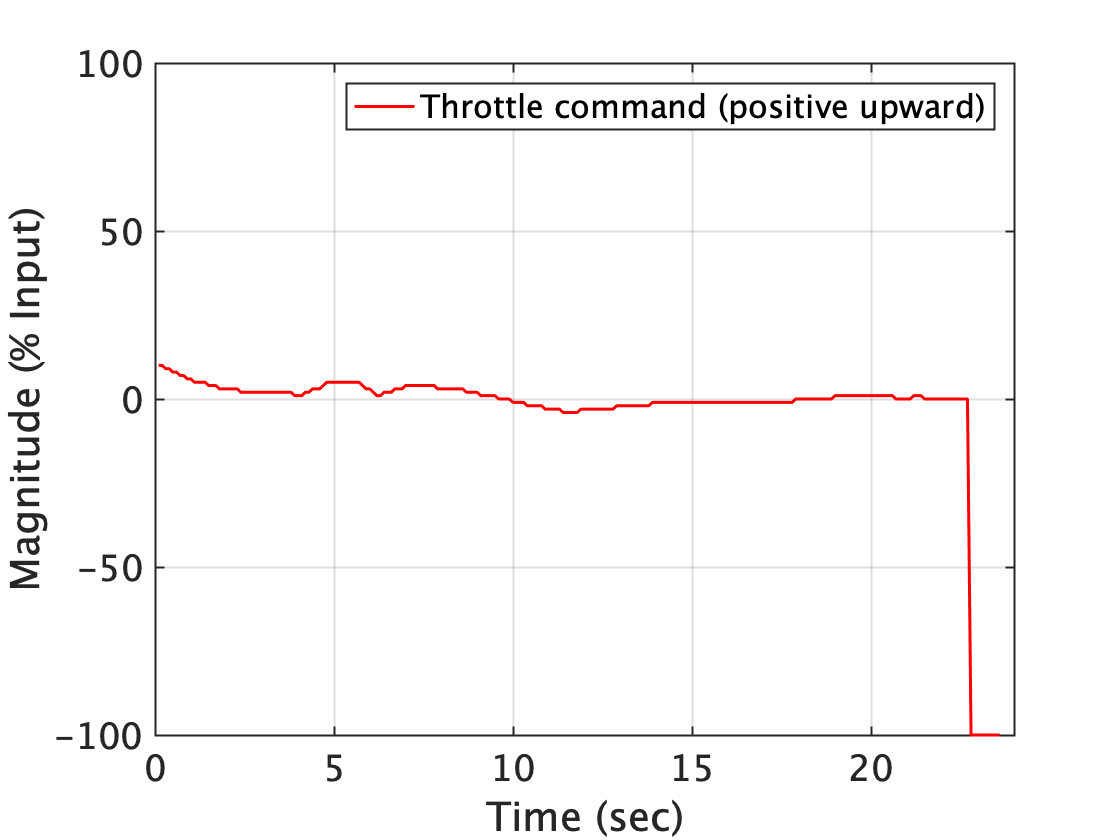}
  \caption{\label{throttle_8}Throttle Command Input (Case-2)} 
\end{figure}

It shows the time history of vertical relative distance with respect to the setpoint which is 15 cm above the landing pad as well as throttle command input after take-off. At 23 seconds, the immediate landing is executed by applying the maximum throttle down command. 

The relative yaw angle and yaw command input in time are shown in Figs. \ref{yaw_8} and \ref{pedal_8}, respectively. After the directional approach of safety mode commands zero yaw for an initial 1 second, it starts regulating the yaw angle by the yaw PID controller with the moving average method using 10 previous data which allows stable tracking. However, it experiences limited control authority due to maximum pitch forward commands between 4 and 17 seconds. Therefore, the yaw angle is fully regulated after 17 seconds. 

\begin{figure}[!hbt]
  \ContinuedFloat* \raggedleft
   \includegraphics[width=0.9\linewidth]{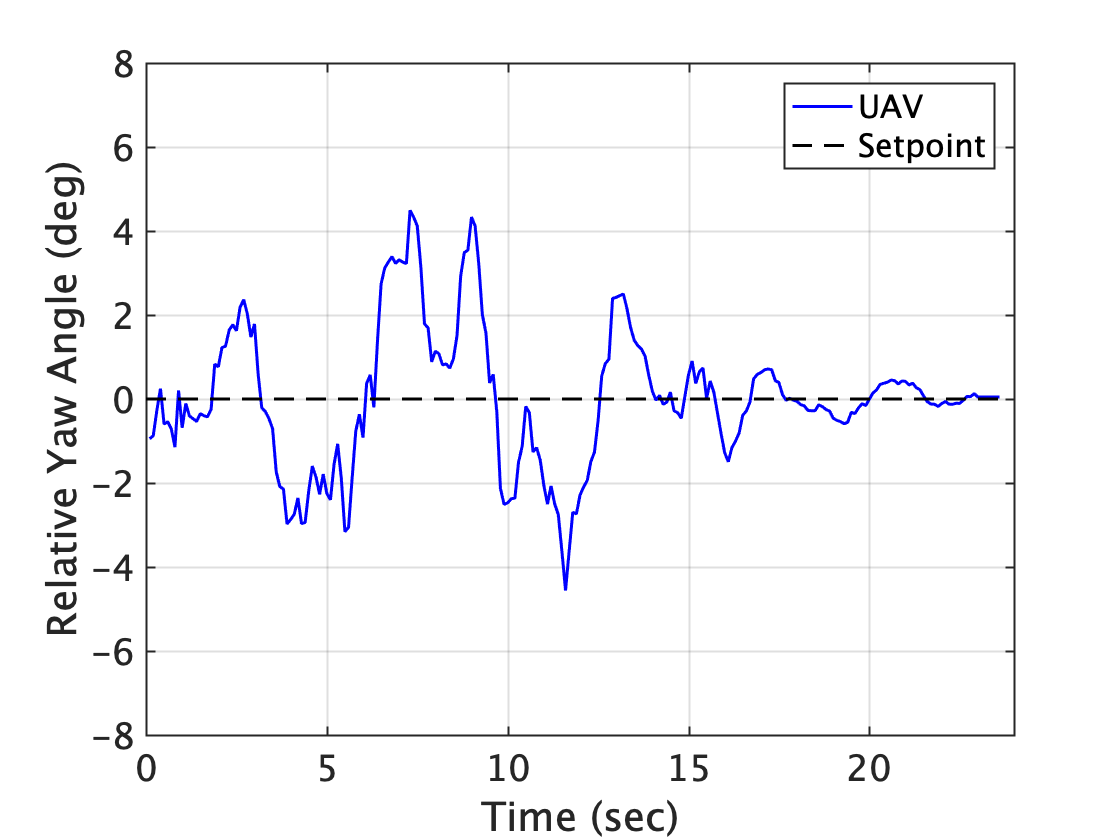}
  \caption{\label{yaw_8}Relative Yaw Angle (Case-2)} 
\end{figure}
\begin{figure}[!hbt]
  \ContinuedFloat \raggedleft
   \includegraphics[width=0.9\linewidth]{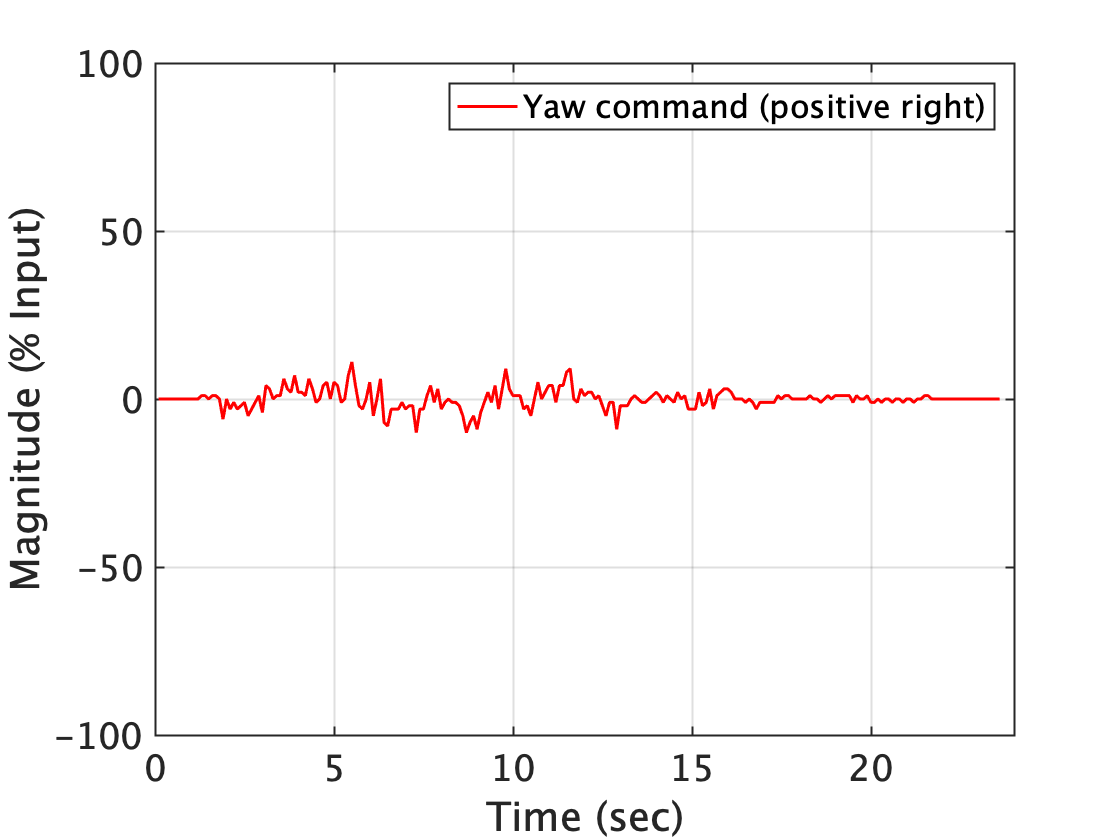}
  \caption{\label{pedal_8}Yaw Command Input (Case-2)} 
\end{figure}

\item \textbf{Case-3: Platform moving in S-pattern:} The platform is programmed to move in an S-pattern at a constant speed of 1 m/s and the trajectories viewed from the top and side have minimal overshoot as shown in Figs. \ref{top_view_s} and \ref{side_view_s}, respectively. \\

\newpage
\begin{figure}[t]
  \ContinuedFloat* \raggedleft
   \includegraphics[width=0.9\linewidth]{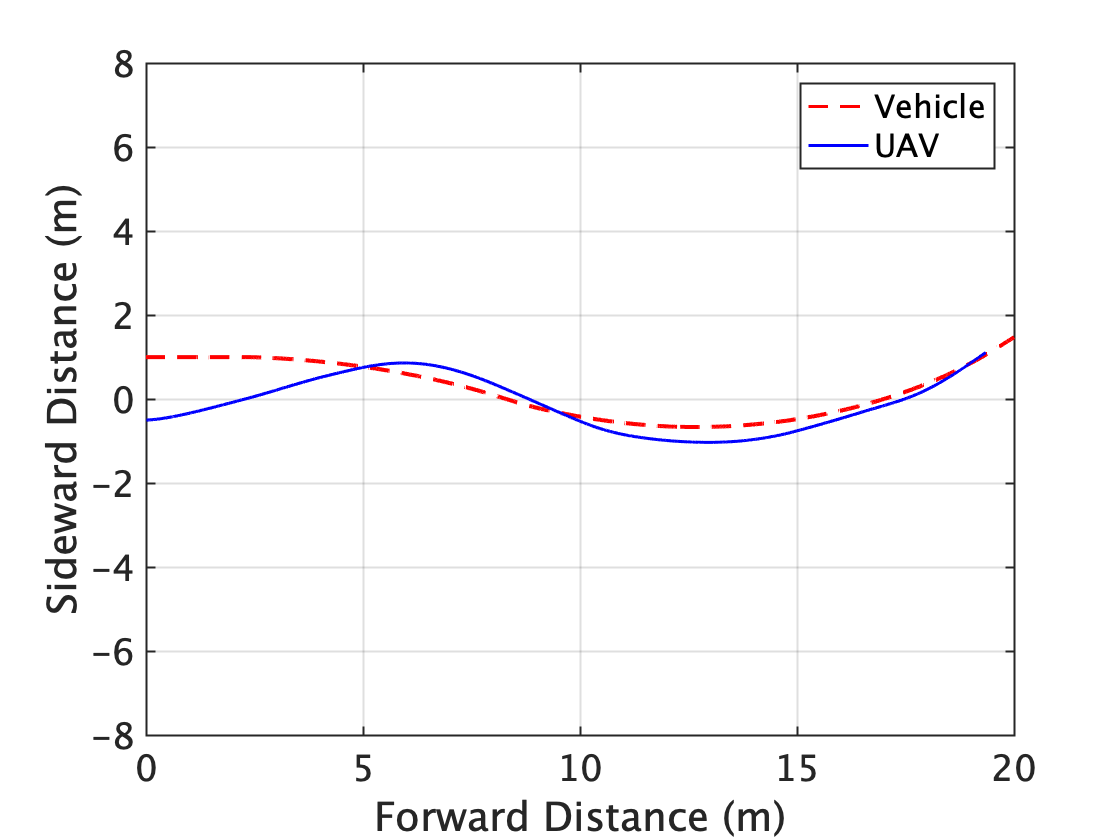}
  \caption{\label{top_view_s}Trajectories in Top View (Case-3)} 
\end{figure}
\begin{figure}[!hbt]
 \vspace{0.4cm}
  \ContinuedFloat \raggedleft
   \includegraphics[width=0.9\linewidth]{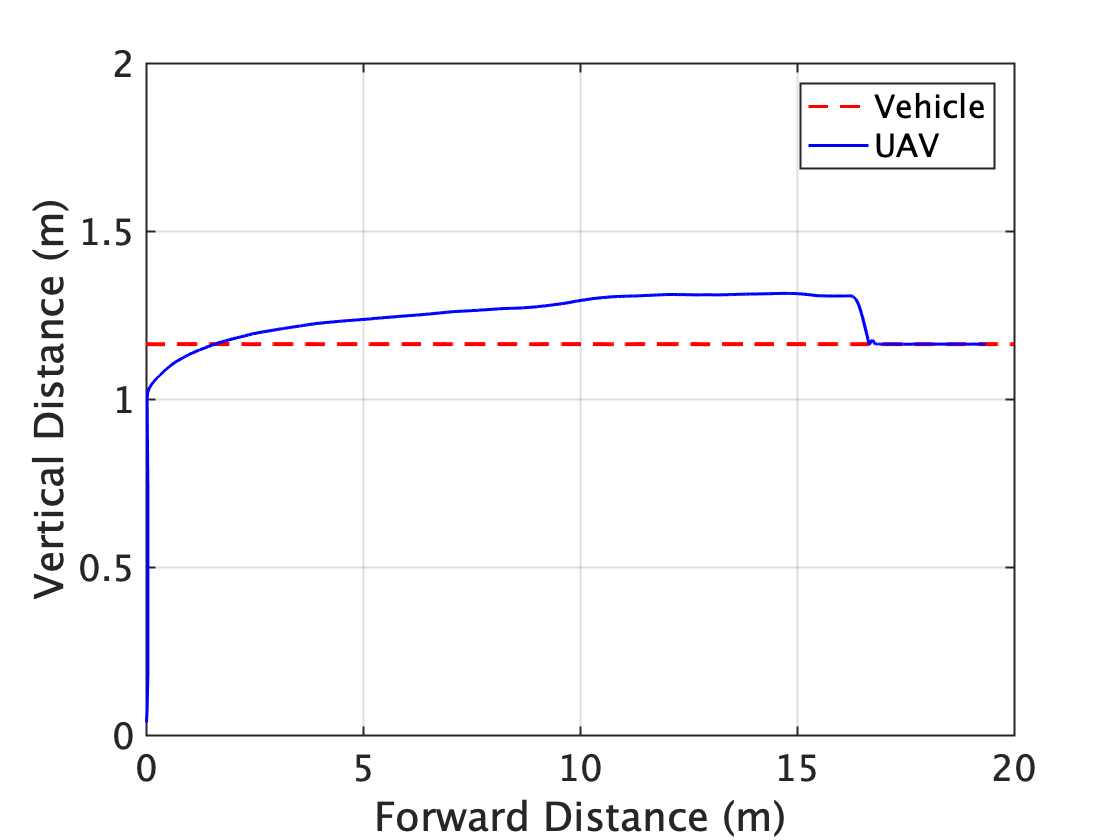}
  \caption{\label{side_view_s}Trajectories in Side View (Case-3)} 
\end{figure}

The forward relative distance and pitch command input in time are shown in Figs. \ref{forward_s} and \ref{pitch_s}, respectively. The pitch controller is switched from level 2 to 3 upon reaching the distance of 1.5 meters at 3.7 seconds and the final forward landing deviation is 9 cm.\\

\begin{figure}[!hbt]

  \ContinuedFloat* \raggedleft
   \includegraphics[width=0.9\linewidth]{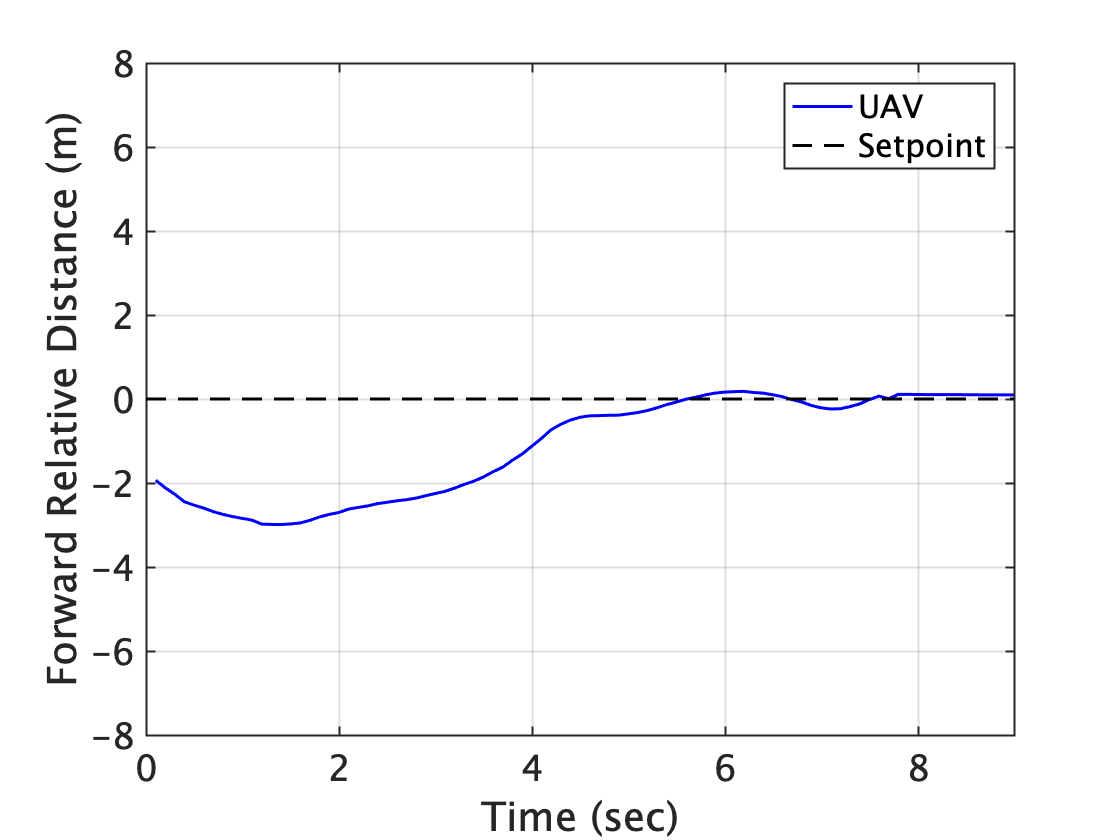}
  \caption{\label{forward_s}Forward Relative Distance (Case-3)} 
\end{figure}
\begin{figure}[t]
  \ContinuedFloat \raggedleft
   \includegraphics[width=0.9\linewidth]{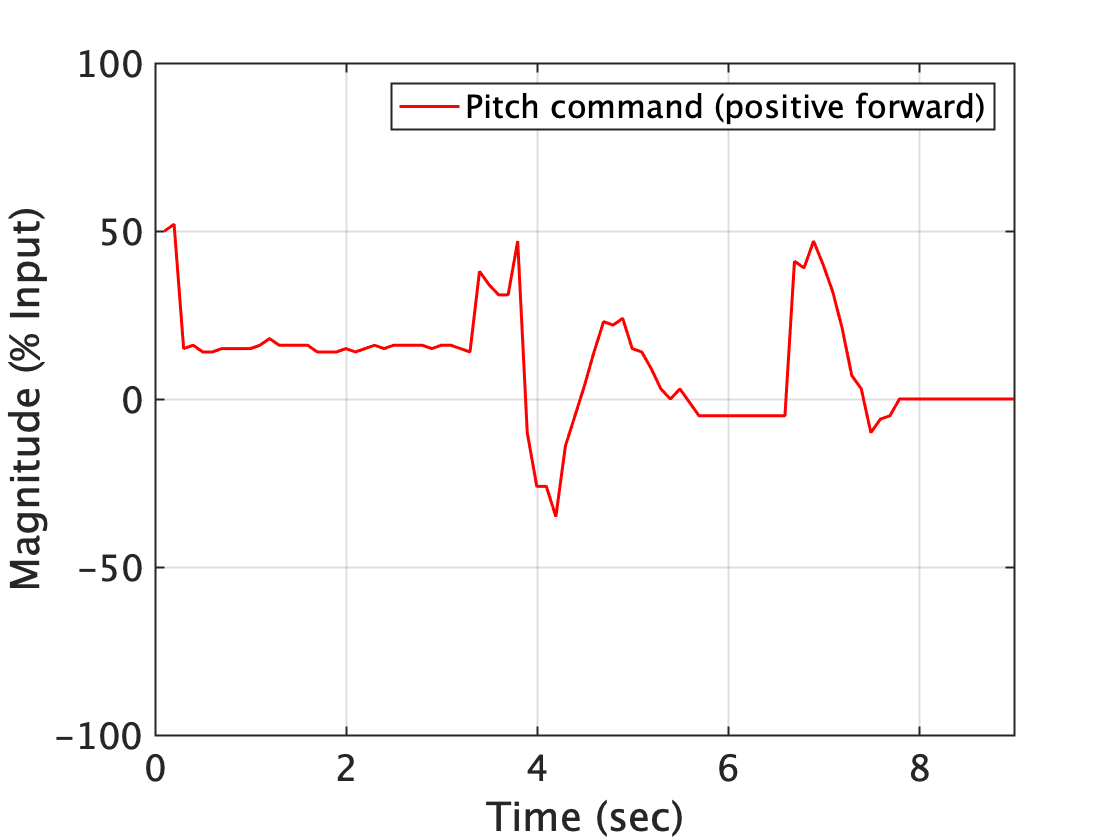}
  \caption{\label{pitch_s}Pitch Command Input (Case-3)} 
\end{figure}

The sideward relative distance and roll command input in time are shown in Figs. \ref{side_s} and \ref{roll_s}, respectively. The directional approach of safety mode is activated for an initial 1 second followed by level 1 roll controller. At 2.5 seconds, the level 2 controller is activated for precise tracking and the sideward landing deviation is 13 cm.

\begin{figure}[!hbt]
  \ContinuedFloat* \raggedleft
   \includegraphics[width=0.9\linewidth]{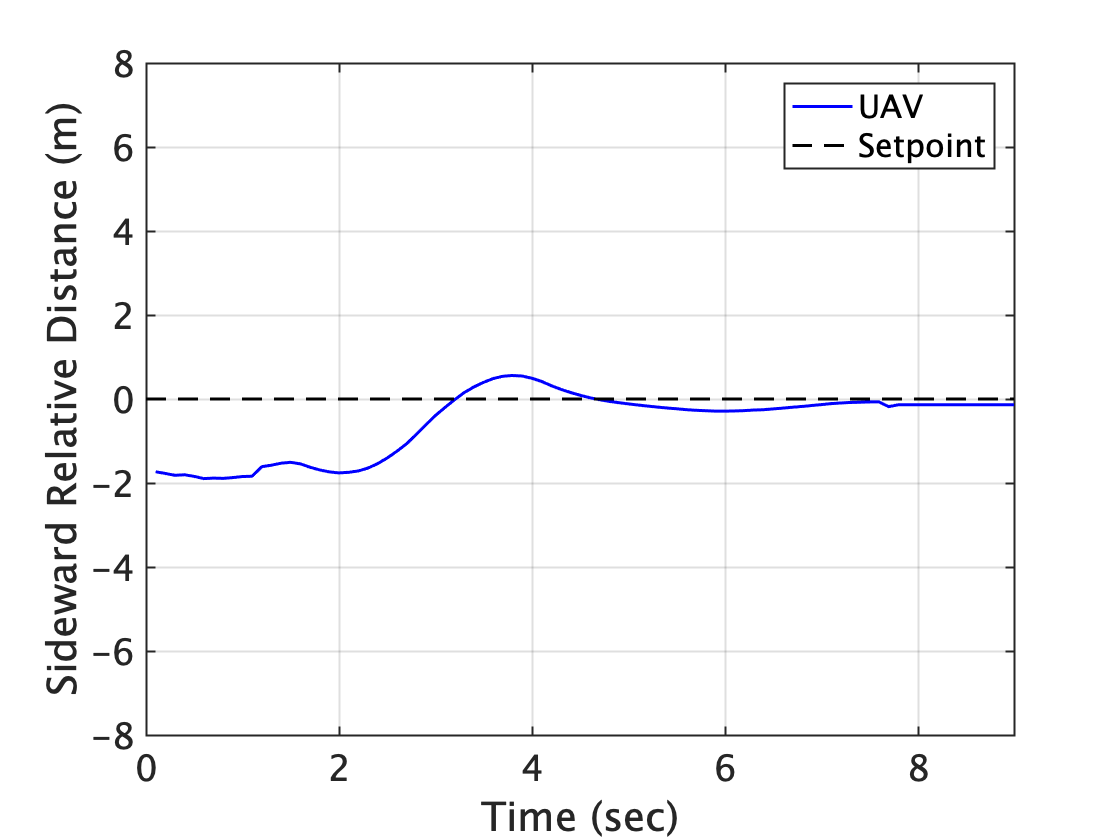}
  \caption{\label{side_s}Sideward Relative Distance (Case-3)} 
\end{figure}
\begin{figure}[!hbt]
  \ContinuedFloat \raggedleft
   \includegraphics[width=0.9\linewidth]{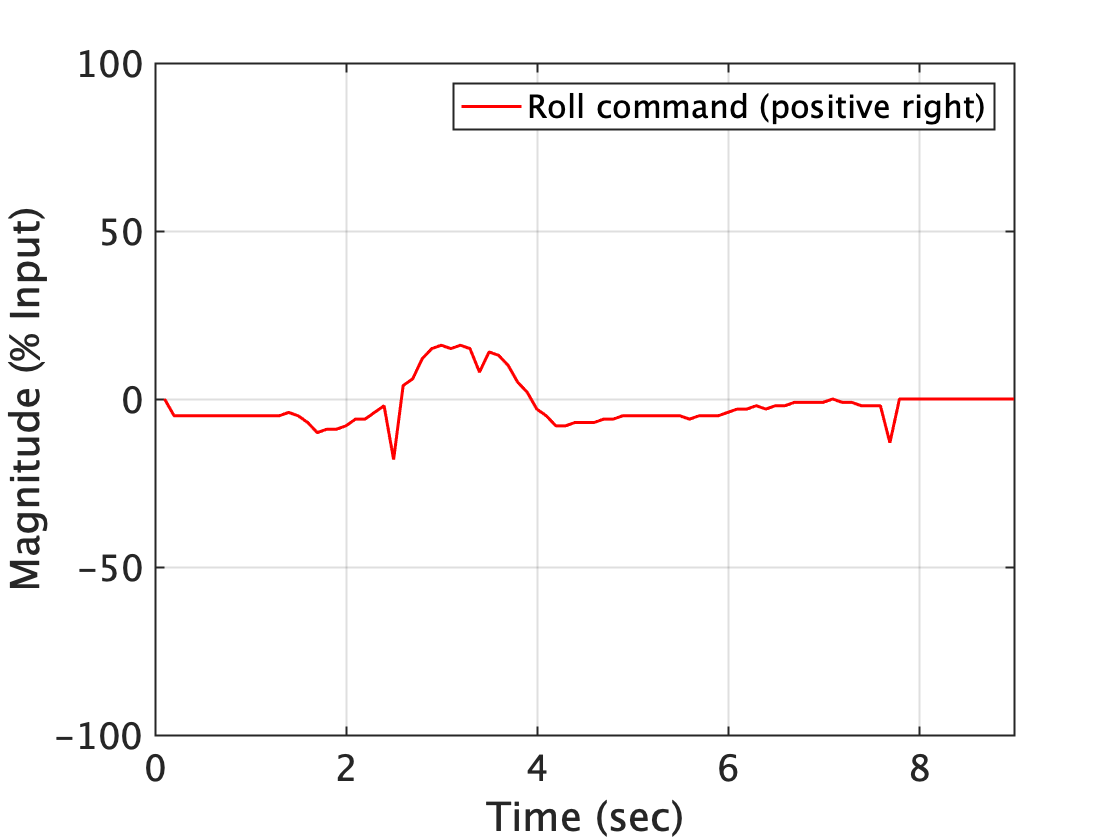}
  \caption{\label{roll_s}Roll Command Input (Case-3)} 
\end{figure}

The vertical relative distance and throttle command input in time are shown in Figs. \ref{vertical_s} and \ref{throttle_s}, respectively. The zero setpoint is 15 cm above the landing pad and it commands maximum throttle down to land immediately at 7.8 seconds where the landing conditions are satisfied.

\begin{figure}[!hbt]
 \vspace{-0.3cm}
  \ContinuedFloat* \raggedleft
   \includegraphics[width=0.9\linewidth]{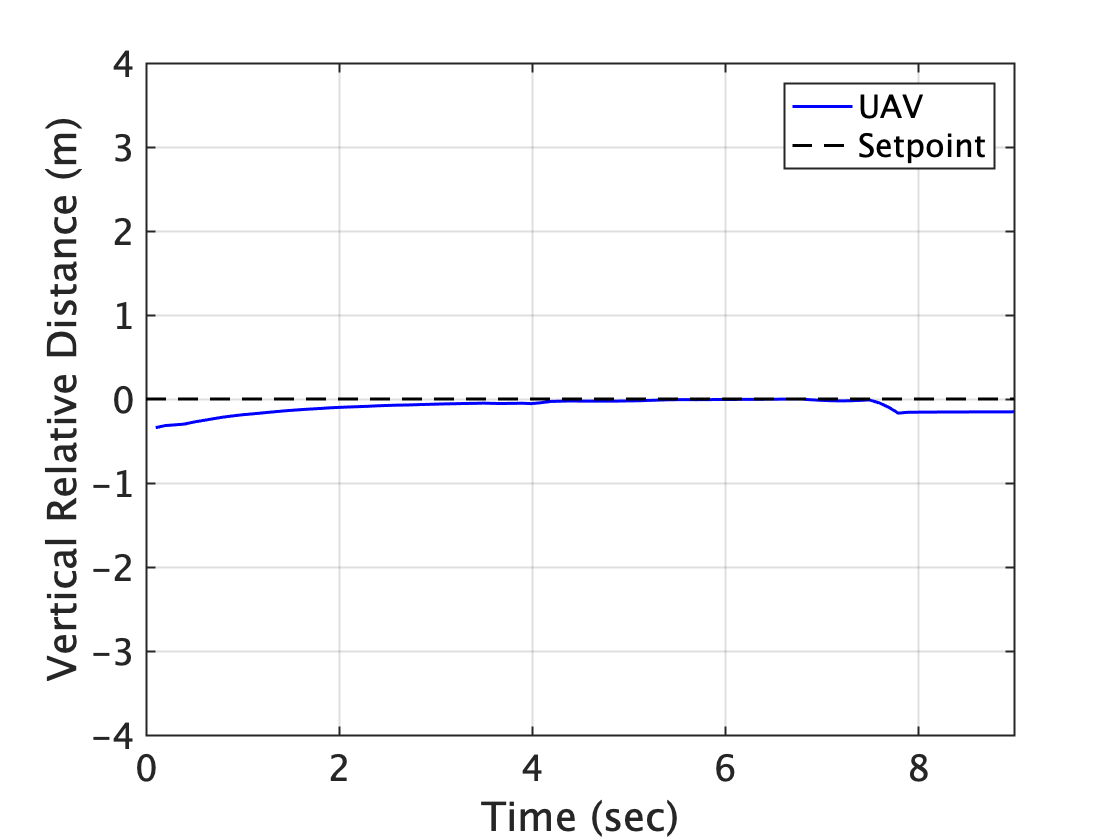}
  \caption{\label{vertical_s}Vertical Relative Distance (Case-3)} 
\end{figure}
\begin{figure}[!hbt]
 \vspace{-0.2cm}
  \ContinuedFloat \raggedleft
   \includegraphics[width=0.9\linewidth]{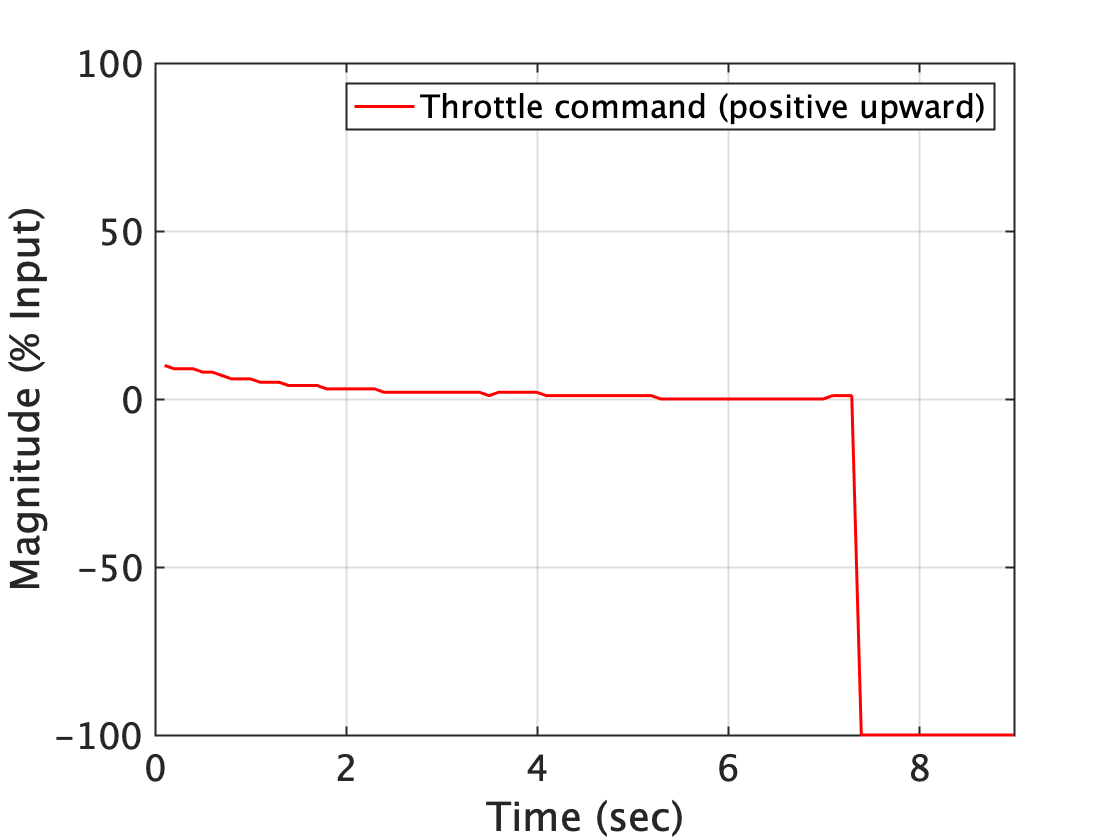}
  \caption{\label{throttle_s}Throttle Command Input (Case-3)} 
\end{figure}

The relative yaw angle and yaw command input in time are shown in Figs. \ref{yaw_s} and \ref{pedal_s}, respectively.

\begin{figure}[b]
 \vspace{-0.1cm}
  \ContinuedFloat* \raggedleft
   \includegraphics[width=0.9\linewidth]{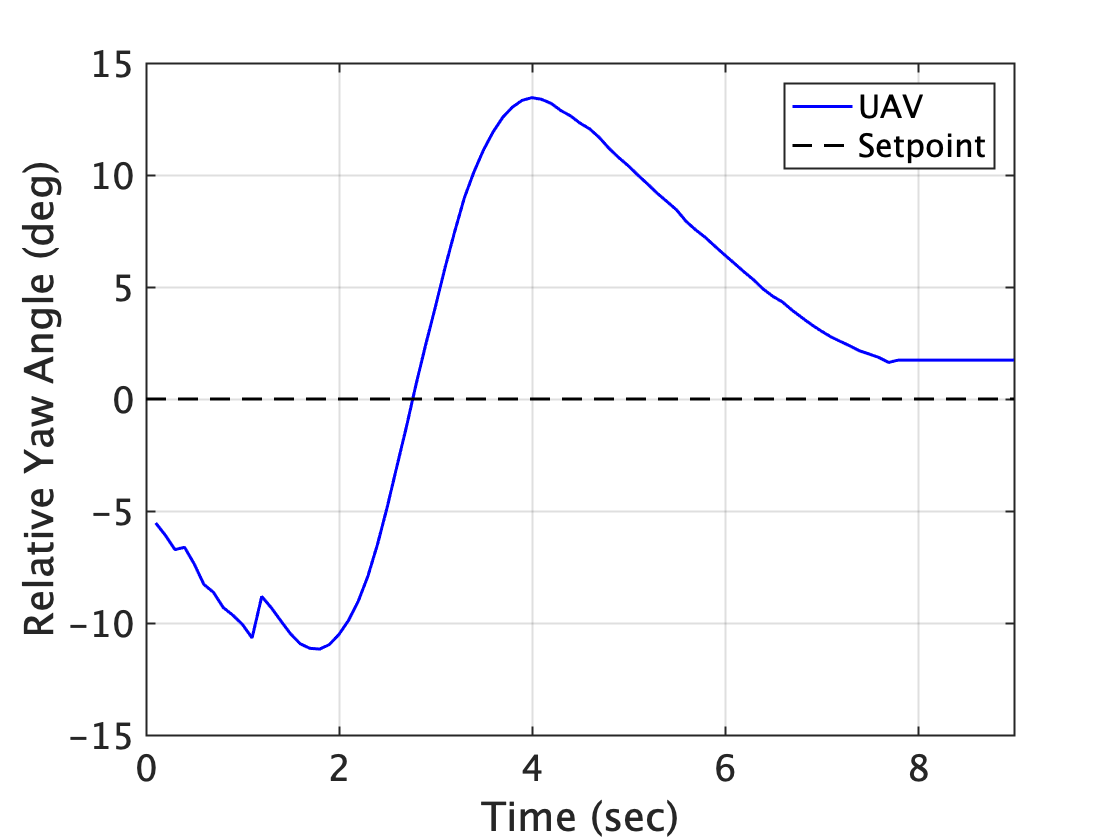}
  \caption{\label{yaw_s}Relative Yaw Angle (Case-3)} 
\end{figure}
\begin{figure}[!hbt]
 \vspace{-0.3cm}
  \ContinuedFloat \raggedleft
   \includegraphics[width=0.9\linewidth]{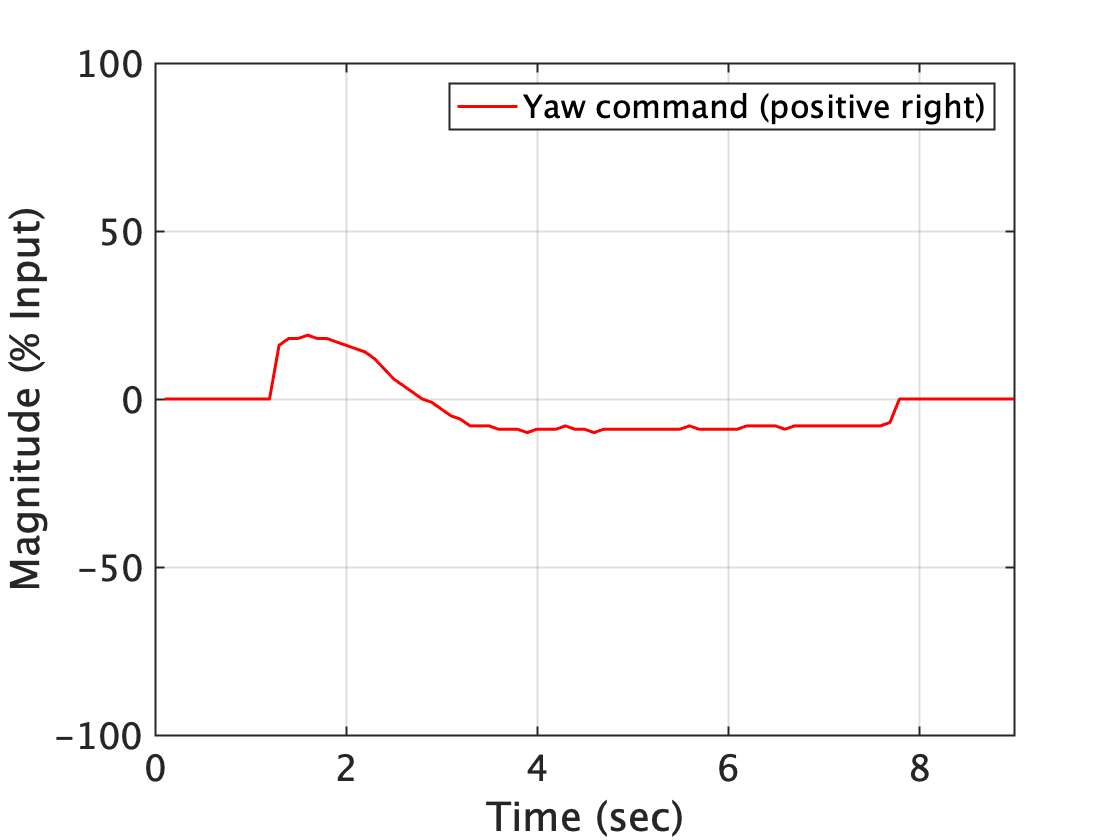}
  \caption{\label{pedal_s}Yaw Command Input (Case-3)} 
\end{figure}

Due to the platform changing its course, the yaw angle is relatively big. However, the level 2 yaw controller regulates the yaw angle effectively. Since the yaw angle is not one of the landing conditions, the UAV lands with a yaw angle of 1.7 degrees.

\item \textbf{Case-4: Platform moving in circular pattern:} The platform is programmed to move in a circular pattern at a speed of 1 m/s and the trajectories viewed from the top and side are shown in Figs. \ref{top_view_c} and \ref{side_view_c}, respectively.  

\begin{figure}[!hbt]
 \vspace{-0.3cm}
  \ContinuedFloat* \raggedleft
   \includegraphics[width=0.9\linewidth]{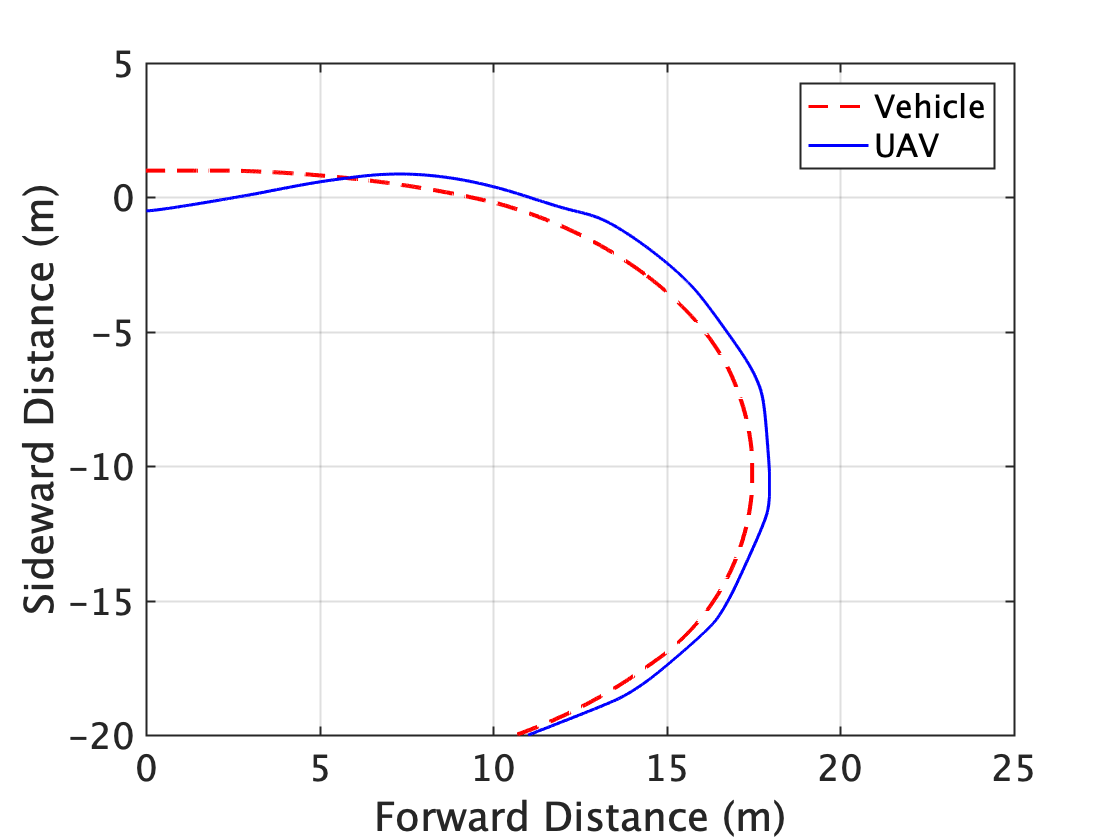}
  \caption{\label{top_view_c}Trajectories in Top View (Case-4)} 
\end{figure}
\begin{figure}[!hbt]
 \vspace{-0.6cm}
  \ContinuedFloat \raggedleft
   \includegraphics[width=0.9\linewidth]{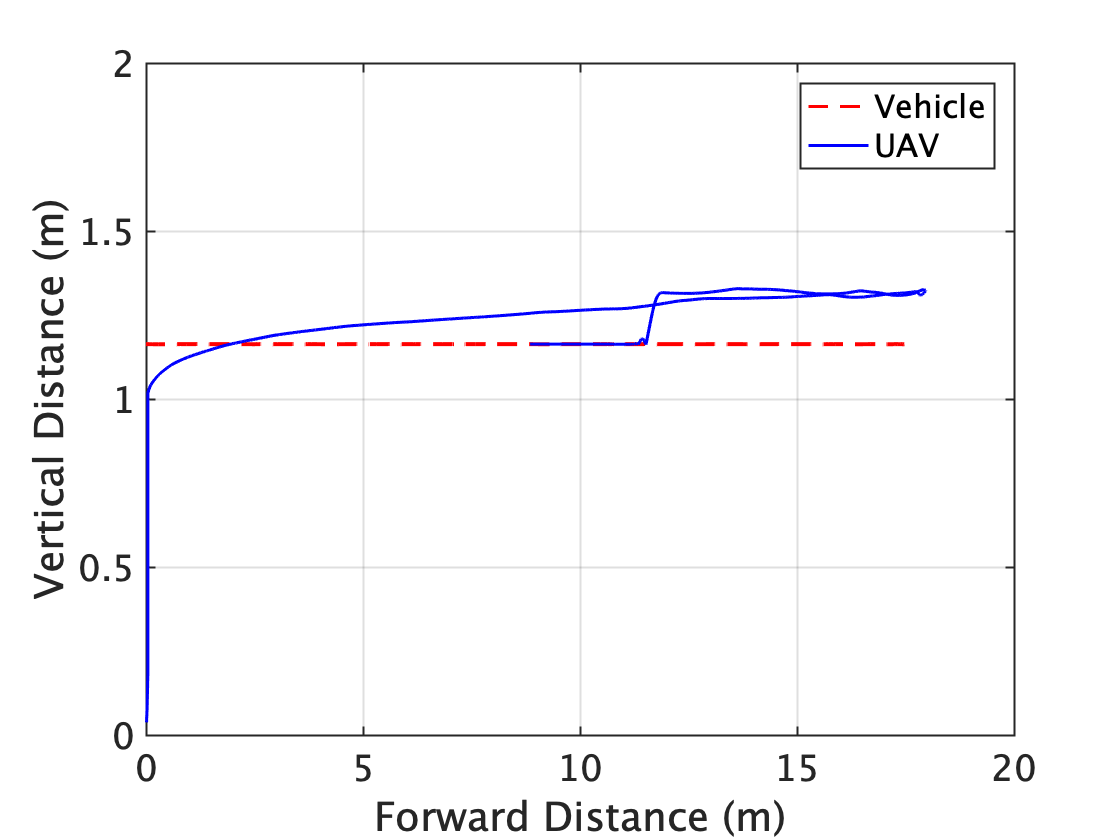}
  \caption{\label{side_view_c}Trajectories in Side View (Case-4)} 
\end{figure}

Since the platform turns more than 90 degrees, the trajectory in the side view overlaps as the platform comes back. The entire trajectory shows that the UAV tracks the circling platform with minimal overshoot.

The forward relative distance and pitch command input in time are shown in Figs. \ref{forward_c} and \ref{pitch_c}, respectively. The level 2 pitch controller operates until 3.8 seconds followed by level 3 controller. While tracking, the safety mode is intermittently activated to prevent a potential collision with platform structure, which commands -10 pitch input if the UAV crosses the pad center. It lands with a forward deviation of 8 cm. 

\begin{figure}[!hbt]
  \ContinuedFloat* \raggedleft
   \includegraphics[width=0.9\linewidth]{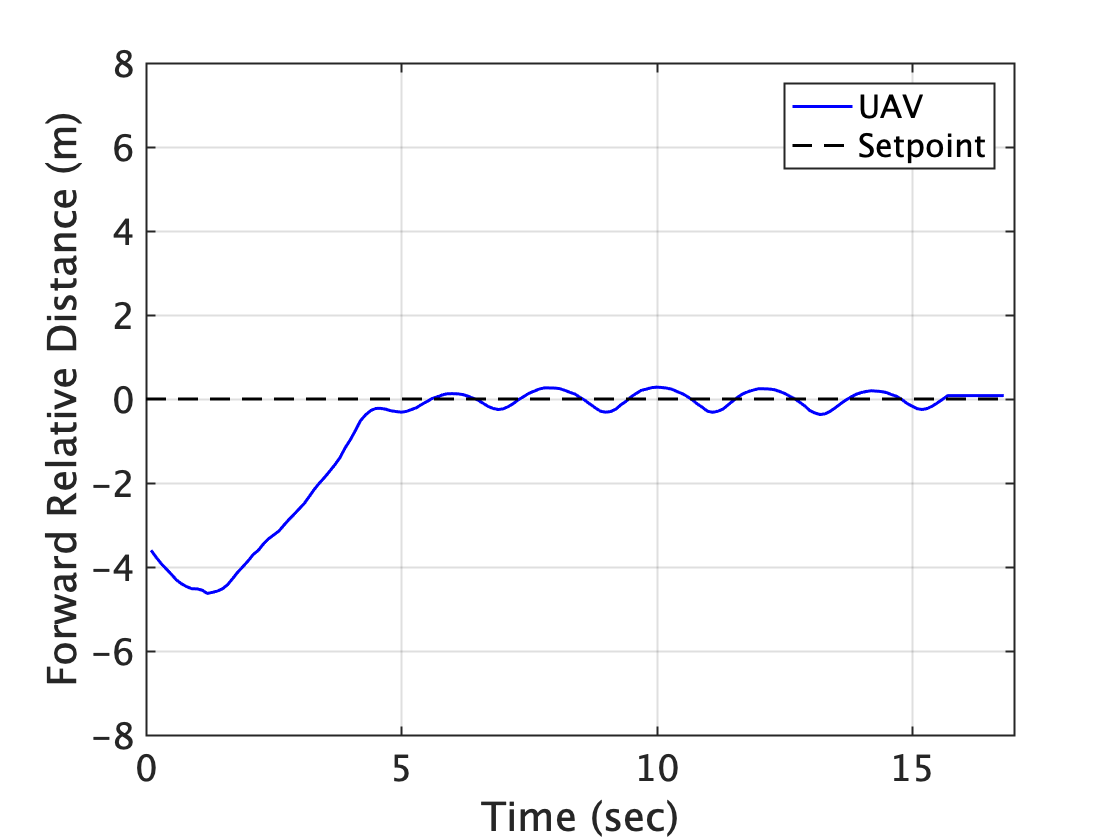}
  \caption{\label{forward_c}Forward Relative Distance (Case-4)} 
\end{figure}
\begin{figure}[!hbt]
 \vspace{0.3cm}
  \ContinuedFloat \raggedleft
   \includegraphics[width=0.9\linewidth]{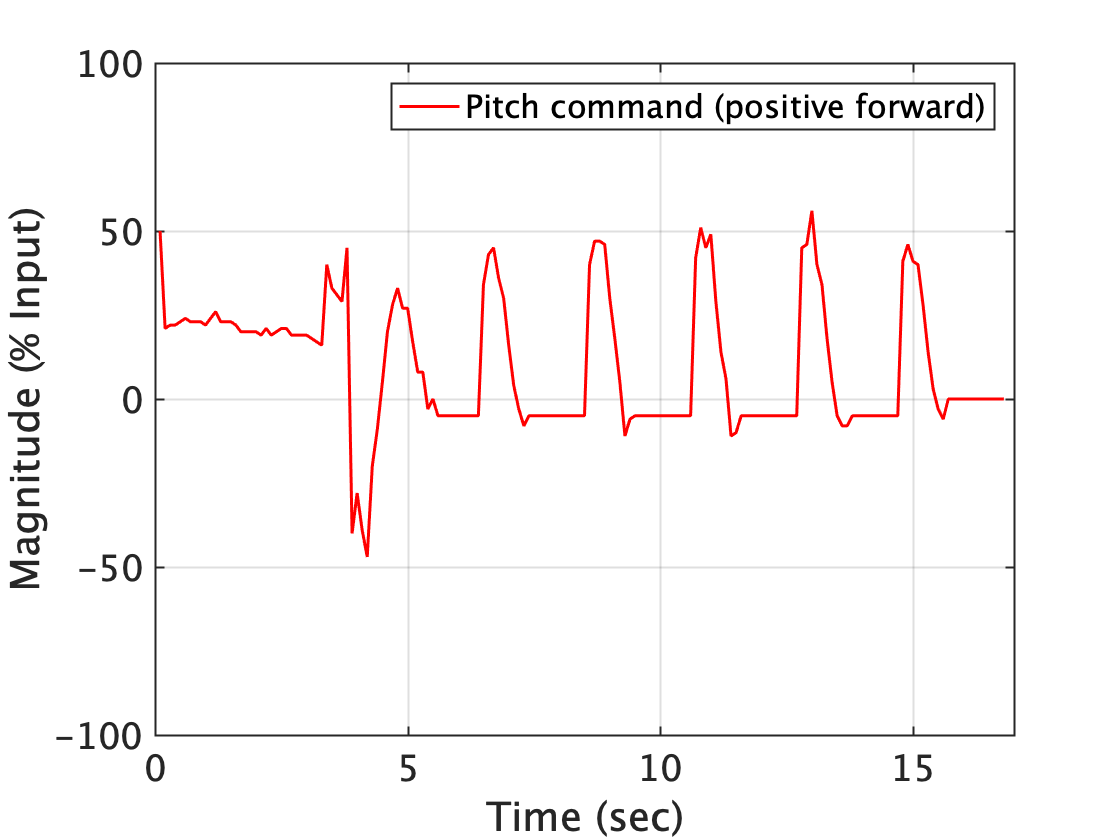}
  \caption{\label{pitch_c}Pitch Command Input (Case-4)} 
\end{figure}

The sideward relative distance and roll command input in time are shown in Figs. \ref{side_c} and \ref{roll_c}, respectively. The directional approach of safety mode is activated for an initial 1 second followed by level 1 controller. At 2.5 seconds, the level 2 controller is activated for precise tracking and it takes a relatively long time to regulate because the vehicle moves in a way to increase sideward relative distance. The final landing deviation in the sideward direction is 14 cm.

\begin{figure}[t]
 \vspace{-0.3cm}
  \ContinuedFloat* \raggedleft
   \includegraphics[width=0.9\linewidth]{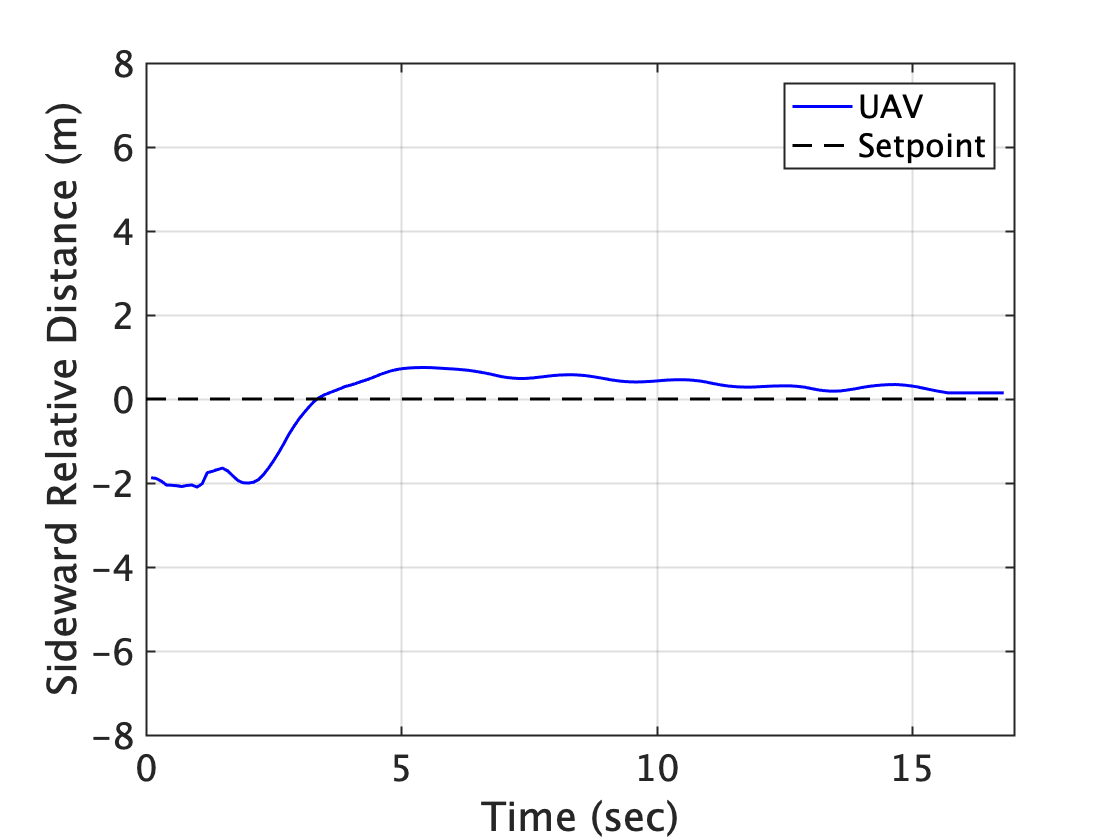}
  \caption{\label{side_c}Sideward Relative Distance (Case-4)} 
\end{figure}
\begin{figure}[!hbt]
 \vspace{0.3cm}
  \ContinuedFloat \raggedleft
   \includegraphics[width=0.9\linewidth]{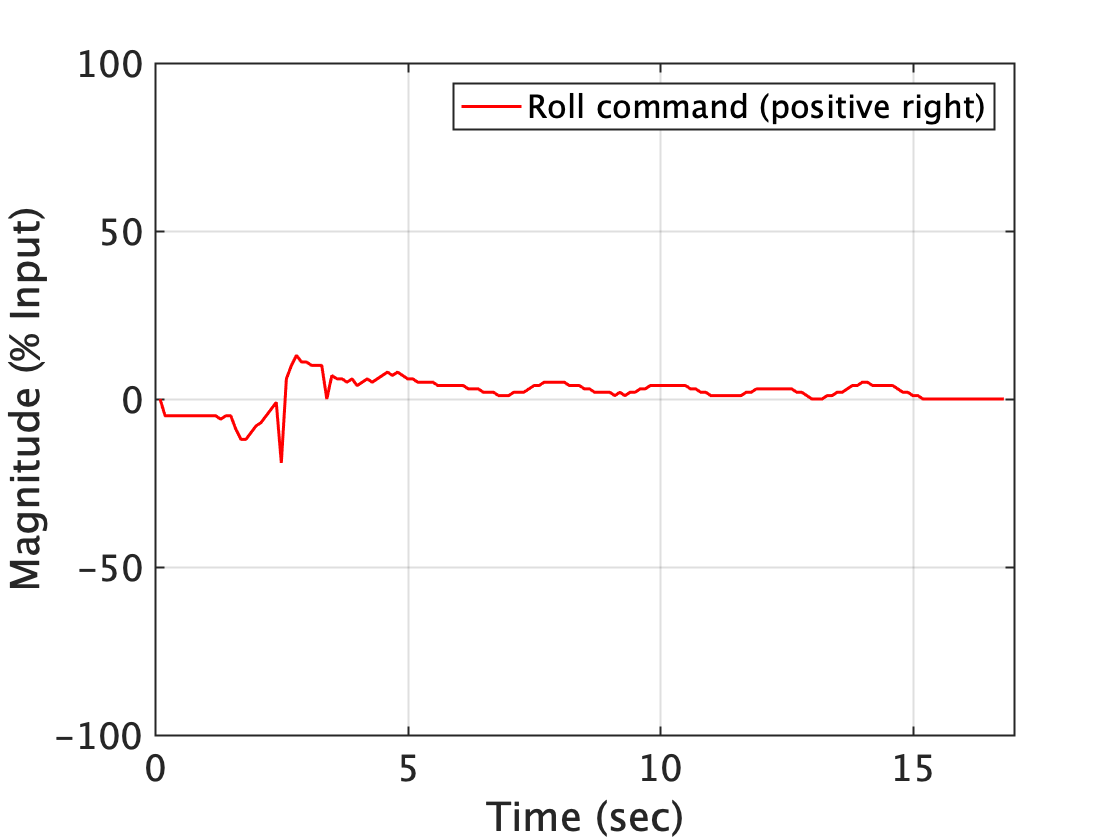}
  \caption{\label{roll_c}Roll Command Input (Case-4)} 
\end{figure}

The vertical relative distance and throttle command input in time are shown in Figs. \ref{vertical_c} and \ref{throttle_c}, respectively. The zero setpoint means 15 cm above the landing pad and it commands the maximum throttle down to land immediately at 15.7 seconds at which the landing conditions are satisfied.

\begin{figure}[!hbt]
 \vspace{0.4cm}
  \ContinuedFloat* \raggedleft
   \includegraphics[width=0.9\linewidth]{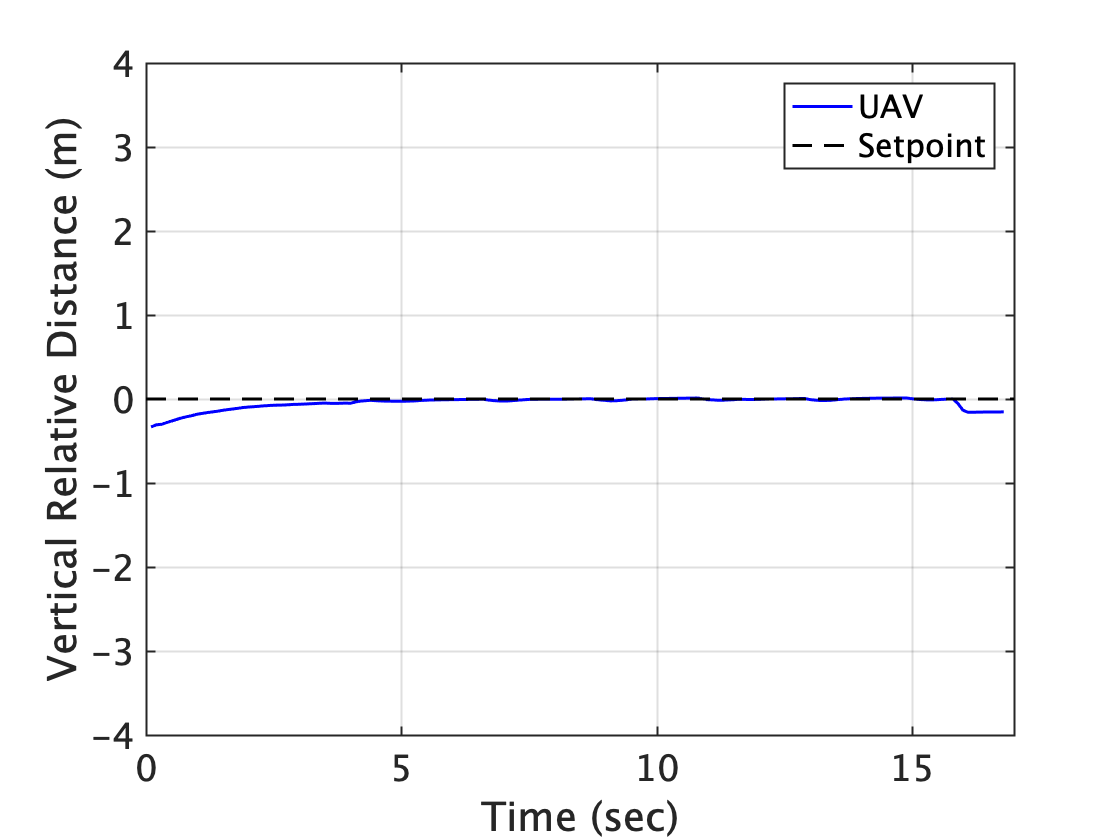}
  \caption{\label{vertical_c}Vertical Relative Distance (Case-4)} 
\end{figure}
\begin{figure}[!hbt]
 \vspace{-0.3cm}
  \ContinuedFloat \raggedleft
   \includegraphics[width=0.9\linewidth]{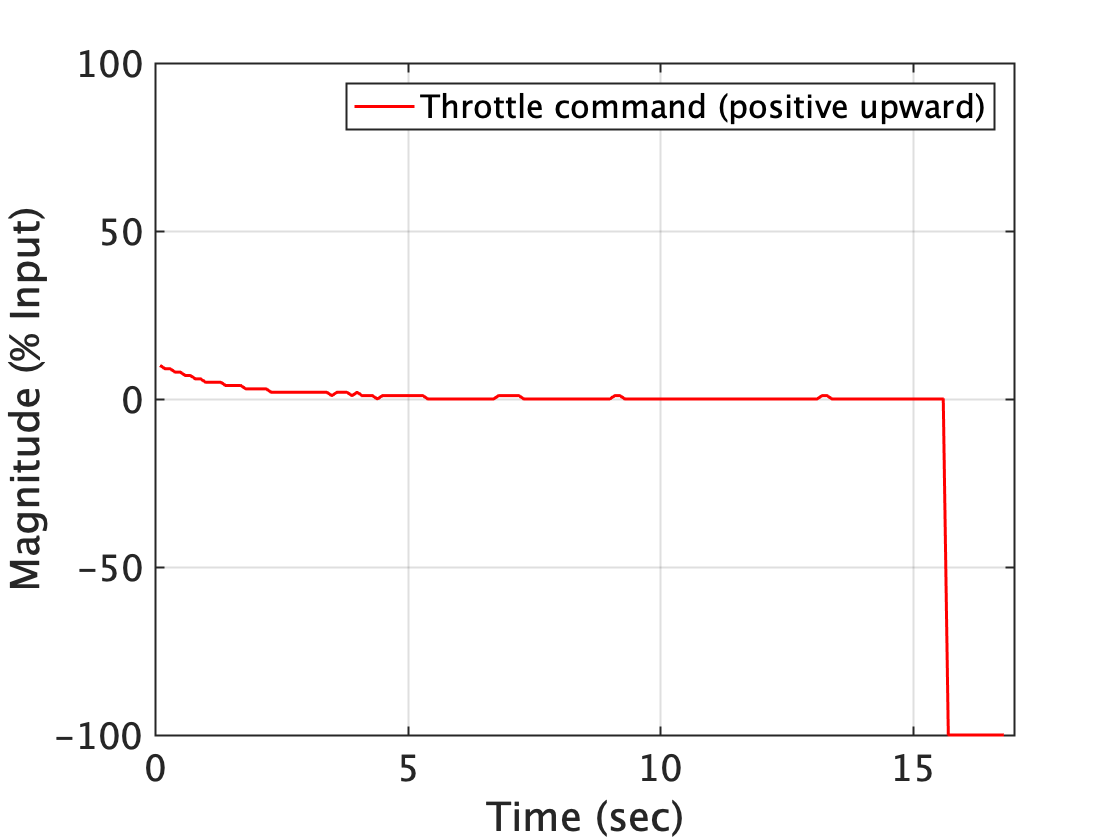}
  \caption{\label{throttle_c}Throttle Command Input (Case-4)} 
\end{figure}

The relative yaw angle and yaw command input in time are shown in Figs. \ref{yaw_c} and \ref{pedal_c}, respectively. The directional approach of safety mode commands zero yaw for an initial 1 second. Due to the circular trajectory of the platform, the yaw angle is relatively large, however, kept less than 15 degrees. Thus, the level 2 yaw controller with the moving average method is activated to regulate the yaw angle. Since the yaw angle is not one of the landing conditions, the UAV lands with having a yaw angle of 1.8 degrees at 15.7 seconds.

\begin{figure}[!hbt]
 \vspace{-0.4cm}
  \ContinuedFloat* \raggedleft
   \includegraphics[width=0.9\linewidth]{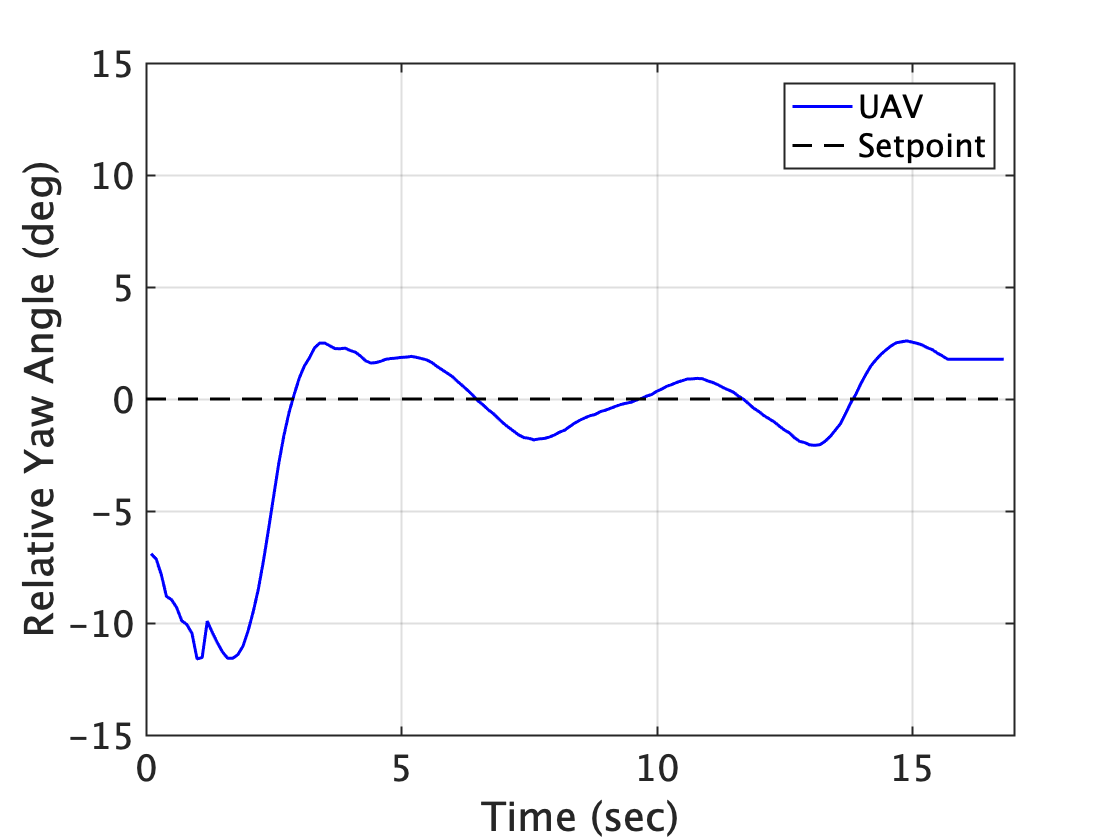}
  \caption{\label{yaw_c}Relative Yaw Angle (Case-4)} 
\end{figure}
\begin{figure}[b]
 \vspace{-0.4cm}
  \ContinuedFloat \raggedleft
   \includegraphics[width=0.9\linewidth]{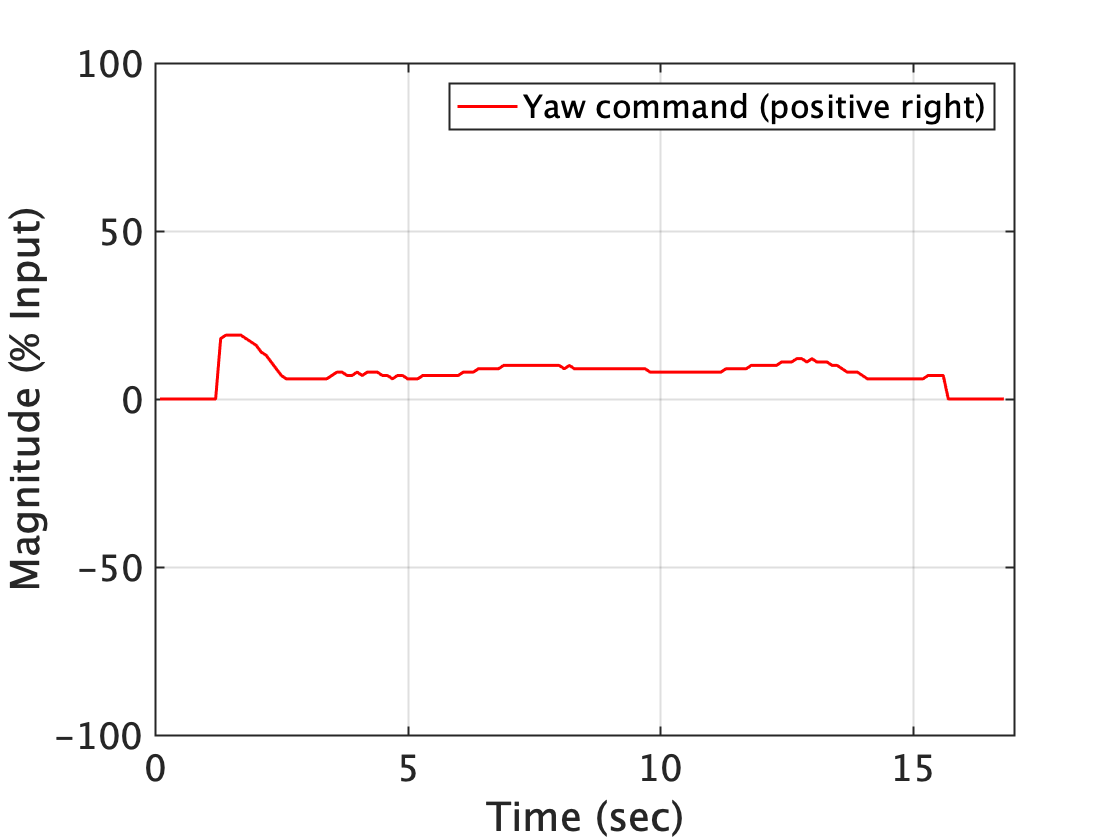}
  \caption{\label{pedal_c}Yaw Command Input (Case-4)} 
\end{figure}

\end{enumerate}

\section{Flight Tests}

\begin{figure}[hbt!]
\centering
\includegraphics[width=0.485\textwidth]{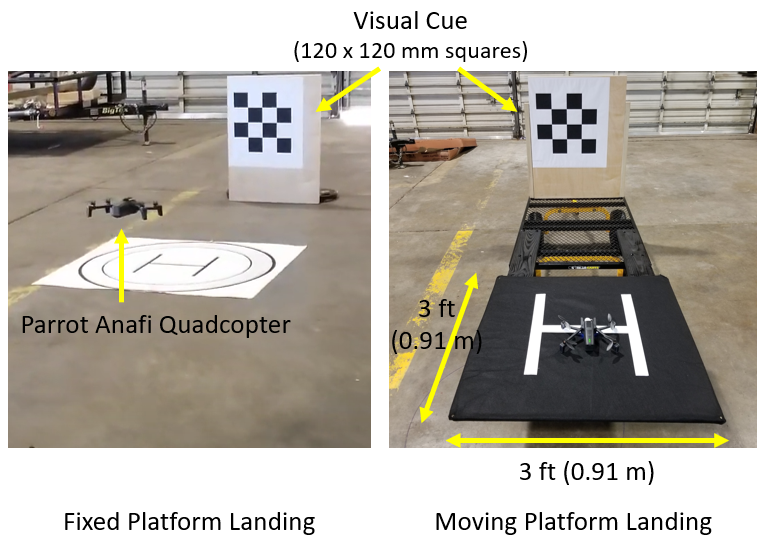}
\caption{Indoor Flight Experimental Setup}
\label{experiment}
\end{figure}

Flight tests are conducted to demonstrate the ability of the developed vision-based autonomous VTOL UAV landing controller to achieve precise landing in practice. The experimental setup for both fixed and moving platform landings is shown in Fig. \ref{experiment}. The visual cue and landing pad dimensions are the same as in simulations. Flights are performed indoor where GPS signal is not available and shown in Videos: \href{https://youtu.be/w0dzVwBzFGk}{Fixed Platform Landing} \cite{fixvideo}, \href{https://youtu.be/IboT80OR1T8}{Moving Platform Landing} \cite{movvideo}.

\begin{figure}[hbt!]
\centering
\includegraphics[width=0.485\textwidth]{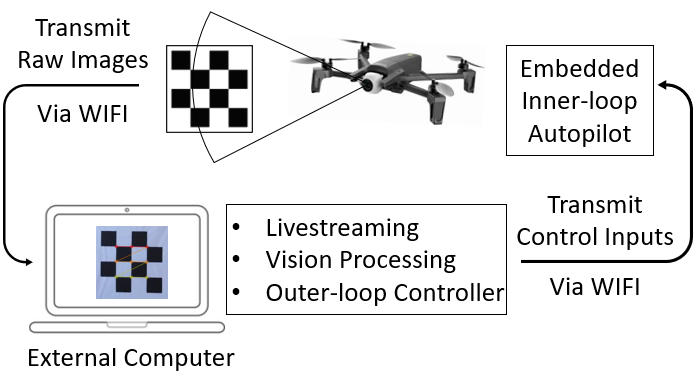}
\caption{Schematic of Flight Testing Process}
\label{process}
\end{figure}

The schematic of the configured vision-based control system is shown in Fig. \ref{process}. 

The UAV is controlled by a Python script that runs on an external base-station computer. The external computer communicates with the UAV through the WIFI connection. The UAV transmits raw images captured by the onboard camera to the computer in real-time. Then, the computer processes images to detect the visual cue and estimate the relative positions and heading angle. The image resolution used is 1280 x 720p and this affects the effective range for detection and estimation. Once the relative positions and heading angle are computed, the feedback control loop yields the corresponding UAV command inputs which are roll, pitch, throttle, and yaw. The commands are sent back to the UAV and then the embedded inner-loop autopilot controls the rotating speed of each propeller to achieve the commanded inputs. One cycle of this whole process takes 0.1 seconds, which is proven to be fast enough to control the UAV behavior. 

Flight tests are progressively conducted from a stationary platform landing to a moving platform landing. It requires camera calibration for the real camera sensor since the simulated camera is designed to only mimic the dimensions and gimbal movement, but not the same sensor quality. Considering flight conditions, it is designed to take 40 previous yaw data for the moving average method, which can reduce yaw fluctuations. Since the update rate is 0.1 seconds, it takes 4 seconds for the moving average method to be in effect. The number of yaw data for the moving average method can be chosen properly according to the flight conditions; for example, the simulations took 10 previous yaw data for the moving average method. Even though realistic simulations for the same flight cases have been done, the flight tests still require more PID gain tuning. The specific gain values obtained from the simulations are used as initial gains for flight tests, and the final gains are determined via multiple flight tests. The control strategy is designed in a way that the UAV stops streaming and makes a quick landing once it achieves the desired landing condition in actual flights. The following results show the relative positions, heading angle, and control inputs while the streaming is operational, which is from after take-off to before landing. More than 100 flight tests are conducted and the representative results are presented.

\subsection{Landing on a stationary platform}

Multiple tests are performed with various initial positions and heading angles. As a representative challenging case of stationary platform landing, 45 degrees of initial yaw angle case is presented. The final landing point is shown in Fig. \ref{set_f}. It is designed to land when the UAV enters the landing threshold which is 15 x 15 cm from the landing pad center. The final landing deviation from the center is 5 cm. 

\begin{figure}[!hbt]
 \raggedleft
   \includegraphics[width=0.9\linewidth]{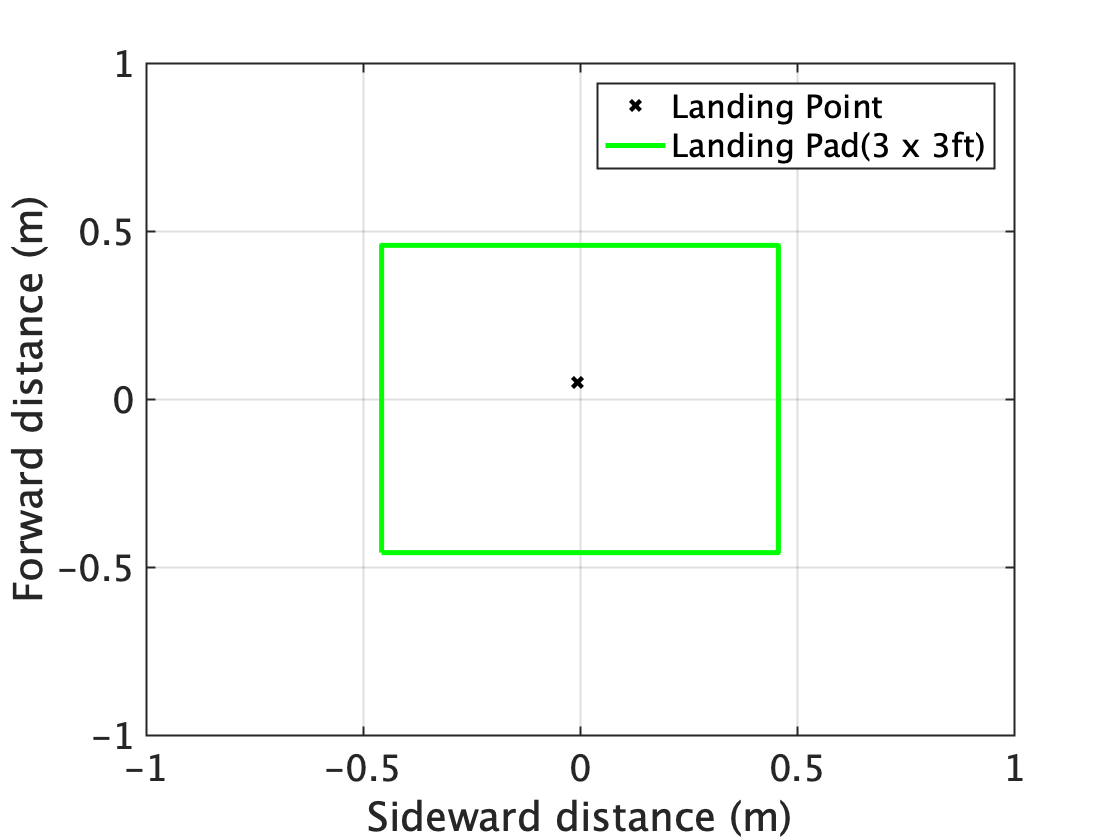}
\caption{Final Landing Point (stationary platform)}
\label{set_f} 
\end{figure}

The time history of forward relative distance and pitch command input are shown in Figs. \ref{forward_f} and \ref{pitch_f}, respectively.

\begin{figure}[!hbt]
 \vspace{-0.3cm}
  \ContinuedFloat* \raggedleft
   \includegraphics[width=0.9\linewidth]{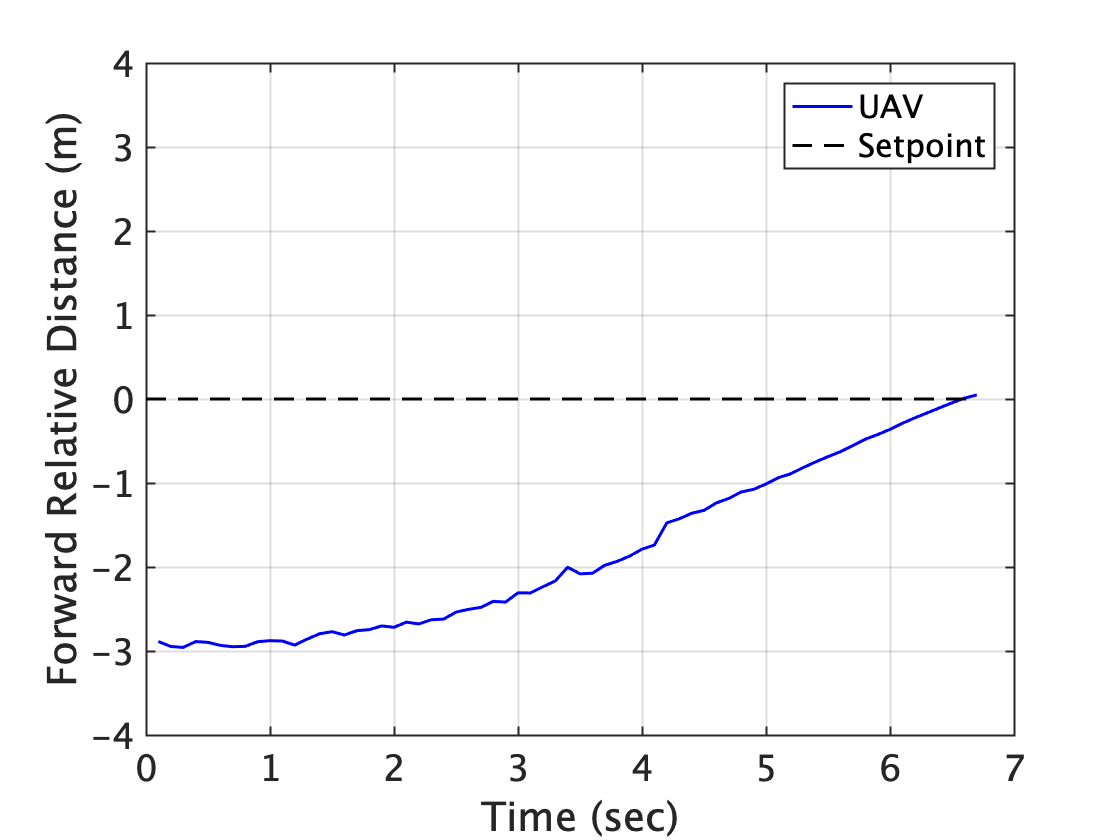}
  \caption{\label{forward_f}Forward Relative Distance (stationary platform)} 
\end{figure}
\begin{figure}[!hbt]
 \vspace{-0.0cm}
  \ContinuedFloat \raggedleft
   \includegraphics[width=0.9\linewidth]{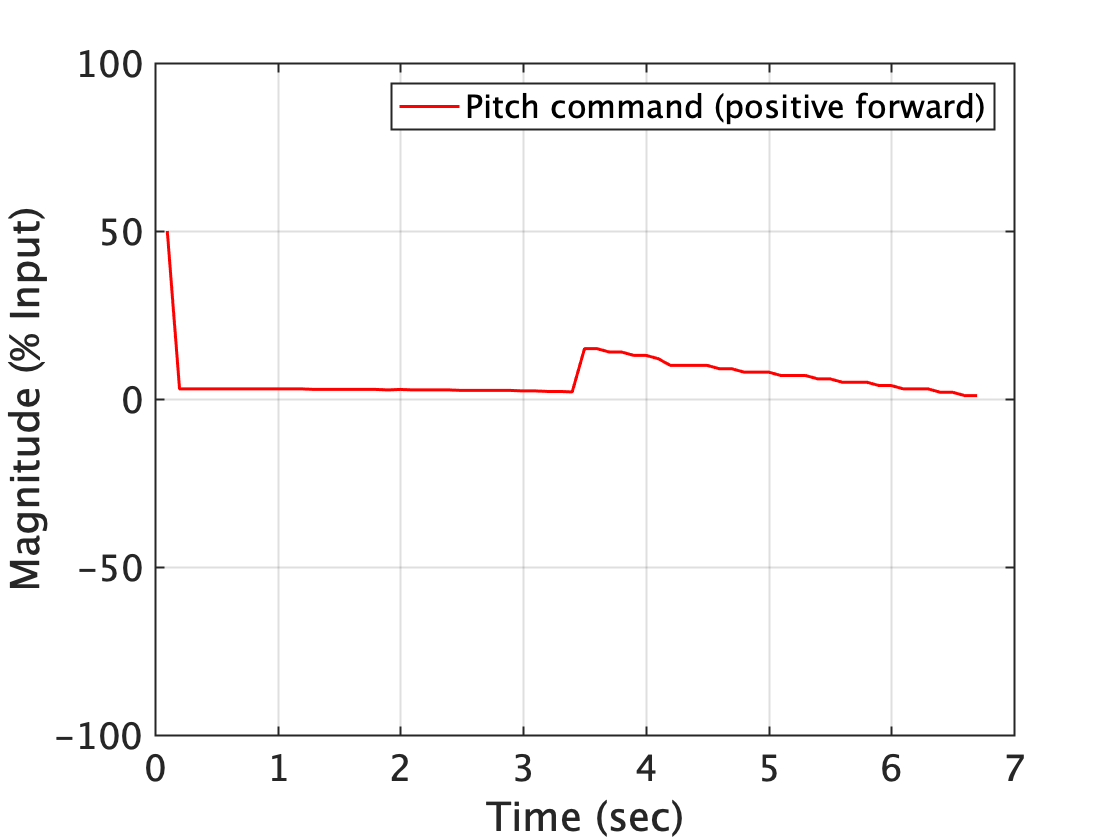}
  \caption{\label{pitch_f}Pitch Command Input (stationary platform)} 
\end{figure}

The time history of sideward relative distance and roll command input during tracking are shown in Figs. \ref{side_f} and \ref{roll_f}, respectively. Up to 3.4 seconds, the magnitudes of pitch and roll commands are small due to the activation of the level 1 yaw controller which gives priority to yaw correction. It takes 6.7 seconds to satisfy the landing conditions.

\begin{figure}[!hbt]
 \vspace{-0.0cm}
  \ContinuedFloat* \raggedleft
   \includegraphics[width=0.9\linewidth]{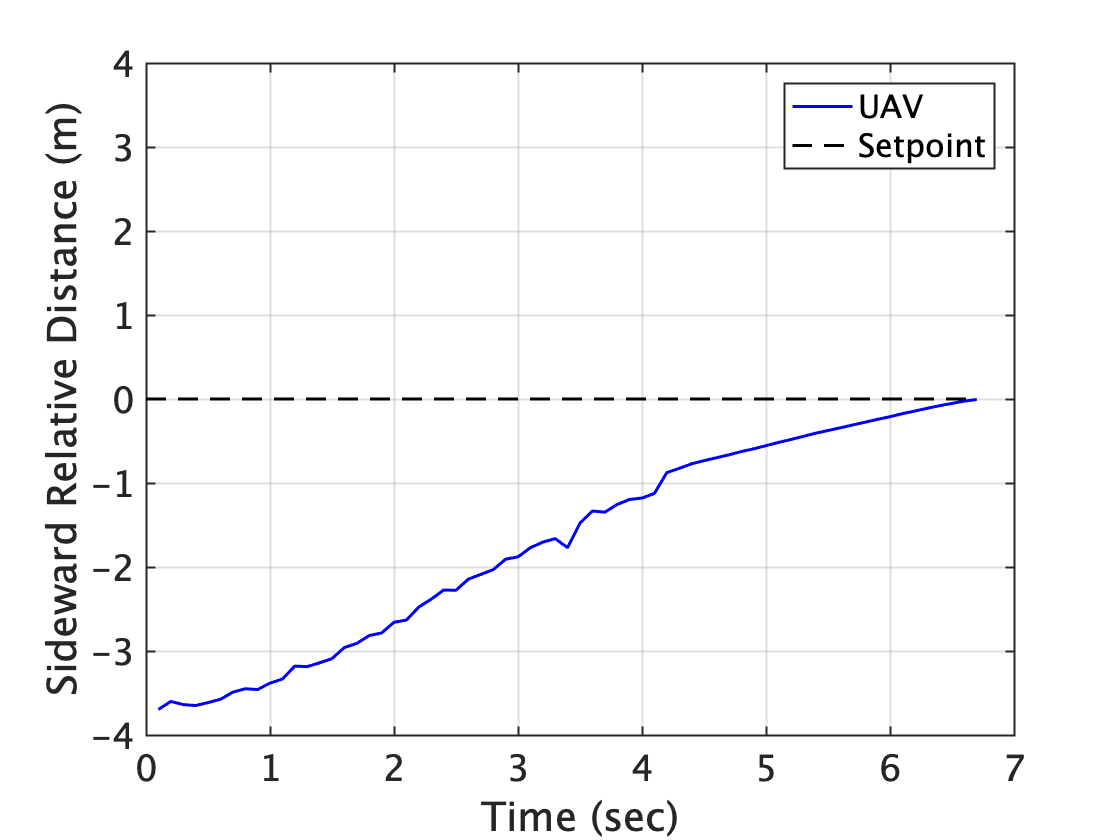}
  \caption{\label{side_f}Sideward Relative Distance (stationary platform)} 
\end{figure}
\begin{figure}[!hbt]
 \vspace{-0.3cm}
  \ContinuedFloat \raggedleft
   \includegraphics[width=0.9\linewidth]{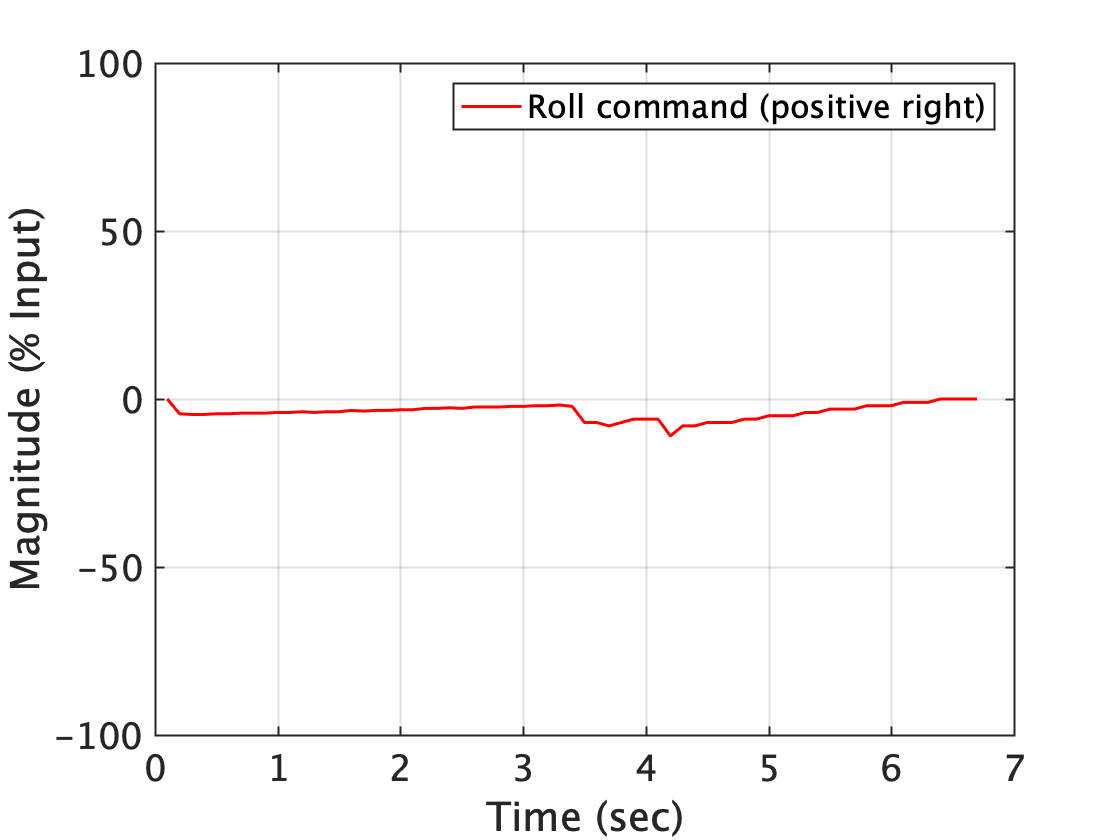}
  \caption{\label{roll_f}Roll Command Input (stationary platform)} 
\end{figure}

The time history of vertical relative distance and throttle command input during tracking are shown in Figs. \ref{vertical_f} and \ref{throttle_f}, respectively.

\begin{figure}[!hbt]
 \vspace{-0.0cm}
  \ContinuedFloat* \raggedleft
   \includegraphics[width=0.9\linewidth]{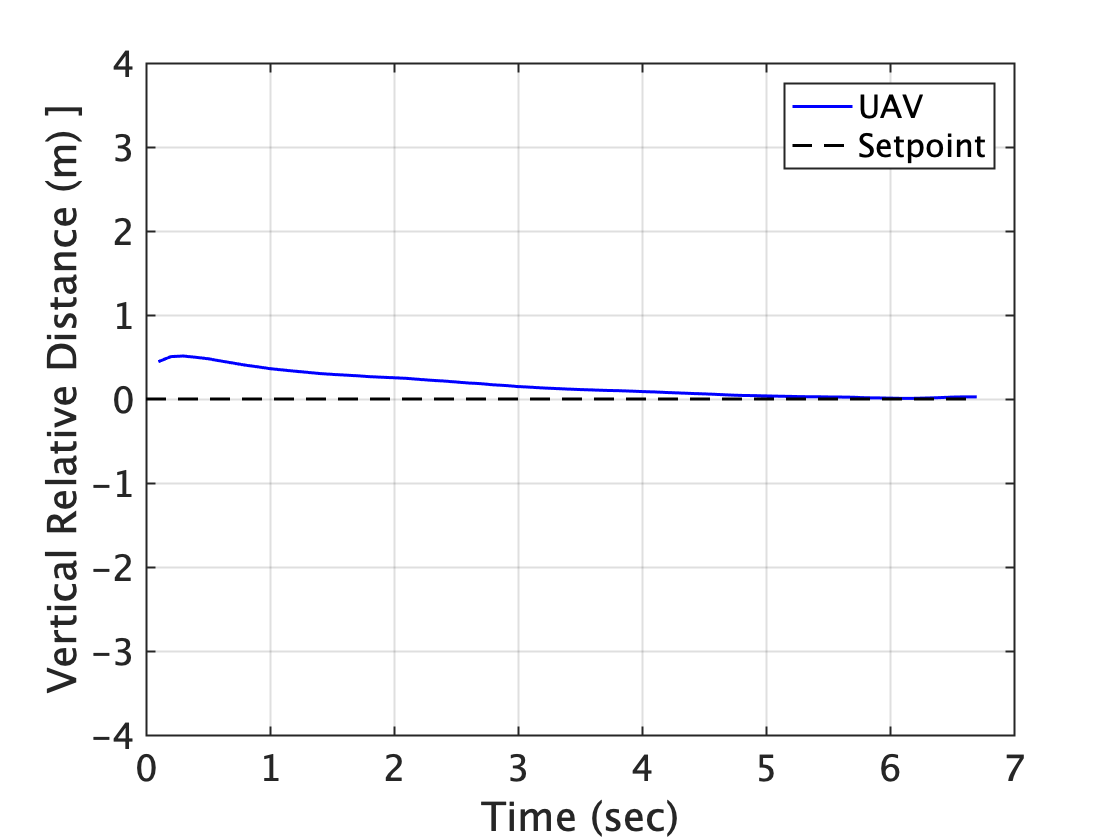}
  \caption{\label{vertical_f}Vertical Relative Distance (stationary platform)} 
\end{figure}
\begin{figure}[!hbt]
 \vspace{-0.0cm}
  \ContinuedFloat \raggedleft
   \includegraphics[width=0.9\linewidth]{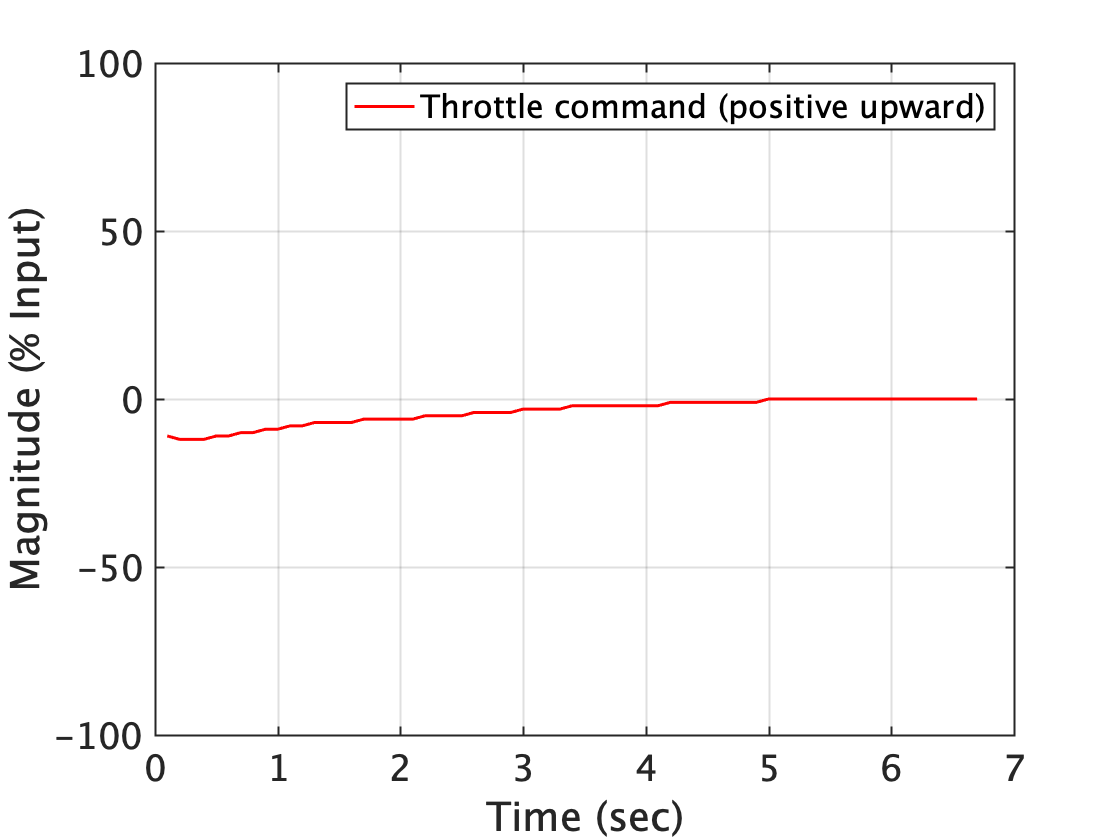}
  \caption{\label{throttle_f}Throttle Command Input (stationary platform)} 
\end{figure}

The time history of relative yaw angle and yaw command input during tracking are shown in Figs. \ref{yaw_f} and \ref{pedal_f}, respectively. Initially, the UAV has 45 degrees of heading angle, which is near the maximum that can be assigned without losing the visual cue in the camera view. Due to the higher image quality of the real camera, the relative position and heading estimates in real experiments experience fewer fluctuations compared to the simulations. Thus, it commands yaw from the beginning instead of applying the safety mode for the directional approach, which had to be activated in the simulations. Until 3.4 seconds, the level 1 yaw controller corrects yaw and also reduces the magnitudes of pitch and roll commands not to lose the visual cue in the camera view when the yaw angle is large. After 4.1 seconds, the estimation and control commands become smoother due to the moving average method. By the time the UAV satisfies the landing conditions, the vertical distance is regulated to the visual cue height and the yaw angle is decreased to zero degrees.

\begin{figure}[!hbt]
  \ContinuedFloat* \raggedleft
   \includegraphics[width=0.9\linewidth]{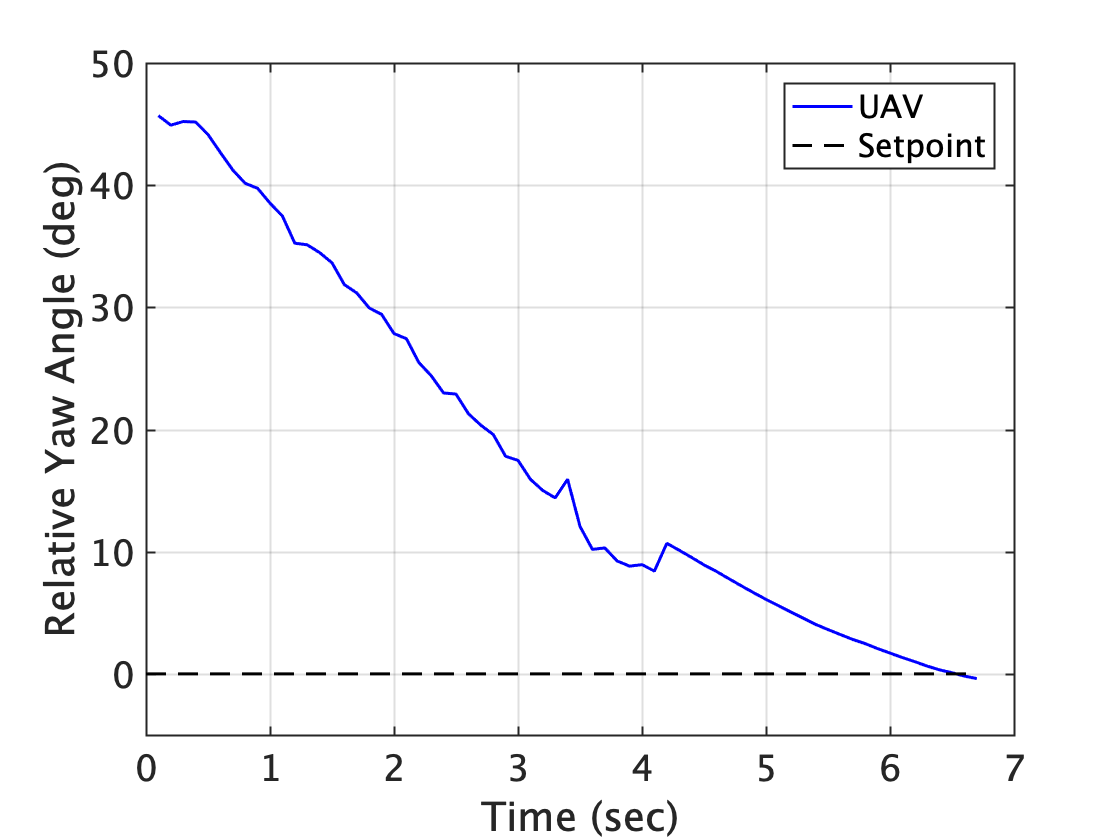}
  \caption{\label{yaw_f}Relative Yaw Angle (stationary platform)} 
\end{figure}
\begin{figure}[!hbt]
  \ContinuedFloat \raggedleft
   \includegraphics[width=0.9\linewidth]{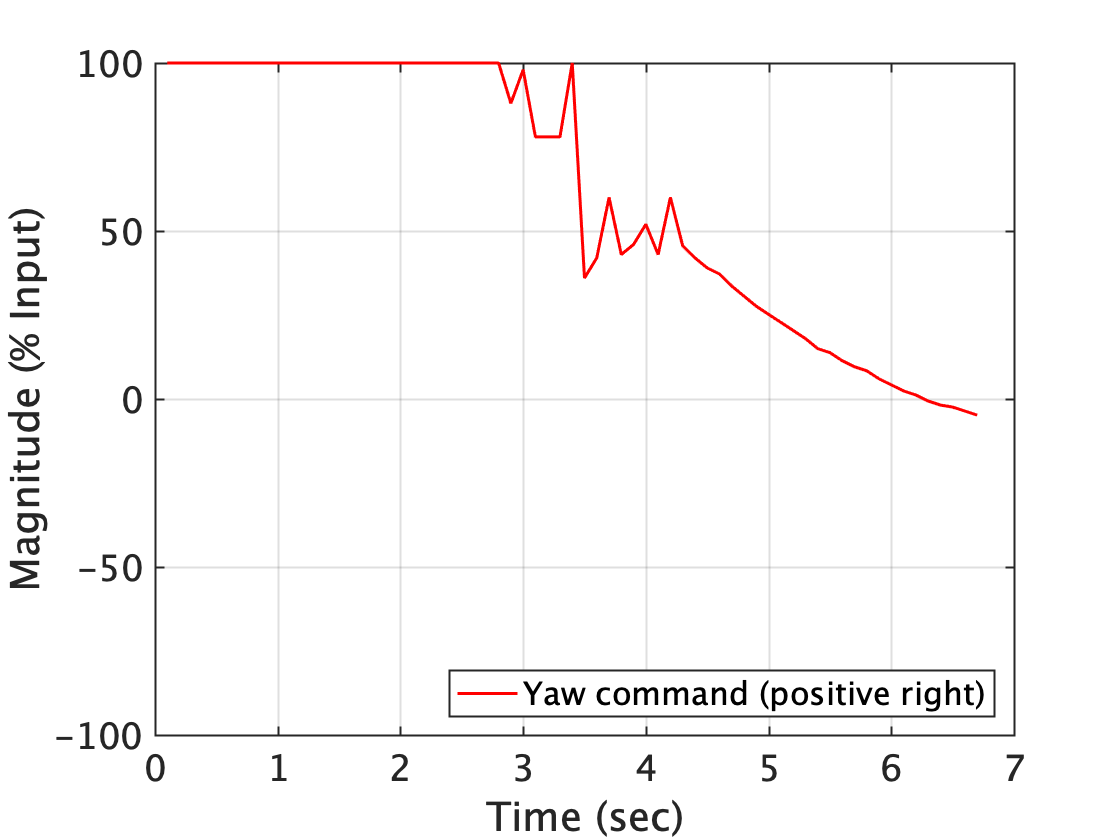}
  \caption{\label{pedal_f}Yaw Command Input (stationary platform)} 
\end{figure}

\subsection{Landing on a moving platform}

Multiple tests are performed while the platform is slowly moving forward and the results from a representative case are presented in detail. The landing deviation is 4 cm as shown in Fig. \ref{set_m}.

\begin{figure}[!hbt]
 \vspace{-0.3cm}
 \raggedleft
   \includegraphics[width=0.9\linewidth]{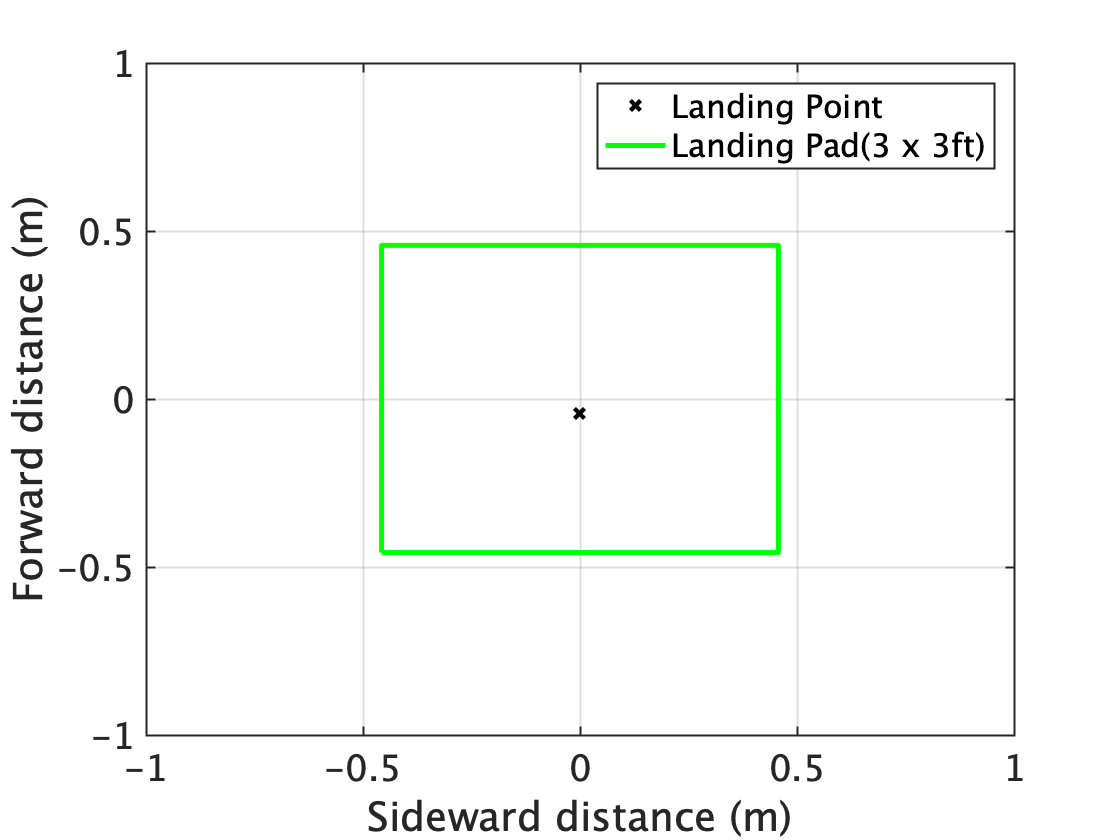}
\caption{Final Landing Point (moving platform)}
\label{set_m} 
\end{figure}

The time history of forward relative distance and pitch command input during tracking are shown in Figs. \ref{forward_m} and \ref{pitch_m}, respectively.

\begin{figure}[!hbt]
 \vspace{-0.0cm}
  \ContinuedFloat*  \raggedleft
   \includegraphics[width=0.9\linewidth]{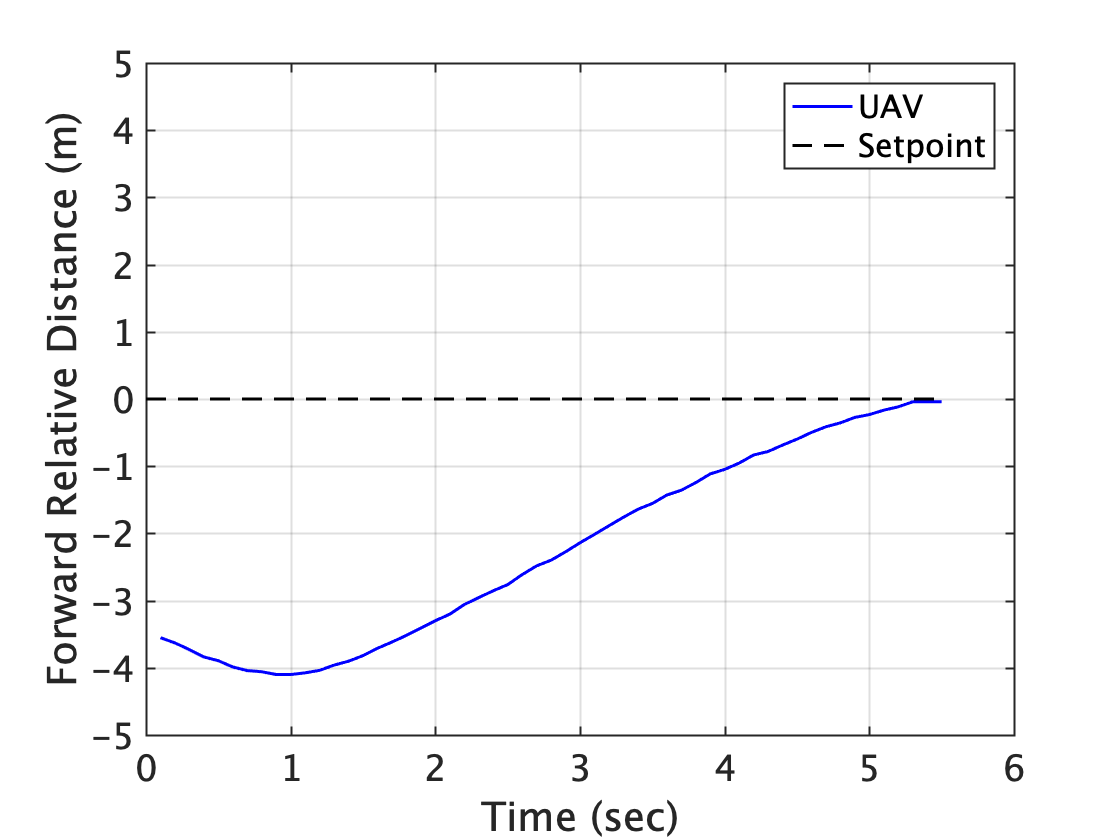}
  \caption{\label{forward_m}Forward Relative Distance (moving platform)} 
\end{figure}
\begin{figure}[!hbt]
 \vspace{0.3cm}
  \ContinuedFloat  \raggedleft
   \includegraphics[width=0.9\linewidth]{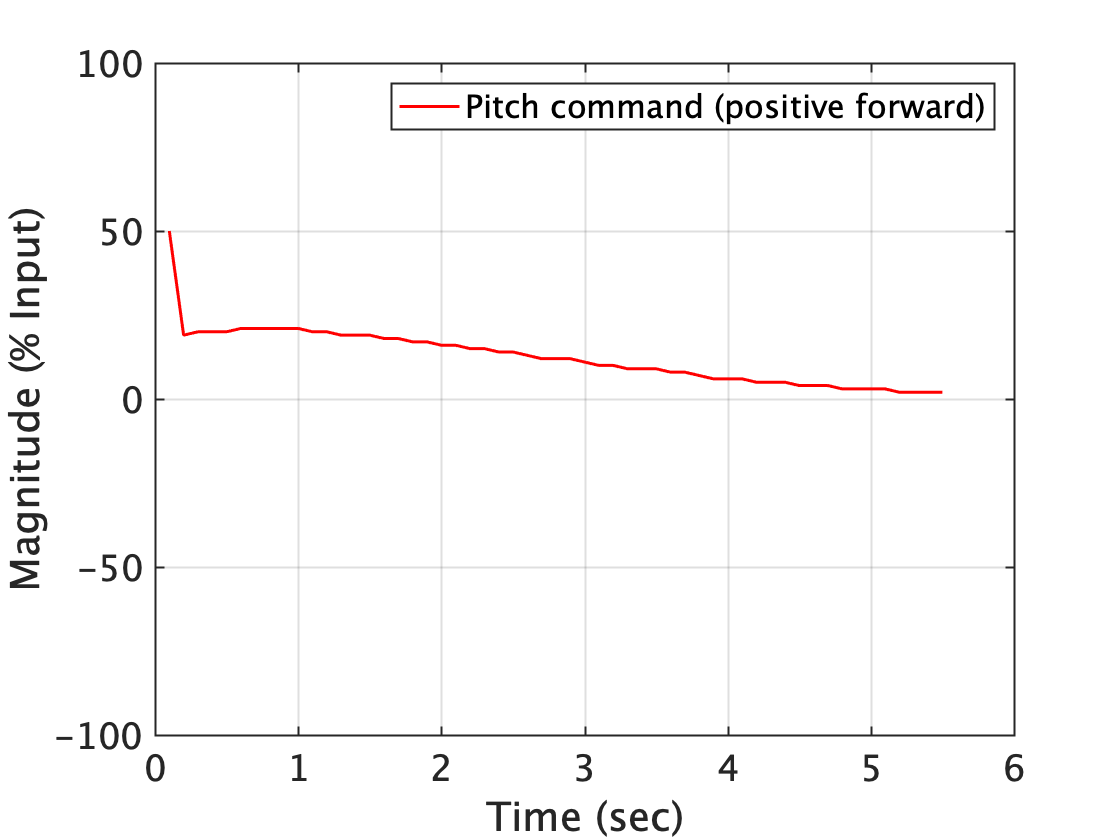}
  \caption{\label{pitch_m}Pitch Command Input (moving platform)} 
\end{figure}

The time history of sideward relative distance and roll command input during tracking are shown in Figs. \ref{side_m} and \ref{roll_m}, respectively.

\begin{figure}[!hbt]
 \vspace{-0.3cm}
  \ContinuedFloat*  \raggedleft
   \includegraphics[width=0.9\linewidth]{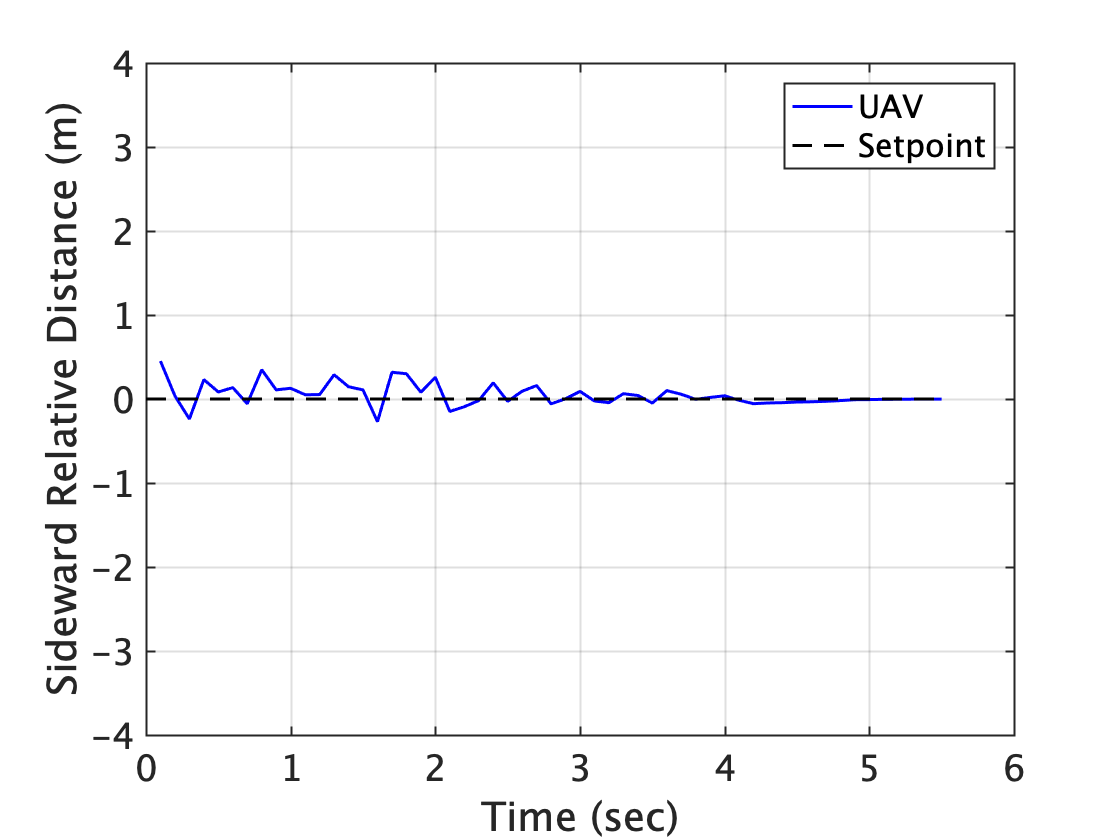}
  \caption{\label{side_m}Sideward Relative Distance (moving platform)} 
\end{figure}
\begin{figure}[!hbt]
 \vspace{-0.5cm}
  \ContinuedFloat  \raggedleft
   \includegraphics[width=0.9\linewidth]{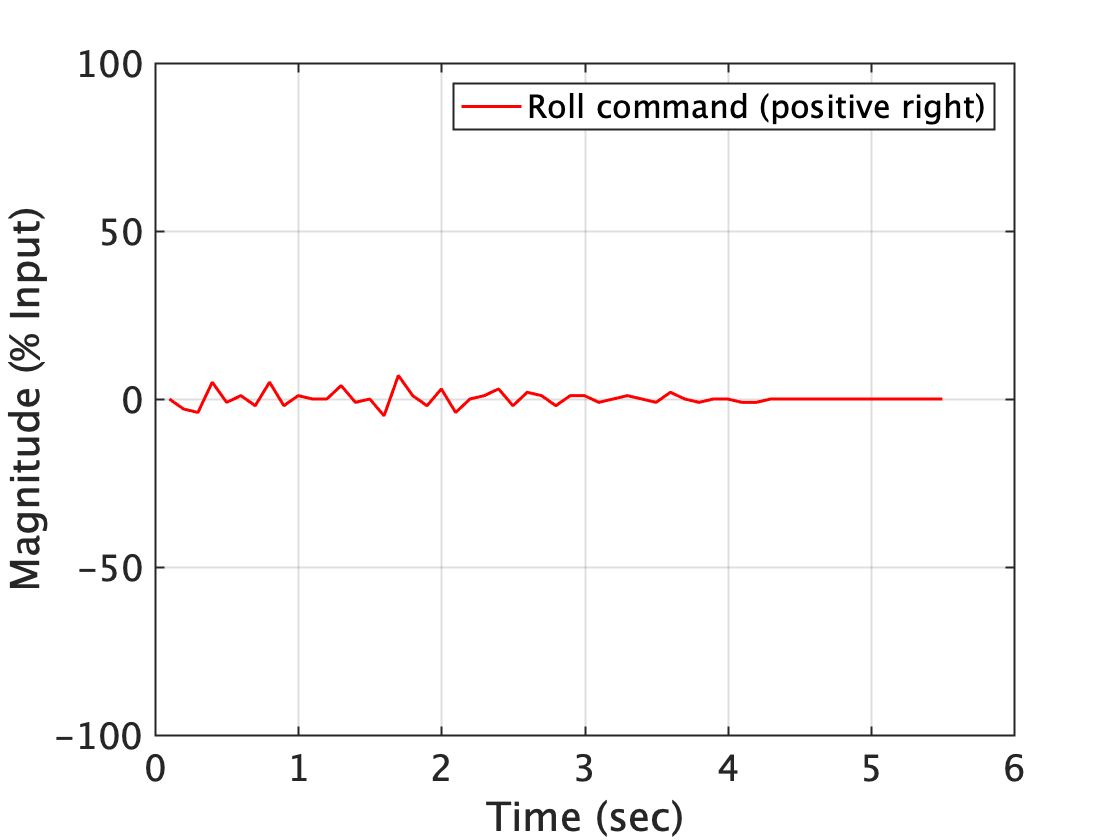}
  \caption{\label{roll_m}Roll Command Input (moving platform)} 
\end{figure}

The time history of vertical relative distance and throttle command input during tracking are shown in Figs. \ref{vertical_m} and \ref{throttle_m}, respectively. The forward distance increases due to the platform already moving in the beginning, and sideward distance is kept small throughout the flight. The setpoint for the vertical relative distance is the visual cue height on the moving platform and it is almost the same height when the UAV starts streaming right after take-off. Thus, it maintains the height until landing. The landing conditions are satisfied at 5.5 seconds, it stops streaming and executes landing.

\begin{figure}[!hbt]
 \vspace{-0.4cm}
  \ContinuedFloat*  \raggedleft
   \includegraphics[width=0.9\linewidth]{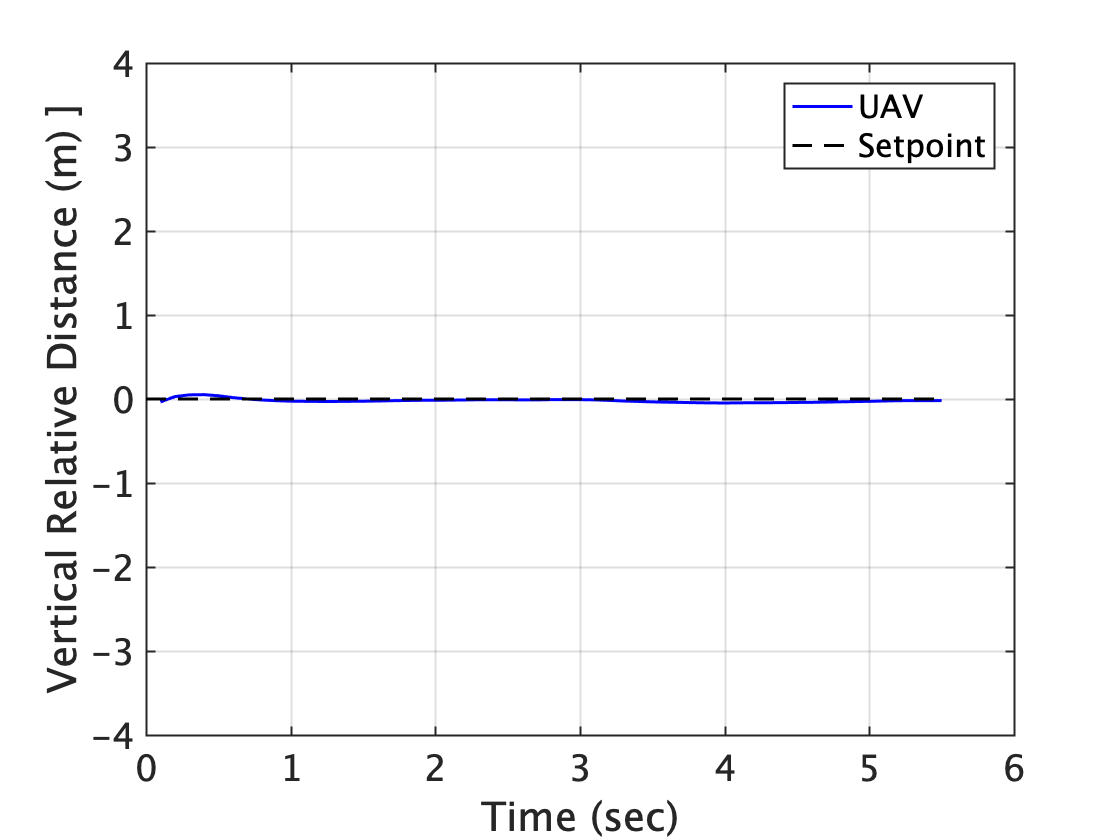}
  \caption{\label{vertical_m}Vertical Relative Distance (moving platform)} 
\end{figure}
\begin{figure}[!hbt]
 \vspace{-0.3cm}
  \ContinuedFloat  \raggedleft
   \includegraphics[width=0.9\linewidth]{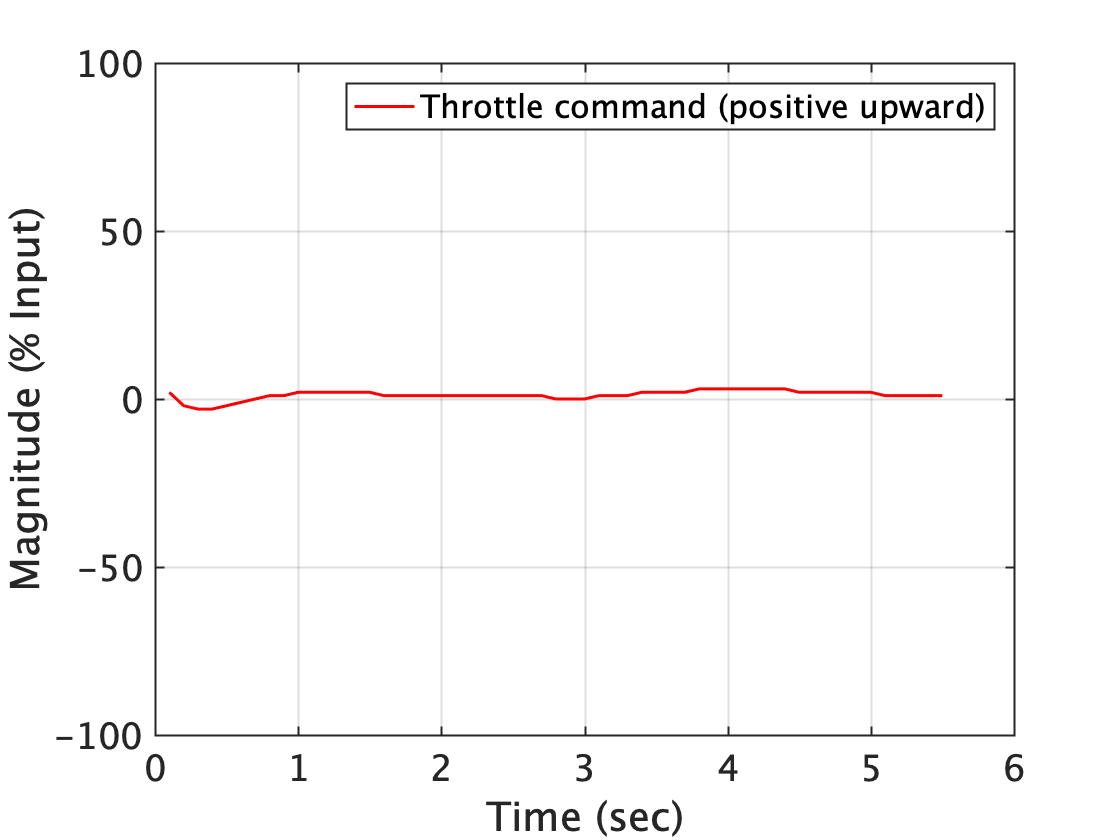}
  \caption{\label{throttle_m}Throttle Command Input (moving platform)} 
\end{figure}

The relative yaw angle and yaw command input during tracking are shown in Fig. \ref{yaw_m} and \ref{pedal_m}, respectively. Since the initial heading angle of the UAV is intended to be parallel to the visual cue, it has small yaws during flight. The fluctuations that exist until 4.1 seconds are at the noise level which is $\pm$ 1 degree and become smooth after 4.1 seconds as the moving average method is activated. When the UAV satisfies the landing condition at 5.5 seconds, the final yaw angle is less than 1 degree.

\begin{figure}[!hbt]
 \vspace{-0.4cm}
  \ContinuedFloat*  \raggedleft
   \includegraphics[width=0.9\linewidth]{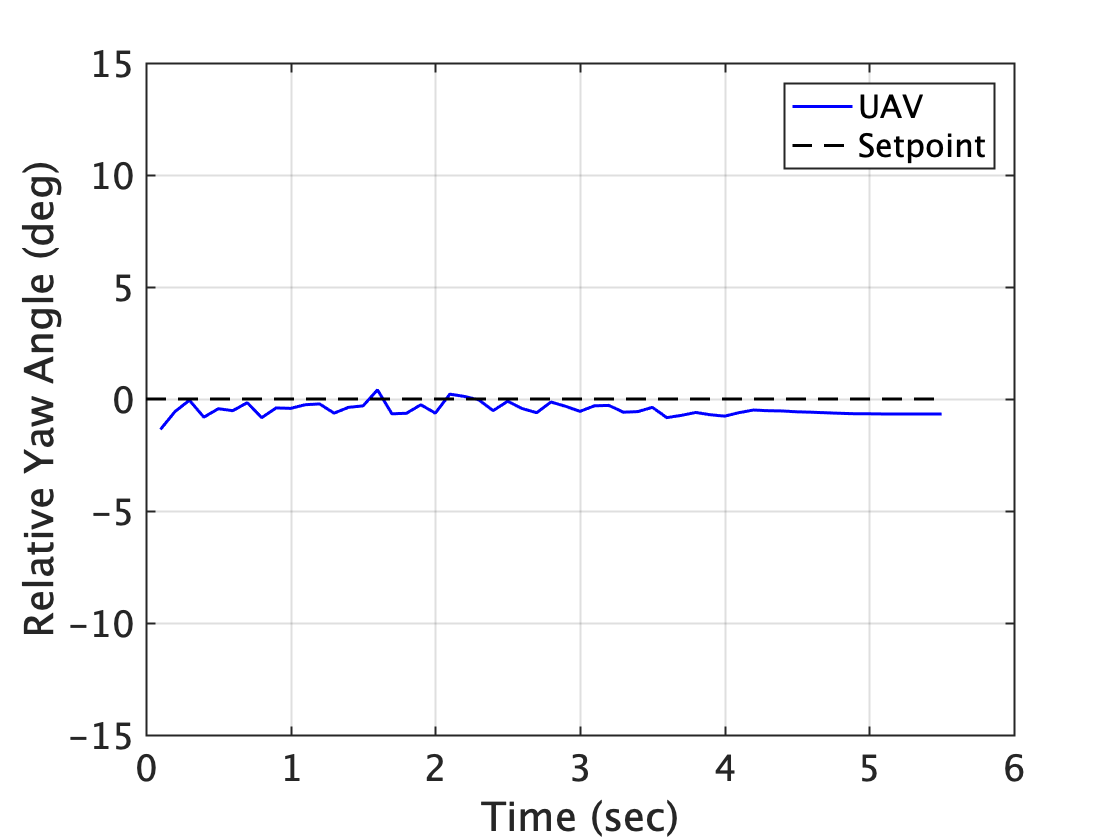}
  \caption{\label{yaw_m}Relative Yaw Angle (moving platform)} 
\end{figure}
\begin{figure}[!hbt]
 \vspace{-0.7cm}
  \ContinuedFloat  \raggedleft
   \includegraphics[width=0.9\linewidth]{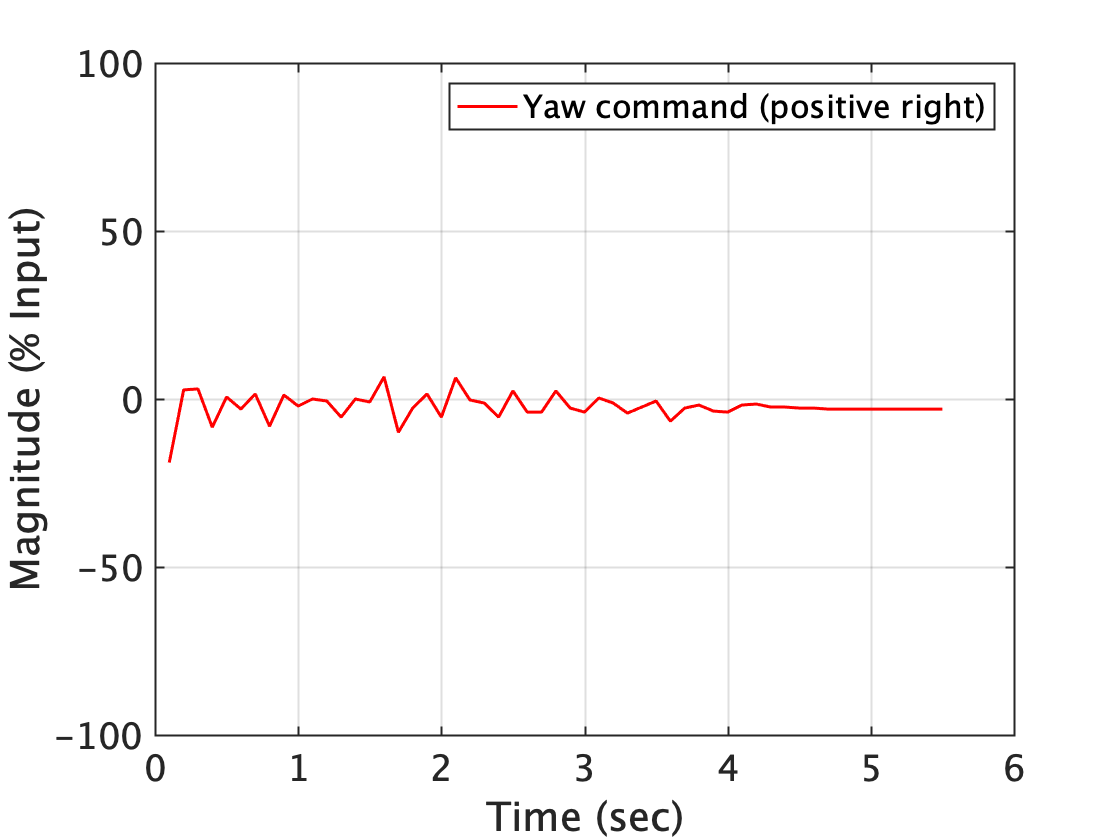}
  \caption{\label{pedal_m}Yaw Command Input (moving platform)} 
\end{figure}

\section{Summary and Conclusions}

The study focuses on the development and validation (via simulations and flight testing) of a novel method inspired by manned helicopter ship landing procedure for autonomous tracking and landing of VTOL UAVs on moving platforms without relying on a GPS signal. The main novelty of this method is that instead of looking at the landing spot, it utilizes a standardized visual cue to track, approach and land on the moving platform analogous to a pilot landing a helicopter on a moving ship by referencing the horizon bar (a visual aid).

To implement this method a computer vision system is developed, which enables long-range detection and precise pose estimation by incorporating improved F\"orstner corner detection and PnP estimation algorithm with a moving average filter. With the use of a checkerboard type visual cue formed by 120mm squares, its maximum detection range is 25 meters. Within this range, by correlation with a Vicon-based motion capture system, it is demonstrated that the present vision system has sub-centimeter position accuracy and sub-degree heading angle accuracy. The vision-based control system is configured with gain-scheduled PID controllers along with different flight modes, which allows a smart, situation-adaptive control approach to the landing maneuver, rather than blindly moving to the landing pad. Based on the flight conditions and the relative distance to the landing platform, the controller will choose from the following modes: high-speed gross tracking, precise tracking at low speeds, scanning for the visual cue, collision prevention, or immediate landing. The different flight modes are verified through the simulations under various flight scenarios including the landing platform moving at high speeds, as well as following circular and S-shaped trajectories. The simulation results confirmed that the proposed visual cue tracking landing method is able to achieve safe, accurate, and robust landings repeatedly. The factor that affects landing accuracy is the assigned landing threshold and there is a trade-off between the accuracy and time. Even in a scenario where the UAV had to fly at its maximum speed, it landed with an accuracy of 4 cm when a landing threshold was set to 5 x 5 cm from the pad center. The transition from the simulation to flight testing required only a few changes such as camera calibration and minor adjustment in gains. Hundreds of flight tests were safely conducted, which included both stationary and moving platform autonomous landings with a landing error of less than 5 cm. 

The results presented in this paper demonstrate conclusively the feasibility of this novel autonomous approach and landing strategy. In addition, this method offers additional freedom in the selection of the approach trajectory and height since it involves tracking a visual cue installed parallel to the aircraft's view, which could be useful in real-life applications. The next step in this study is to improve the robustness of this method under uncertainties by bringing in the experience and decision-making abilities of a human pilot via machine learning approach.

\newpage
\textbf{Author contact:} Bochan Lee \href{mailto:bochan.lee@tamu.edu}{bochan.lee@tamu.edu}
\\\hspace*{2.5cm}Vishnu Saj \href{mailto:vishnu02saj@tamu.edu}{vishnu02saj@tamu.edu} 
\\\hspace*{2.5cm}Moble Benedict \href{mailto:benedict@tamu.edu}{benedict@tamu.edu}
\\\hspace*{2.5cm}Dileep Kalathil \href{mailto:dileep.kalathil@tamu.edu}{dileep.kalathil@tamu.edu}

\bibliographystyle{ahs}
\label{sec:references}\bibliography{ahsrefs}


\begin{thebibliography}{10}
\newcommand{\enquote}[1]{``#1''}

\bibitem{Ruptly}
Ruptly, \enquote{Helicopter Ship Landing at Rough Sea,}
  \url{https://youtu.be/lDISL-jsF-Q?t=60}, 2016 (accessed June 30, 2020).

\bibitem{lumsden1998challenges}
Lumsden, B., Wilkinson, C., and Padfield, G., \enquote{Challenges at the
  helicopter-ship dynamic interface,} , 1998.

\bibitem{colwell2002maritime}
Colwell, J., \enquote{Maritime helicopter ship motion criteria-Challenges for
  operational guidance,} \emph{Challenges for Operational Guidance-NATO RTO
  Systems Concepts and Integration Panel SCI-120. Berlin, Germany}, 2002.

\bibitem{stingl1970vtol}
Stingl, A.~L., \enquote{Vtol aircraft flight system,} US Patent 3,487,553,
  January~6 1970.

\bibitem{nato}
\enquote{Helicopter Operations from Ships Other Than Aircraft
  Carriers(HOSTAC),} Vol.~I, 2017.

\bibitem{lee2012autonomous}
Lee, D., Ryan, T., and Kim, H.~J., \enquote{Autonomous landing of a VTOL UAV on
  a moving platform using image-based visual servoing,} 2012 IEEE international
  conference on robotics and automation, 2012.

\bibitem{wenzel2011automatic}
Wenzel, K.~E., Masselli, A., and Zell, A., \enquote{Automatic take off,
  tracking and landing of a miniature UAV on a moving carrier vehicle,}
  \emph{Journal of intelligent \& robotic systems}, Vol.~61,~(1-4), 2011,
  pp.~221--238.

\bibitem{vicon}
\enquote{Vicon motion capture system,} \url{http://www.vicon.com/}.

\bibitem{daly2015coordinated}
Daly, J.~M., Ma, Y., and Waslander, S.~L., \enquote{Coordinated landing of a
  quadrotor on a skid-steered ground vehicle in the presence of time delays,}
  \emph{Autonomous Robots}, Vol.~38,~(2), 2015, pp.~179--191.

\bibitem{ghamry2016real}
Ghamry, K.~A., Dong, Y., Kamel, M.~A., and Zhang, Y., \enquote{Real-time
  autonomous take-off, tracking and landing of UAV on a moving UGV platform,}
  2016 24th Mediterranean conference on control and automation (MED), 2016.

\bibitem{araar2017vision}
Araar, O., Aouf, N., and Vitanov, I., \enquote{Vision based autonomous landing
  of multirotor UAV on moving platform,} \emph{Journal of Intelligent \&
  Robotic Systems}, Vol.~85,~(2), 2017, pp.~369--384.

\bibitem{falanga2017vision}
Falanga, D., Zanchettin, A., Simovic, A., Delmerico, J., and Scaramuzza, D.,
  \enquote{Vision-based autonomous quadrotor landing on a moving platform,}
  2017 IEEE International Symposium on Safety, Security and Rescue Robotics
  (SSRR), 2017.

\bibitem{bi2013implementation}
Bi, Y., and Duan, H., \enquote{Implementation of autonomous visual tracking and
  landing for a low-cost quadrotor,} \emph{Optik-International Journal for
  Light and Electron Optics}, Vol.~124,~(18), 2013, pp.~3296--3300.

\bibitem{herisse2011landing}
Heriss{\'e}, B., Hamel, T., Mahony, R., and Russotto, F.-X., \enquote{Landing a
  VTOL unmanned aerial vehicle on a moving platform using optical flow,}
  \emph{IEEE Transactions on robotics}, Vol.~28,~(1), 2011, pp.~77--89.

\bibitem{gazebo}
\enquote{Gazebo simulation program,} \url{http://gazebosim.org/}.

\bibitem{compvision1}
Harris, C., and Stephens, M., \enquote{A combined corner and edge detector,}
  \emph{International Journal of Computer Vision}, 1988.
  DOI:~10.1023/A:1007963824710

\bibitem{compvision2}
Lindeberg, T., \enquote{Edge detection and ridge detection with automatic scale
  selection,} \emph{International Journal of Computer Vision}, 1988.
  DOI:~10.1023/A:1008097225773

\bibitem{compvision3}
Ethan~Rublee, K.~K., Vincent~Rabaud, and Bradski, G., \enquote{ORB: An
  efficient alternative to SIFT or SURF,} International Conference on Computer
  Vision, 2011. DOI:~10.1109/ICCV.2011.6126544

\bibitem{projtrans}
Hartley, R., and Zisserman, A., \emph{Multiple View Geometry in Computer
  Vision}, Cambridge University Press, 2003.

\bibitem{Cheng01colorimage}
Cheng, H.~D., Jiang, X.~H., Sun, Y., and Wang, J.~L., \enquote{Color image
  segmentation: Advances and prospects,} \emph{Pattern Recognition}, Vol.~34,
  2001, pp.~2259--2281.

\bibitem{forstner1987fast}
F\"orstner, W., and G\"ulch, E., \enquote{A Fast Operator for Detection and
  Precise Location of Distict Point, Corners and Centres of Circular Features,}
  Proceedings of the ISPRS Conference on Fast Processing of Photogrammetric
  Data, 1987.

\bibitem{findchess}
Bradski, G., \enquote{The OpenCV Library,} \emph{Dr Dobbs Journal of Software
  Tools}, 2000. DOI:~10.1111/0023-8333.50.s1.10.

\bibitem{est1}
Zhang, Z., \enquote{Flexible Camera Calibration By Viewing a Plane From Unknown
  Orientations,} \emph{International Conference on Computer Vision}, 1999.

\bibitem{est2}
Nozomu~Araki, Y.~K., Takao~Sato, and Ishigaki, H., \enquote{Vehicle's
  orientation measurement method by single camera image using known shaped
  planar object,} \emph{International Journal of Innovative Computing,
  Information and Control}, 2011.

\bibitem{solvepnp}
OPenCV, \enquote{Camera Calibration and 3D Reconstruction,}
  \url{https://docs.opencv.org/master/d9/d0c/group__calib3d.html}, 2020
  (accessed June 30, 2020).

\bibitem{lev1}
Levenberg, K., \enquote{A Method for the Solution of Certain Non-Linear
  Problems in Least Squares,} \emph{Quarterly of Applied Mathematics}, 1944.
  DOI:~10.1090/qam/10666

\bibitem{lev2}
Marquardt, D.~W., \enquote{An Algorithm for Least-Squares Estimation of
  Nonlinear Parameters,} \emph{SIAM Journal on Applied Mathematics}, 1963.
  DOI:~10.1137/0111030

\bibitem{ransac}
Bolles, M. A. F. . R.~C., \enquote{Random Sample Consensus: A Paradigm for
  Model Fitting with Applications to Image Analysis and Automated Cartography,}
  , 1981. DOI:~10.1145/358669.358692

\bibitem{ang2005pid}
Ang, K.~H., Chong, G., and Li, Y., \enquote{PID control system analysis,
  design, and technology,} \emph{IEEE transactions on control systems
  technology}, Vol.~13,~(4), 2005, pp.~559--576.

\bibitem{simvideo}
Lee, B., \enquote{Vision-based VTOL UAV Landings on a Moving Platform in Gazebo
  Simulation,} \url{https://youtu.be/8kAeVzGJyN8}, 2020 (accessed June 30,
  2020).

\bibitem{fixvideo}
Lee, B., \enquote{Flight Testing of Vision-based VTOL UAV Landings on a Fixed
  Platform,} \url{https://youtu.be/w0dzVwBzFGk}, 2020 (accessed June 30, 2020).

\bibitem{movvideo}
Lee, B., \enquote{Flight Testing of Vision-based VTOL UAV Landings on a Moving
  Platform,} \url{https://youtu.be/IboT80OR1T8}, 2020 (accessed June 30, 2020).

\end{thebibliography}

\end{document}